\documentclass{article}

\usepackage[preprint]{neurips_2026}

\usepackage[utf8]{inputenc} %
\usepackage[T1]{fontenc}    %
\usepackage{hyperref}       %
\usepackage{url}            %
\usepackage{booktabs}       %
\usepackage{amsfonts}       %
\usepackage{nicefrac}       %
\usepackage{microtype}      %
\usepackage{xcolor}         %

\usepackage{amsmath}        %
\usepackage{amssymb}        %
\usepackage{graphicx}
\usepackage{wrapfig}
\usepackage{multirow}
\usepackage{float}
\usepackage{caption}
\usepackage{subfig}         %
\usepackage{mathrsfs}
\usepackage[scr=boondoxo]{mathalfa}
\usepackage{etoc}
\etocsetlevel{title}{6}
\etocsetlevel{author}{6}
\providecommand{\authcount}[1]{}
\usepackage{algorithm2e}

\graphicspath{{./}}

\newcommand{\R}{\mathbb{R}}
\newcommand{\grad}[1]{\nabla#1}
\newcommand{\jac}[1]{\mbox{\textbf{J}} #1}
\newcommand{\cdott}{\boldsymbol{\cdot}}
\newcommand\norm[1]{\left\lVert#1\right\rVert}

\newcommand{\method}{\textsf{M-plicits}}

\title{\method{}: Neural Implicit Surfaces via Nested Multiscale Residuals}

\author{%
  Vinícius da Silva\textsuperscript{1}\thanks{Corresponding author: \texttt{viniciuss@tecgraf.puc-rio.br}} \quad
  Isabelle Melo\textsuperscript{1} \quad
  Matheus Bessa\textsuperscript{1} \quad
  Guilherme Schardong\textsuperscript{2} \\[2pt]
  \bfseries Luiz Schirmer\textsuperscript{2} \quad
  André Araújo\textsuperscript{3} \quad
  Nuno Gonçalves\textsuperscript{4} \quad
  Hélio Lopes\textsuperscript{1} \\[2pt]
  \bfseries Alberto Raposo\textsuperscript{1} \quad
  Luiz Velho\textsuperscript{5} \quad
  Tiago Novello\textsuperscript{5} \\[6pt]
  {\footnotesize \textsuperscript{1}PUC-Rio \quad \textsuperscript{2}Universidade Federal de Santa Maria \quad
  \textsuperscript{3}Google DeepMind \quad \textsuperscript{4}University of Coimbra \quad \textsuperscript{5}IMPA}
}

\begin{document}

\maketitle

\begin{abstract}
Encoding input coordinates with sinusoidal functions into multi-layer perceptrons (MLPs) has proven effective for implicit neural representations (INRs) of surfaces defined as zero-level sets. However, existing methods often struggle to balance training efficiency, rendering speed, and noise robustness: single-MLP approaches are expensive at inference, grid-based representations are fast but can limit surface smoothness and overfit input noise, and previous multiscale approaches frequently capture noise and produce artifacts due to hard spectral truncation. To address these limitations, we propose M-plicits, a multiscale framework that models surfaces as a residual sum of MLPs trained via a sequence of nested neighborhoods. Unlike existing residual approaches that rely on standard domain-wide sampling and require costly mesh extraction for visualization, our method strictly localizes supervision to narrow bands around the previous zero-level sets. This nested design naturally provides robustness against noisy input data: the coarse network acts as a low-pass filter that establishes a clean geometric prior, while subsequent residuals progressively refine the geometry without fitting to high-frequency artifacts. We further introduce a multiscale sphere-tracing algorithm and a GEMM-based analytical normal computation that bypasses auto-differentiation entirely, yielding high-fidelity real-time rendering. On Stanford and Thingi32, M-plicits achieves the best mean Chamfer distance in the coarse configuration and the best median Chamfer distance and IoU in the fine configuration, with substantially better noise robustness than iNGP, BACON, and IDF, while using an order of magnitude fewer parameters than grid-based baselines. Code, models, and data are available at \url{https://github.com/dsilvavinicius/m-plicits}.
\end{abstract}

\section{Introduction}
\label{sec:intro}

Reconstructing surfaces from point clouds is a long-standing problem in vision and graphics~\cite{kazhdan2006poisson}, with applications in augmented/virtual reality~\cite{tkach2016sphere}, digital twins~\cite{sun2005bio}, cultural heritage preservation~\cite{scopigno20113d}, and autonomous robotics~\cite{whelan2016elasticfusion}—where high-quality 3D geometry is essential for perception, interaction, and decision-making. The emergence of high-resolution depth sensors has further motivated the development of accurate and efficient surface reconstruction methods.

Representing the surfaces as zero-level sets of multi-layer perceptrons (MLPs) has become a prominent approach, due to their strong representational capacity and the ability to incorporate geometric regularizations~\cite{wang2021neus}.
To enhance model expressiveness, input coordinates are projected into a set of sinusoidal functions~\cite{tancik2020ffm, novello2024taming}, allowing the network's bandlimit to be controlled through appropriate frequency initialization.
For geometric regularization, it is common to assume that the underlying function represents a signed distance function (SDF) of the ground-truth surface~\cite{schirmer2024geometric}, which satisfies the Eikonal equation.
Incorporating this constraint into the implicit neural representation (INR) loss function acts as geometric regularization, enforcing the SDF property away from the data.

\begin{figure}[t!]
\centering
\includegraphics[width=0.9\linewidth]{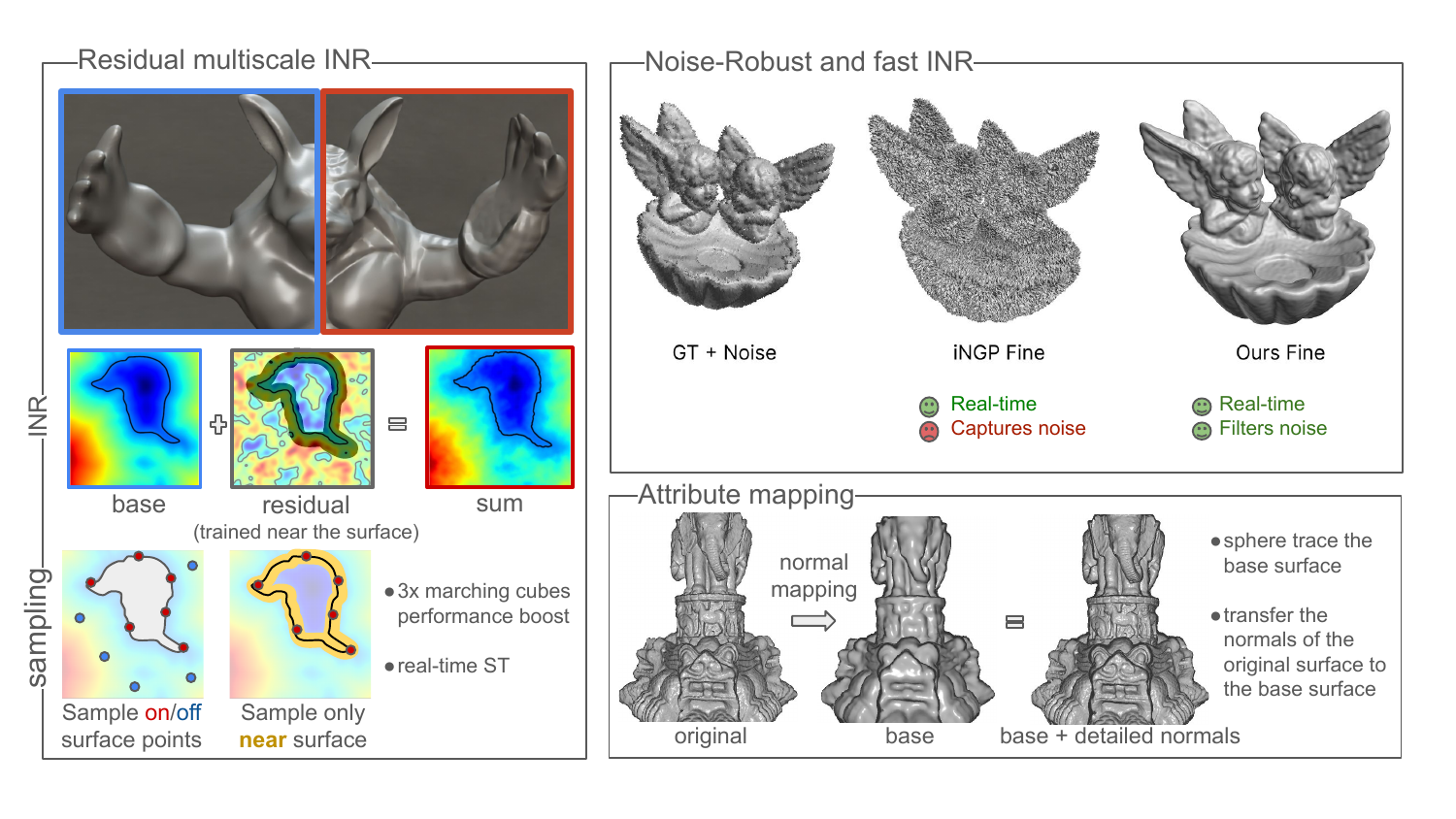}
\caption{We introduce \method{}, a multiscale INR framework for SDFs, based on a nested neighborhood scheme. (Left) Geometry is decomposed into base and residual SIRENs, trained at progressively finer scales with localized, near-surface sampling. (Top-right) \method{} is noise-robust while supporting real-time sphere tracing (ST), unlike grid-based methods such as iNGP that capture noise. (Bottom-right) \method{} supports downstream tasks such as normal mapping.
}
\label{fig:teaser}
\end{figure}

Most INRs rely on a single large MLP modeling the SDF across the whole domain. Scaling up captures finer details but produces large models with prohibitive inference costs, especially for level-set extraction methods—marching cubes~\cite{lorensen1987marching} or sphere tracing~\cite{hart1989ray}—which require dense off-surface evaluations. SIREN~\cite{sitzmann2020implicit} requires full MLP evaluation at every query point, regardless of proximity to the surface; grid-based MLPs~\cite{muller2022instant} enable localized evaluations but limit surface smoothness and overfit input noise; and previous multiscale methods frequently capture noise and introduce artifacts due to hard frequency truncation.

To address these limitations, we introduce \method{} ({\textbf{M}ultiscale Im\textbf{plicit} Neural \textbf{S}urfaces}), which models the SDF as a residual sum of MLPs trained within a sequence of nested neighborhoods around the previous zero-level sets. The coarse network acts as a low-pass filter that establishes a clean, noise-resilient prior, and finer residuals progressively refine geometry without fitting high-frequency artifacts. Surface attributes (normals, textures) are modeled within the same neighborhood structure; for real-time rendering we introduce multiscale sphere tracing and a GEMM-based~\cite{dongarra1990set} analytical normal computation. Fig.~\ref{fig:teaser} illustrates \method{}'s outputs across fitting, rendering, and downstream tasks.
In summary, our contributions~are:

\begin{itemize}
\renewcommand{\labelitemi}{$\bullet$}

\item A compact and efficient multiscale INR model for accurate SDF representation, formulated as a residual sum of MLPs, each capturing a specific frequency band—achieving high representational capacity with fast inference for detailed surface modeling.

\item A nested neighborhood training strategy that refines each residual component by supervising only near the previous zero-level set. This localized approach acts as a natural noise filter, enhancing geometric fidelity and improving data efficiency for oriented point clouds, even under noisy conditions.

\item \method{} enables fast inference through real-time multiscale sphere tracing, a GEMM-based technique for normal computation, and efficient mesh extraction with support for normal and texture mapping, delivering real-time rendering and up to a $5\times$ mesh-extraction speedup over a SIREN baseline.
\end{itemize}

\section{Related Work}\label{s-related_works}Implicit functions are central to graphics and vision~\cite{velho2007implicit, macedo2009hrbf, mescheder2019occupancy}, with SDFs serving as a fundamental tool for modeling and manipulating surfaces~\cite{bloomenthal1990interactive,sang2025implicit}. MLPs have been shown effective as INRs to model SDFs~\cite{park2019deepsdf, gropp2020implicit}, including SIRENs~\cite{sitzmann2020implicit}, which use periodic activations to capture high-frequency details.{\paragraph{Multiscale Neural SDFs.}}
Various methods have investigated multiscale or frequency-aware training to enhance the expressiveness of INRs. BACON~\cite{lindell2021bacon} employs multiplicative filter networks (MFNs)~\cite{fathony2020multiplicative} to band-limit the spectrum of INRs. However, this approach introduces artifacts and is susceptible to capturing input noise due to hard spectral truncation, and the lack of non-linear activations limits the capacity to represent fine details with small networks. ~\cite{dou2023multiplicative} improves the use of MFNs by integrating a feature grid and small architectural changes. BANF~\cite{shabanov2024banf}, MINER~\cite{saragadam2022miner}, and MRNet~\cite{paz2023mr} follow a multiscale design using a Laplacian pyramid. BANF uses grid-based MLPs, which increases memory usage and introduces a dependency on spatial grids.
Also, these methods supervise off-surface regions at all scales, limiting their efficiency and making real-time inference more challenging.
Residual and detail decompositions also appear in the multi-view setting: HF-NeuS~\cite{wang2022hfneus} adapts the IDF displacement composition to a NeuS pipeline, and PR-NeuS~\cite{xu2023prneus} learns a residual on top of a data prior to accelerate convergence. Both supervise the full domain, with no nesting guarantee between levels. A complementary line suppresses spurious off-surface geometry through second-order regularizers loss-level curvature penalties developed largely for unoriented inputs, orthogonal to our representation and combinable with our per-level losses (DiGS~\cite{benshabat2022digs}, StEik~\cite{yang2023steik}, and Neural-Singular-Hessian~\cite{wang2023neuralsingular}).
In contrast, our approach constructs a residual sum of SIRENs, with each component supervised to capture a distinct frequency band improving geometric fidelity and data efficiency by concentrating learning near the surface. Unlike all of these prior methods, \method{} combines a continuous (non-grid) residual representation with strictly localized nested-band supervision and a multiscale tracing / mesh-extraction inference path that exploits this nesting; Tab.~\ref{t-prior-work} contrasts these four axes against the published methods used as baselines in our experiments.

\begin{table}[h]
\scriptsize
\centering
\setlength{\tabcolsep}{4pt}
\vspace{-2mm}
\caption{Differentiation of \method{} from the methods compared in the experiments. \method{} uniquely combines a continuous (non-grid) residual representation with strictly localized nested-band supervision and a multiscale tracing/extraction inference path that exploits this nesting. Citations in surrounding text.}
\label{t-prior-work}
\begin{tabular}{@{}lp{2.3cm}p{2.1cm}p{2.0cm}p{2.4cm}@{}}
\toprule
Method & Representation & Supervision domain & Inference / tracing & Noise mechanism \\
\midrule
BACON & Multiplicative filter network & Full domain, multiscale via architecture & Multiscale MC & Hard spectral truncation \\
BANF & Regular grid + interpolation kernel & Grid sample locations, full domain & Standard MC & Output kernel filter (anti-aliasing) \\
IDF & Coarse base MLP + displacement MLP & Full domain & Mesh extraction only & Coarse-base smoothness; off-surface blobs persist \\
iNGP & Multiresolution hash grid & Full domain & Hash-grid ST & None (grid-based) \\
\textbf{\method{} (ours)} & \textbf{Coarse MLP + multi-level residual MLPs} & \textbf{Nested narrow $\delta_i$-band per residual} & \textbf{Multiscale ST + adaptive MC} & \textbf{Coarse low-pass + nested band excludes off-surface blobs} \\
\bottomrule
\end{tabular}
\vspace{-3mm}
\end{table}
{\paragraph{Inference and Rendering.}}
Traditional visualization of SDFs relies on marching cubes~\cite{lorensen1987marching} or sphere tracing (ST)~\cite{hart1996sphere}. Performance-focused approaches such as~\cite{davies2020overfit} leverage ST to enable real-time rendering of INR level sets. NGLOD, in particular, interpolates hierarchical features stored in a sparse voxel octree and decodes them with shallow MLPs. Although efficient, the discretized octree structure introduces gradient discontinuities across voxel boundaries, which affect normal estimation and surface smoothness. In addition, it is not designed for real-time rendering of detailed level sets.
Instant-NGP~\cite{muller2022instant} can be viewed as an evolution of NGLOD. It replaces the sparse octree with a multiresolution hash-grid encoding, improving memory efficiency and enabling fully GPU-parallelized training and inference. This results in significantly faster optimization and real-time rendering, making Instant-NGP a strong and relevant baseline for comparison. However, its grid-based capacity makes it prone to overfitting, often struggling to recover smooth, coherent surfaces from noisy input point clouds.
In contrast, \method{} adopts a continuous representation by modeling each frequency band explicitly with SIREN components. This avoids discretization artifacts typical of grid-based encodings, provides smooth spatial gradients by construction, and naturally filters noise through its coarse-to-fine nested architecture, improving normal estimation and level-set fidelity.

{\paragraph{Attribute Mapping.}}Finally, classical attribute mapping techniques, such as normal mapping~\cite{cohen1998appearance}, enhance surface detail but require explicit parameterizations and are sensitive to geometric distortions. Recent neural approaches~\cite{wang2022geometry} extend these ideas by propagating learned features off-surface using convolutional modules. Our method simplifies this process by avoiding interpolation entirely: inspired by variational inpainting techniques~\cite{bertalmio2001variational}, we regularize attribute fields to remain smooth along normals near the surface. For texture mapping, traditional parameterized approaches~\cite{catmull1974subdivision} and neural texture fields~\cite{oechsle2019texture, gao2022get3d} require known meshes or image-depth pairs and often involve complex training. In contrast, we define texture fields directly from colored point clouds using compact MLPs, regularized along the surface, enabling fast, texture-aware rendering without dense supervision or UV mapping.

\section{M-plicits: \textbf{M}ultiscale Im\textbf{plicit} Neural \textbf{S}urfaces}
\begin{figure}[h]
\centering
    \includegraphics[width=0.8\textwidth]{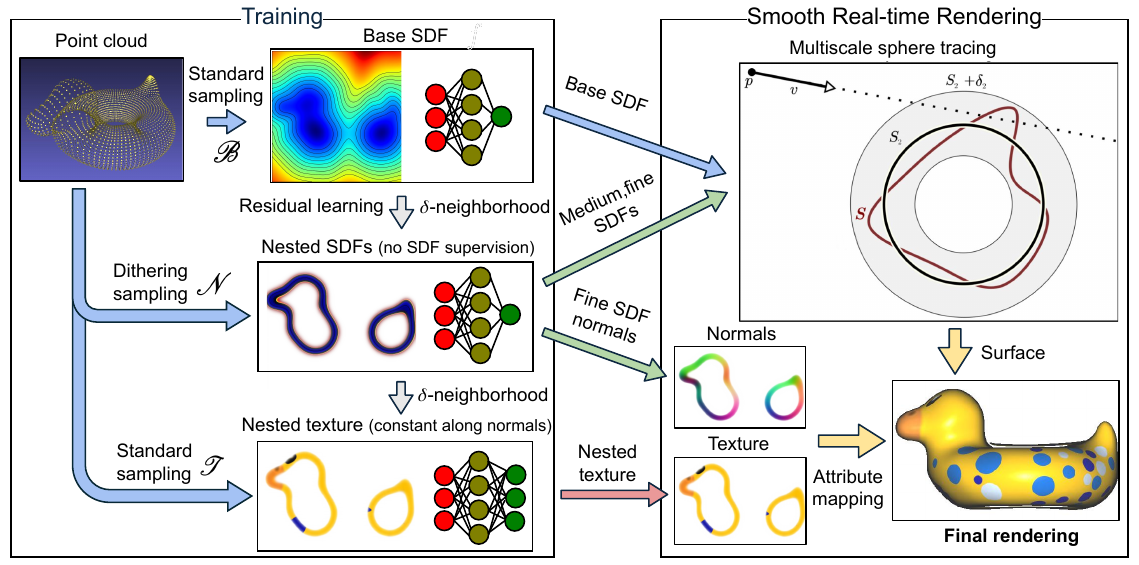}
    \caption{Overview of \method{}. Starting from an oriented point cloud with colors, we combine sampling techniques and loss regularizations ($\mathscr{B}$ and $\mathscr{N}$) to create a base, a medium %
    , and a fine SDF to implicitly represent the SDF in multiscale. The base SDF is defined for the entire domain, while the others are residuals, defined in (nested) neighborhoods of the surface. The colors are also trained in a neighborhood, regularized (by $\mathscr{T}$) to be constant along normals. The resulting multiscale representation can be rendered using novel sphere tracing and attribute mapping algorithms.}
    \label{f-pipeline}
\end{figure}

Our goal is to model SDFs using a multiscale INR based on a residual sum of SIRENs, where each component captures a distinct level of detail. We train these networks using a nested neighborhood strategy: each residual is supervised only near the current zero-level set, concentrating learning near the surface. This improves geometric details and enables fast inference. Additionally, we introduce an attribute mapping scheme that leverages the SDF structure to support textures and normals without relying on mesh parameterizations or interpolation.
Figure~\ref{f-pipeline} gives an overview of our method.
\vspace{-9mm}

\subsection{Preliminaries}

Given an oriented point cloud $\{x_j, N_j\}_{j=1}^n$, consisting of surface points $x_j$ and their normals $N_j$, our goal is to reconstruct the underlying surface \(S\) as the zero level set of a signed distance function (SDF) \(f : \mathbb{R}^3 \to \mathbb{R}\), i.e.,
$
S = f^{-1}(0) = \{x \mid f(x) = 0\},
$
with the additional condition that \(f(x_j) \approx 0\) and \(\nabla f(x_j) \approx N_j\).
To regularize the solution away from the input data, we also recall that true SDFs satisfy the Eikonal equation:
$\norm{\nabla f(x)} = 1$ for all $x \in \Omega$ in the training domain $\Omega$.
Enforcing this condition during training improves generalization in unsupervised regions.
Finally, combining the data constraints with the Eikonal regularization leads to a loss function~\cite{gropp2020implicit}:
\begin{equation}
\label{e-loss_function-siren}
\mathcal{L}(f)=
\frac{1}{n}\sum_{j}
\left[
f(x_j)^2
+ \left(1 - \nabla f(x_j) \cdot N_j \right)
\right]
+ \int_{\Omega} \left(\|\nabla f(x)\| - 1\right)^2 \, dx .
\end{equation}
The first two terms ensure that the network fits the input points and aligns the gradient with the ground-truth normals. The third term enforces the Eikonal constraint, providing geometric regularization over \(\Omega\).
To capture fine geometric detail, it is common to parameterize the SDF \(f\) using a sinusoidal MLP (SIREN). A SIREN with \(n{-}1\) hidden layers is defined as:
\begin{equation}
\label{e-siren}
f(x) = W_n \circ h_{n-1} \circ \cdots \circ h_0(x) + b_n,
\quad
h_i(x) = \sin\!\left( \omega_0 (W_i x + b_i) \right).
\end{equation}
Here, \(\omega_0\) is a frequency parameter controlling the network’s capacity to model high-frequency details~\cite{sitzmann2020implicit}, and each \(W_i, b_i\) are the learnable weight matrices and biases.
This architecture enables the network to fit both surface constraints \(f(x_j) \approx 0\) and normal alignment \(\nabla f(x_j) \approx N_j\), while providing high representational capacity for complex geometries. However, relying on a single large MLP to represent the entire domain is inefficient, especially for high-resolution surfaces, as it leads to high computational cost during inference.

\subsection{Multiscale modeling of neural SDFs}
Standard neural SDFs typically rely on large MLPs to capture fine surface details, leading to fitting inefficiencies and high inference costs. To address this, we propose a multiscale SDF representation by recursively refining a coarse base network. Specifically, we define a sequence of neural SDFs \(\{f_i\}_{i=1}^n\) via a sum of residual MLPs:
    $f_{i+1} = f_i + r_i,$ for $i = 1,2,$
where \(f_1\) is a compact base MLP trained over the full domain, and each residual \(r_i\) refines \(f_i\) by capturing higher-frequency components, modulated by the SIREN frequency parameter~\(\omega_0\).
We refer to the sequence \(\{f_i\}\) as multiscale SDFs, each approximating the SDF of \(S\) at an increasing level of detail.
While this formulation naturally extends to an arbitrary number of residuals, we focus on the case of three MLPs for simplicity and because this configuration performs well in our experiments.

To enable efficient learning and real-time inference, we impose a nesting condition between successive SDF levels. Specifically, each refined surface \( S_{i} = f_{i}^{-1}(0) \), for $i=2,3$ is constrained to lie within a narrow band around the previous level set \( S_{i-1} \), that is,
\begin{align}\label{e-nesting_condition}
    S_3\subset \big[|f_{2}|<\delta_2\big]\subset \big[|f_1|<\delta_1\big]
\end{align}
Throughout, we use the bracket shorthand $\big[|f_i|<\delta_i\big]$ to denote the $\delta_i$-band $\{x\in\Omega : |f_i(x)|<\delta_i\}$ around $S_i$. The inclusions in (\ref{e-nesting_condition}) hold by construction (up to marching-cubes interpolation at extraction): each residual $r_{i+1}$ is supervised only on this set, and $\delta_i$ is set adaptively (Eq.~\ref{e-deltas}) so that all input points lie within it.
This is enforced during training by supervising the residual MLP \( r_i \) only within this \(\delta_i\)-neighborhood, a strategy we call nested neighborhood training. This condition ensures that each residual captures localized corrections, promotes coarse-to-fine refinement, and supports efficient applications such as progressive sphere tracing and surface-aware attribute mapping (see supplementary material). The choice of \(\delta_i\) is tied to the training regime and is discussed in next.

\paragraph{Training with nested neighborhoods.}
\label{s-examples}
In standard SIREN-based SDF fitting, the loss in \eqref{e-loss_function-siren} is applied across the entire domain \(\Omega\). In contrast, our multiscale framework leverages the nesting condition to localize supervision. The hierarchy begins with a coarse-level SDF \(f_1\), modeled by a compact SIREN. Finer levels \(f_2, f_3, \dots\) are added as residual SIRENs with progressively higher-frequency capacity, controlled by increasing values of the sinusoidal parameter \(\omega_0\).

Training proceeds in stages: first, the base SDF \(f_1\) is trained over the full domain \(\Omega_1:=\Omega\); then, each subsequent level \(f_{i+1}\) (for \(i=1,2\)) is trained within the restricted band \(\Omega_{i+1}:=\{x\in\Omega_i : |f_i(x)| < \delta_i\}\), so that each field \(f_i\) is trained on its own domain \(\Omega_i\).
Each SDF \(f_i\) is optimized using a combination of data and Eikonal losses:
\begin{equation}
\label{e-loss_function}
\mathcal{L}_i(f_i)=
\frac{1}{n}\sum_{j}
\left[
f_i(x_j)^2
+ \left(1 - \nabla f_i(x_j) \cdot N_j \right)
\right]
+ \int_{\Omega_{i}} \left(\|\nabla f_i(x)\| - 1\right)^2 \, dx .
\end{equation}
To ensure that \(\{x_j\}\) lies within $\Omega_{i}$ at each stage $i$, we set the band threshold \(\delta_i\) adaptively following:
\begin{align}\label{e-deltas}
    \delta_i = (1 + \varepsilon) \cdot \max_{j=1,\dots,n} |f_i(x_j)|, \text{ with } \varepsilon > 0\text{ a small threshold.}
\end{align}
The maximum runs over the input surface points \(\{x_j\}\): \(\max_j |f_i(x_j)|\) is level \(i\)'s residual fitting error at the data (small by construction) not the field's maximum over the domain. Measured band widths are reported in suppl.~Sec.~\ref{s-deltas}.

\begin{figure}[!h]
\centering
    \includegraphics[width=0.7\textwidth]{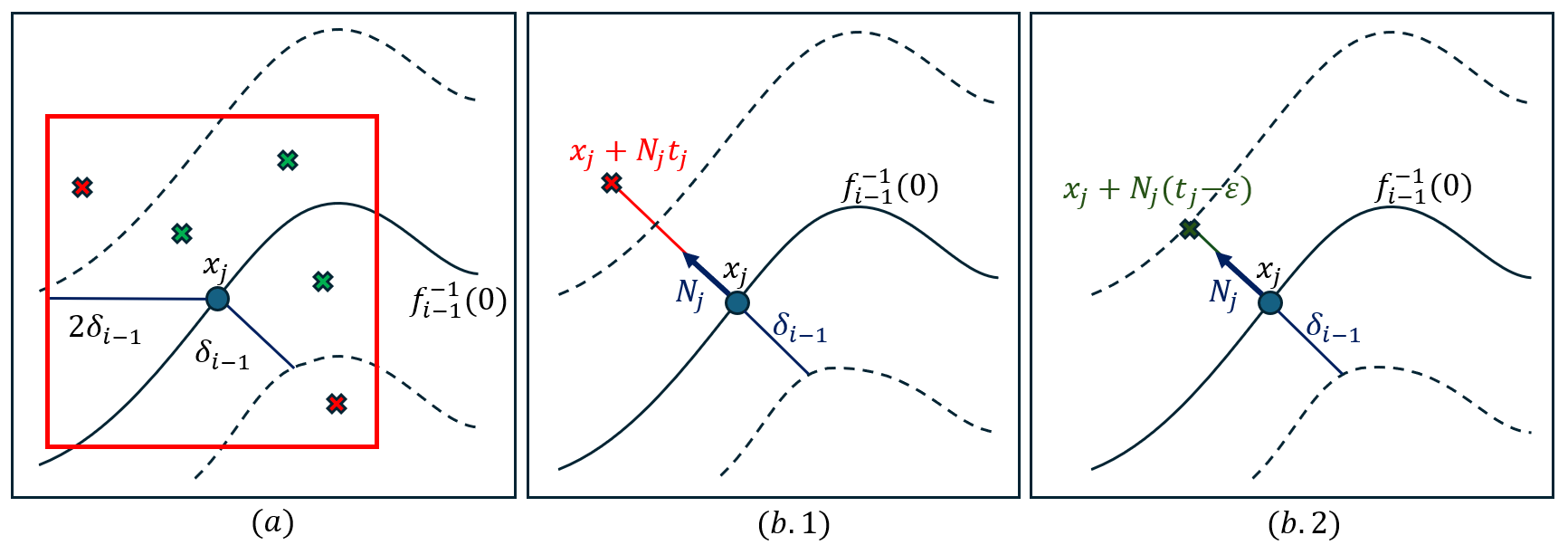}
    \caption{\textbf{(a)} Dithering-based sampling around each input point \(x_j\), where each coordinate is perturbed by a random value in \((-2\delta_{i-1}, 2\delta_{i-1})\), followed by filtering to retain only points in the valid band $\Omega_{i}$ (green points denote those that are kept and red those rejected).
\textbf{(b.1)} Computation of the displacement vector \(t_j N_j\), where the offset point \(x_j + t_j N_j\) lies just outside the narrow band \(\Omega_{i}\).
\textbf{(b.2)} Final accepted sample \(x_j + N_j(t_j - \varepsilon)\), located inside \(\Omega_{i}\), used to supervise the data term.
}
    \label{f-sampling}
\end{figure}

\paragraph{Sampling.}
To discretize the Eikonal term over $\Omega_{i}$,
we employ dithering-based sampling around the input points $\{x_j\}$,
perturbing each coordinate by a random value in the interval
$(-2\delta_{i-1},\, 2\delta_{i-1})$.
Samples falling outside $\Omega_{i}$ are rejected using the condition
$|f_{i-1}(x)| < \delta_{i-1}$; see Fig.~\ref{f-sampling}(a).

Adding extra samples to accelerate SDF training is a common strategy~\cite{novello21diff}, but it often requires evaluating a large MLP over the entire domain, leading to expensive inference. In contrast, \method{} uses residual MLPs that are trained only within a narrow band around the input data, allowing for efficient sampling.
To enrich supervision of the data term in \eqref{e-loss_function}, we propose sampling along the normal \(N_j\) of each point \(x_j\); see Fig.~\ref{f-sampling}(b). Specifically, we compute a scalar \(t_j \leq \delta_{i-1}\) such that for all \(t \in [0, t_j]\), the offset point \(x_j + t N_j\) lies at a distance \(t\) from the surface. This allows us to supervise both the SDF value and its gradient:
$
f_i(x_j + t N_j) = t$ \text{and} $\nabla f_i(x_j + t N_j) = N_j,
$
providing richer sampling near the surface.
We determine each \(t_j\) via an iterative scheme: starting from \(t_j = \delta_{i-1}\), we reduce it by a small step until the distance from \(x_j + t_j N_j\) to the point cloud \(\{x_j\}\) equals \(t_j\), ensuring that the offset point lies inside the band.

\paragraph{Rendering and mesh extraction.}

To render the zero level set \(f^{-1}(0)\) of an SDF \(f\), it is common to use either sphere tracing (ST) or {marching cubes} for mesh extraction followed by standard mesh rendering. When the SDF is represented using \method{}, both strategies become more efficient.

\begin{figure}[!h]
    \centering
    \small
    \includegraphics[width=0.7\textwidth]{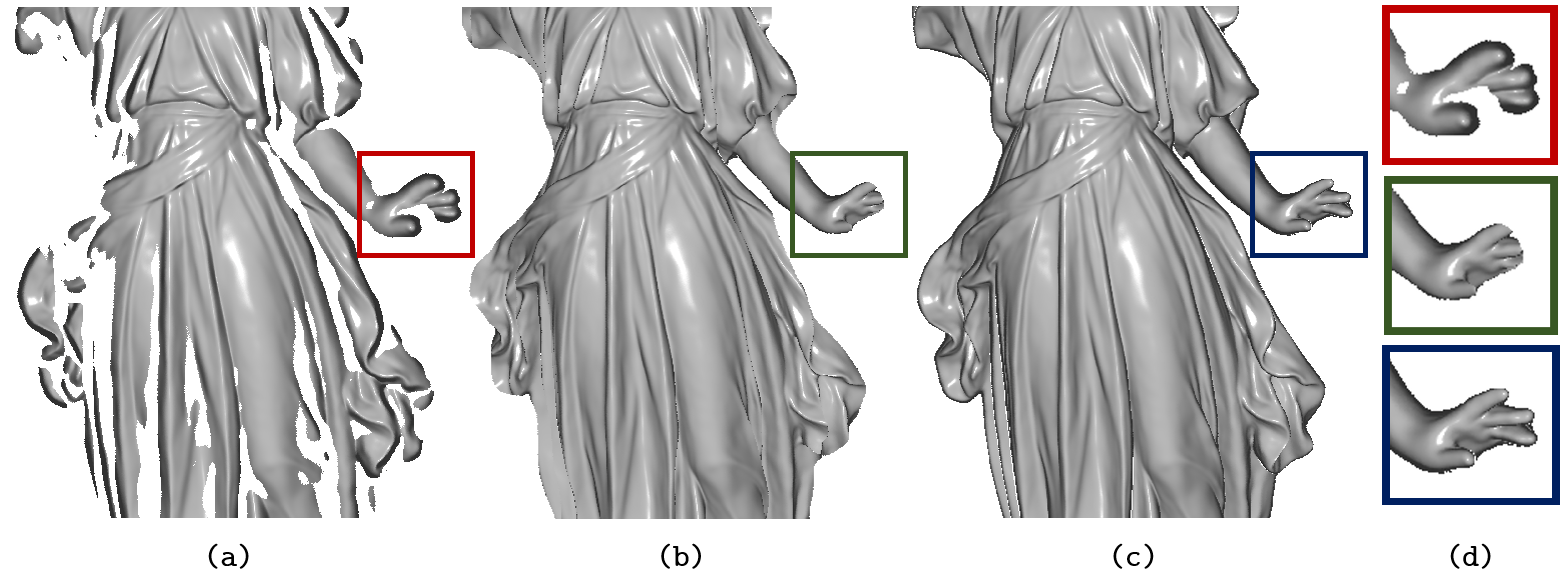}
    \caption{
    (a) Using too large $\delta_1=\delta_2$ may result in holes (the ray does not reach the surface). More iterations would be needed using the finer SDF to fill those holes, defeating the idea of minimizing iterations.
    (b) Conversely, reducing the deltas $\delta_1=\delta_2$ may miss parts of the silhouette since the target surface may not be inside the previous neighborhood (the hand).
    (c) Using $\delta_1$ and $\delta_2$ suited for the nesting condition implies no holes and better silhouette capture.
    }
    \label{f-delta_issues}
    \vspace{-0.3cm}
\end{figure}

We first introduce a \textbf{multiscale ST}. Given a view ray $\gamma(t)\!=\!p_0+tv$, with origin at a point \(p_0\) and direction \(v\), intersecting $f^{-1}(0)$, standard ST approximates the first intersection point by iterating \(p_{i+1} \!=\! p_i \!+\! v f(p_i)\) along $\gamma$.
However, querying a large MLP at each step may be expensive. To reduce this cost, we exploit the multiscale SDF hierarchy \(\{f_i\}\), using coarser networks to guide early steps.
Thanks to the nesting condition in \ref{e-nesting_condition}, coarse levels can be used to trace offset surfaces before switching to finer levels near the surface.
The ray starts tracing \(f_1^{-1}(\delta_1)\) with \(f_1\), then proceeds to \(f_2^{-1}(\delta_2)\) using \(f_2\), and finally reaches the target surface \(f_3^{-1}(0)\) with \(f_3\). Each coarser level uses offset tracing via
$p_{i+1} \!=\! p_i \!+\! v \left(f_j(p_i) \!-\! \delta_j\right)$,
ensuring convergence avoiding high-cost evaluations.
The values \(\delta_i\) (\ref{e-deltas}) play a crucial role in rendering. Using distinct values at each level helps prevent issues such as missed ray-surface intersections: if \(\delta_1\) is too small, parts of \(f_2^{-1}(0)\) might lie outside the region bounded by \(f_1^{-1}(\delta_1)\); Fig.~\ref{f-delta_issues} shows some of these~issues.

\begin{wrapfigure}[12]{R}{0.33\textwidth} %
        \centering
        \vspace{-0.7cm}
        \includegraphics[scale=0.7]{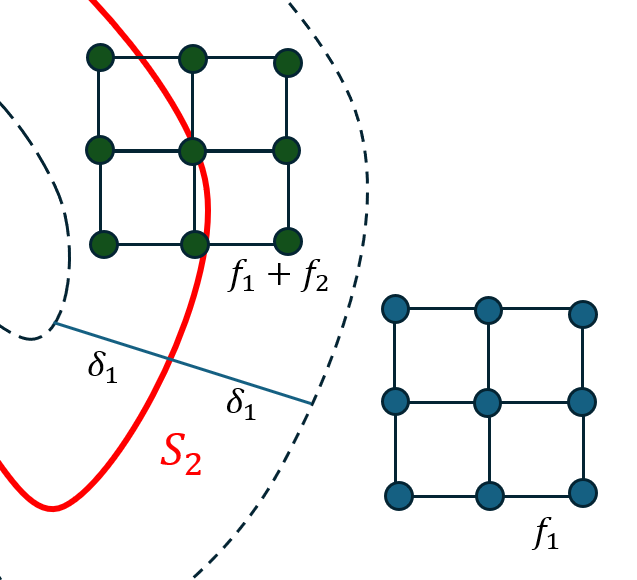}
        \caption{Adaptive marching cubes. For grid vertices outside the $\delta_1$ neighborhood (blue), only the coarse SDF $f_1$ is evaluated. For points in the neighborhood (green) the residual $f_2$ is added. }
        \label{f-marching_cubes}
\end{wrapfigure}%
\method{} also accelerates mesh extraction using \textbf{marching cubes}.
We adopt an adaptive grid inference strategy by first evaluating the coarse SDF \(f_1\) to cull grid vertices,
and querying finer SDFs only for vertices inside the \(\delta_1\)-neighborhood.
This reduces the number of voxel evaluations, thereby accelerating mesh extraction (Fig.~\ref{f-marching_cubes}).

\paragraph{{Normal and texture mapping.}}
Let \(S\) be a surface nested within a \(\delta\)-neighborhood of the zero-level set of a neural SDF \(f\).
Assuming \(f\) is a finer-level neural SDF, we define the \textbf{neural normal mapping} by assigning to each point \(p \in S\) the attribute
$
g(p) := \nabla f(p).
$
If \(S\) corresponds to the zero-level set of a coarser neural SDF, this mapping allows us to bypass additional sphere tracing iterations, reducing computational overhead.

Similarly, we define a network \(g\colon \mathbb{R}^3 \to \mathcal{C}\) to encode a \textbf{texture} within the \(\delta\)-neighborhood of \(f\), where \(\mathcal{C}\) is the \texttt{RGB} color space.
We refer to the attribute mapping defined by the triple \(\{S, f, g\}\) as a \textbf{neural texture mapping}.
To train the parameters \(\phi\) of \(g\), we optimize
$
\mathscr{T}(\phi) = \int_{f^{-1}(0)} (g - \mathscr{g})^2 \, dx + \int_{[|f| \leq \delta]} \langle \nabla g, \nabla f \rangle^2 \, dx,
$
where the first term ensures that \(g\) fits the ground-truth texture \(\mathscr{g}\), while the second term regularizes \(g\) to remain constant along the gradient flow of \(f\), effectively propagating texture information throughout the \(\delta\)-neighborhood.

\paragraph{GEMM-based normal calculation.} To ensure real-time rendering performance, we compute normals without the need of auto-differentiation nor computational graphs. It works as a forward pass on the MLP and is implemented on the GPU using only a GEMM library, resulting in a 2X performance improvement over \texttt{torch.autograd}. Details are in the supplementary material.

\vspace{-0.2cm}
\section{Experiments}
\vspace{-0.2cm}

This section presents experiments to evaluate the proposed method comprehensively, both in comparison to representative multiscale and grid-based neural SDF methods and with ablations designed to understand the importance of its different components.
We also demonstrate several applications of our method, showcasing its versatility.

\paragraph{Implementation details.}
All experiments are conducted on an NVIDIA RTX 5090 with 32 GB.
For sphere tracing, we fix the number of iterations to 20 for coarse and 5 for each residual level, for better control of the parallelism. The $\omega_0$ SIREN parameters are 30 for coarse, 45 for medium, and 100 for fine. We use PyTorch's Adam optimizer for training~\cite{paszke2019pytorch}.

\paragraph{Evaluation protocol.}
All input point clouds were centered and normalized to the unit sphere. For evaluation, we extracted the zero-level set of each Neural SDF using marching cubes at a resolution of $512^3$.
We then uniformly sampled 500K points on the reconstructed meshes and computed the L2 Chamfer Distance (CD), and the Intersection over Union (IoU) against the input point cloud using PyTorch3D’s implementation~\cite{ravi2020pytorch3d}.

\paragraph{MLP notation.}
$(N,d)$ refers to a MLP with $d$ hidden layers of the form $\R^{N}\!\!\to\! \R^{N}$.
Additionally, ($128, 2) \!\rhd (256, 2) \!\rhd \!(400, 2)$ refers to a multiscale SDF with coarse, medium, and fine MLPs with two $\R^{128}\!\!\to\! \R^{128}$, $\R^{256}\!\!\to\! \R^{256}$, and $\R^{400}\!\!\to\!\R^{400}$ hidden layers, respectively. This is our setup for all experiments unless otherwise stated.

\begin{table}[t]
\centering
\scriptsize
\vspace{-0.2cm}
\caption{Surface reconstruction on Stanford + Thingi32 datasets.
Best values are \textbf{bold}, second best are \underline{underlined}, and third best are \textit{italic}.
Chamfer Distance (CD) considers 500K samples. Sampling time is measured on a $512^3$ grid; for our method this uses the adaptive multiscale marching cubes (Sec.~3). Renderer FPS considers $512^2$ images and uses multiscale sphere tracing on geometry only (without normal mapping unless explicitly noted in Tab.~\ref{t-3d}).}
\resizebox{\columnwidth}{!}{
\begin{tabular}{lcccccccc}
\toprule
\textbf{Method}
& mean CD $\downarrow$
& median CD $\downarrow$
& mean IoU $\uparrow$
& median IoU $\uparrow$
& \# params $\downarrow$
& avg Training (min) $\downarrow$
& avg Sampling (s) $\downarrow$
& Renderer FPS $\uparrow$ \\
\midrule

iNGP coarse
& \textit{6.87E-05} & 6.43E-05 & 4.12E-01 & 4.05E-01
& 2,040,864
& \underline{8.22}
& \underline{0.40}
& \textit{86} \\

iNGP fine
& \underline{6.44E-05} & \textit{6.04E-05}  & \textit{4.20E-01} & \textit{3.90E-01}
& 9,113,760
& 10.59
& \textit{0.89}
& \underline{88} \\

IDF
& 6.13E-02 & 7.74E-04 & 4.09E-01 &  9.20E-02
& 1,191,943
& \textit{8.26}
&  42.38
& N/A \\

BACON
& 8.12E-03 & \underline{2.18E-05} & \underline{5.15E-01} & \underline{7.32E-01}
& \textit{530,953}
& 8.43
& \textbf{0.24}
& N/A \\

Ours coarse
& \textbf{5.36E-05} & 3.80E-05 & 4.48E-01 & 4.59E-01
& \textbf{17,153}
& \textbf{4.07}
& 1.51
& \textbf{180} \\

Ours fine
& 6.57E-05 & \textbf{1.87E-05} & \textbf{5.86E-01} & \textbf{8.67E-01}
& \underline{246,627}
& 8.53
& 5.29
& 43 \\

\bottomrule
\end{tabular}}

\vspace{-0.6cm}
\label{t-comparison}
\end{table}

\subsection{Main results}

\paragraph{Surface reconstruction.} We compare our neighborhood nesting approach against representative multiscale and grid-based neural SDF methods for surface representation on the Stanford and Thingi32 datasets~\cite{curless1996volumetric,zhou2016thingi10k,takikawa21nglod}. We include Instant-NGP (iNGP)~\cite{muller2022instant}, focusing on real-time rendering performance, Implicit Displacement Fields (IDF)~\cite{wang2022geometry}, which disentangles shape and detail, and BACON~\cite{lindell2021bacon}, a multiscale representation where we report results from its 8th hidden layer.
We do not include NGLOD~\cite{takikawa21nglod} as a separate baseline because Instant-NGP supersedes it on this task: iNGP replaces NGLOD's sparse-voxel octree with a multiresolution hash-grid encoding that improves memory efficiency, training speed, and rendering quality, so the iNGP comparison already covers the strongest representative of this lineage. For BANF~\cite{shabanov2024banf}, the public implementation does not provide a pipeline for surface reconstruction from point clouds (the released code targets image fitting, NeRFing, and multi-view 3D reconstruction); we therefore report a comparison against our own faithful re-implementation separately, in suppl.~Sec.~\ref{s-banf}, with its full configuration disclosed. Its numbers reflect a custom implementation rather than the authors' code. We additionally compare against classical Screened Poisson reconstruction~\cite{kazhdan2013screened} in suppl.~Sec.~\ref{s-spsr}.

Tab.~\ref{t-comparison} summarizes the results. We evaluate geometric fidelity using Chamfer Distance (CD), volumetric consistency using Intersection over Union (IoU), computational cost, and rendering throughput. While CD measures pointwise surface deviation, IoU reflects global occupancy agreement and structural correctness.

\begin{figure}[h]
\centering
\scriptsize
\begin{minipage}{0.32\textwidth}
\centering
\includegraphics[width=\linewidth]{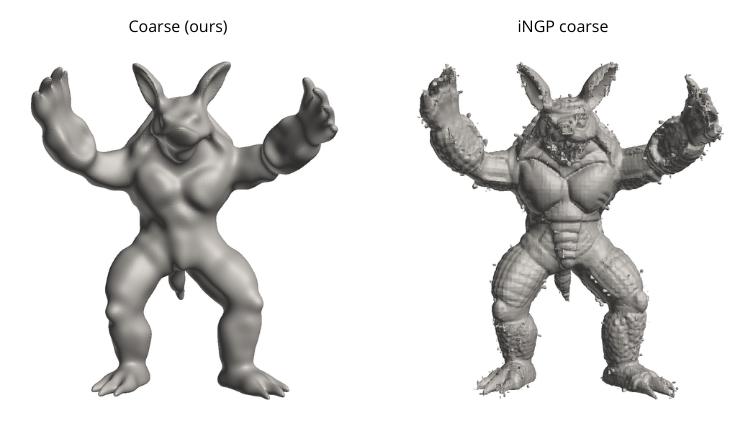}
\caption{Coarse reconstruction of the Armadillo (Stanford). Ours vs.\ iNGP.}
\label{coarse}
\end{minipage}
\hfill
\begin{minipage}{0.66\textwidth}
\centering
\includegraphics[width=\linewidth]{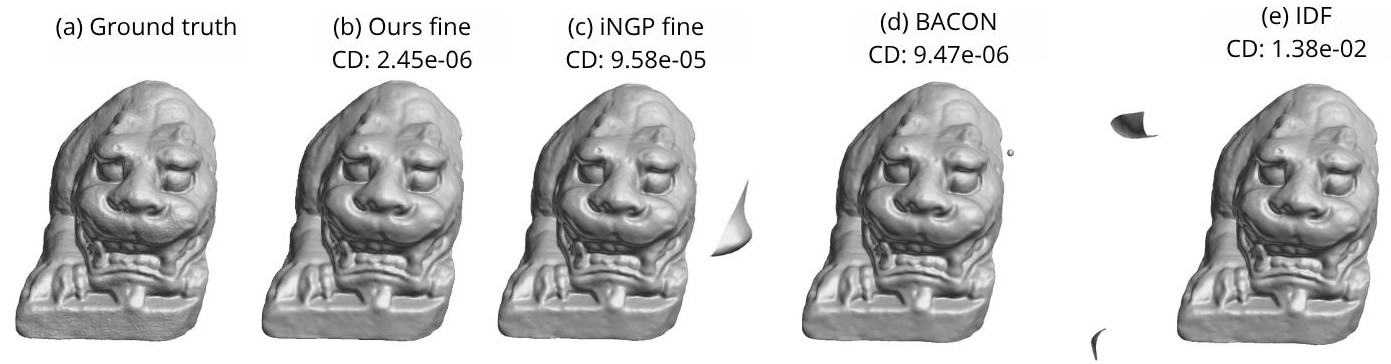}
\caption{Qualitative comparison on a Thingi32 sample. (a) GT; (b) ours (fine); (c) iNGP fine; (d) BACON, showing blob-like artifacts; (e) IDF. All with their respectives CD values.}
\label{all_fines}
\end{minipage}
\end{figure}

In the coarse configuration, our method achieves the best mean CD while also improving IoU over both iNGP variants, indicating better global geometric alignment rather than merely local surface fitting (Fig.~\ref{coarse}). In the fine configuration, mean CD remains comparable to iNGP fine, while our method achieves the best median CD and significantly higher IoU (Fig.~\ref{all_fines}). Most reconstructions closely match the ground truth, with only a few challenging Thingi32 cases (non-watertight surfaces, multiple components, thin structures, topological defects; see suppl.~Fig.~\ref{sailor}) affecting the mean.

Qualitatively, iNGP recovers slightly sharper high-frequency details but often introduces a geometric offset relative to the ground truth (suppl.~Fig.~\ref{comparison_ingp}). We acknowledge this trade-off explicitly: the smoothness prior in the coarse SDF favors faithful global geometry over the highest-frequency micro-detail, and our neural normal mapping (Sec.~\ref{s-ablation}, Fig.~\ref{f-armadillo_ov}) partially mitigates it by transferring fine-frequency normals from a higher-capacity SDF onto the coarse geometry without rendering-cost penalty. IDF frequently produces unstable surfaces with blob-like artifacts (suppl.~Fig.~\ref{comparison_idf}) and requires explicit mesh extraction, preventing direct real-time rendering.

\begin{figure}[h]
\centering
\begin{minipage}{0.58\textwidth}
\centering
\includegraphics[width=\linewidth]{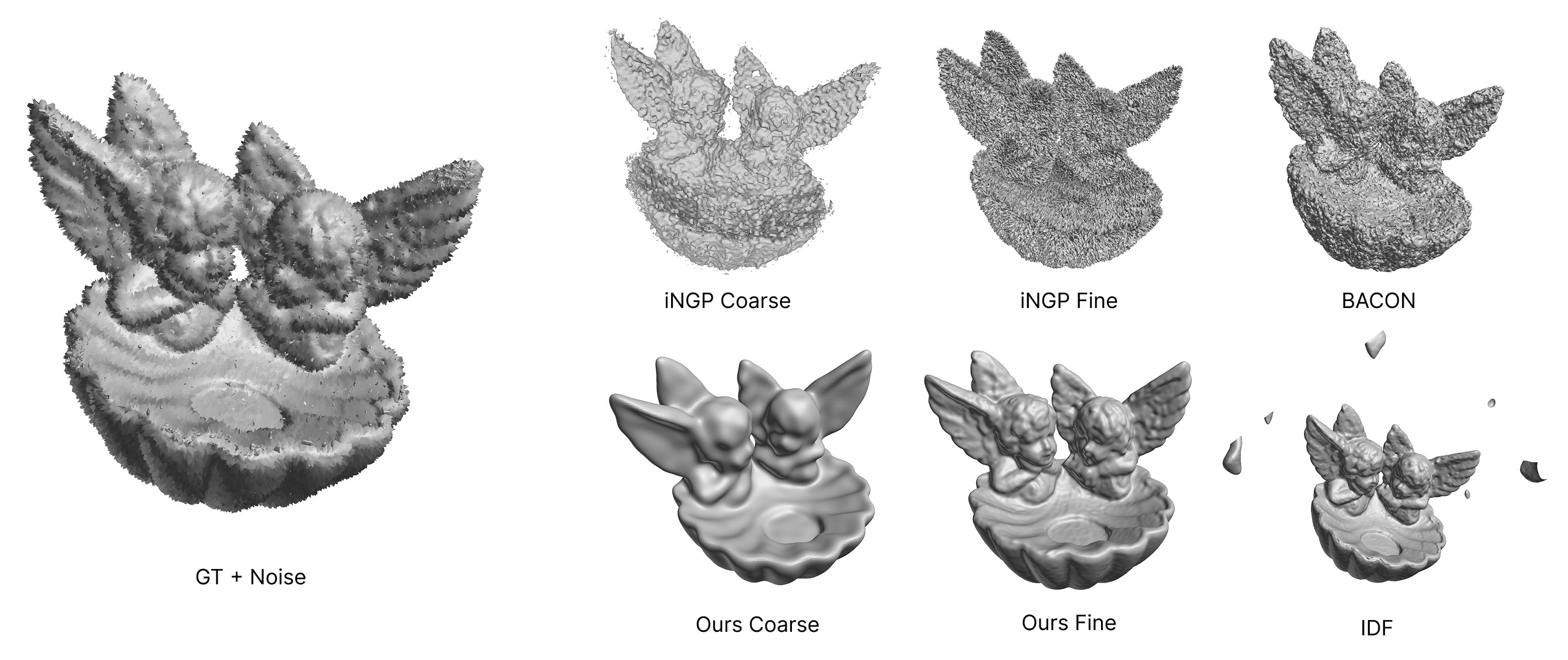}
\captionof{figure}{Qualitative comparison of surface reconstruction on sample 354371 with 1\% noise (Thingi32). Left: GT with noise. Right: reconstructions. IDF is the only baseline that does not retain noise on the surface, but it generates blobs around it.}
\label{fig_noise_comparision}
\end{minipage}
\hfill
\begin{minipage}{0.40\textwidth}
\centering
\tiny
\setlength{\tabcolsep}{3pt}
\captionof{table}{Quantitative robustness to noise. CD ($\downarrow$) and IoU ($\uparrow$) for Stanford and Thingi32 under vertex noise. \textbf{Bold}, \emph{italic}, \underline{underlined} = best, second, third.}
\label{table:noise_comparison}
\begin{tabular}{lcccc}
\toprule
Method & mean CD$\downarrow$ & median CD$\downarrow$ & mean IoU$\uparrow$ & median IoU$\uparrow$ \\
\midrule
Ours coarse & \underline{7.42E-05} & 5.33E-05 & \underline{3.81E-01} & \underline{4.03E-01} \\
Ours medium & \emph{4.62E-05} & \underline{3.03E-05} & \emph{5.37E-01} & \textbf{6.38E-01} \\
Ours fine & \textbf{3.36E-05} & \textbf{1.62E-05} & \textbf{5.82E-01} & \emph{5.91E-01} \\
iNGP coarse & 6.70E-04 & 3.76E-04 & 1.90E-01 & 1.88E-01 \\
iNGP fine & 2.80E-04 & 1.09E-04 & 3.09E-01 & 3.05E-01 \\
IDF & 8.37E-02 & \emph{2.49E-05} & 3.22E-01 & 3.08E-01 \\
BACON & 1.68E-03 & 1.51E-04 & 1.76E-01 & 1.65E-01 \\
\bottomrule
\end{tabular}
\end{minipage}
\end{figure}

\paragraph{Trade-offs and design space.}
No single method in Tab.~\ref{t-comparison} dominates all axes. iNGP fine attains the second-best mean CD and real-time FPS but is highly noise-sensitive (Tab.~\ref{table:noise_comparison}: mean CD $2.80\!\times\!10^{-4}$ vs.\ ours $3.36\!\times\!10^{-5}$). BACON has the fastest sampling and competitive median CD/IoU but is also noise-sensitive (mean CD $1.68\!\times\!10^{-3}$) and provides no real-time rendering. \method{} is the only method in the (noise-robust, real-time) region, achieving the best mean CD in coarse, the best median CD/IoU in fine, and the best noise robustness across configurations, while remaining roughly an order of magnitude smaller in parameter count than grid-based baselines and supporting native real-time rendering (180 FPS coarse, 35--43 FPS fine, at $512^2$; Tabs.~\ref{t-comparison} and~\ref{t-3d}). Per-mesh CD distributions in suppl.~Figs.~\ref{fig:chamfer_ours+ignp_coarse}, \ref{fig:chamfer_ours+ignp_fine}, \ref{fig:chamfer_bacon} and~\ref{fig:chamfer_idf} are tightly concentrated around the mean for \method{} and substantially wider for BACON and IDF, indicating more stable reconstruction across shapes.

\vspace{-3mm}
\vspace{-3mm}

\paragraph{Robustness to Noise.}
We evaluate noise robustness on a perturbed version of Stanford and Thingi32 in which vertices are randomly displaced along the ground-truth normal by up to 1.0\% of the bounding box. The coarse level of \method{} acts as a low-pass filter, removing high-frequency noise and providing a clean geometric prior for subsequent residual levels (Fig.~\ref{fig_noise_comparision}). In contrast, iNGP and BACON absorb noise into the reconstructed surface; IDF avoids surface noise but produces unstable blobs around it, lowering its mean CD (Tab.~\ref{table:noise_comparison}).

\label{s-comparisons}

\subsection{Ablations and additional experiments}
\label{s-ablation}
\begin{figure}[h!]
\centering
\includegraphics[width=0.75\textwidth]{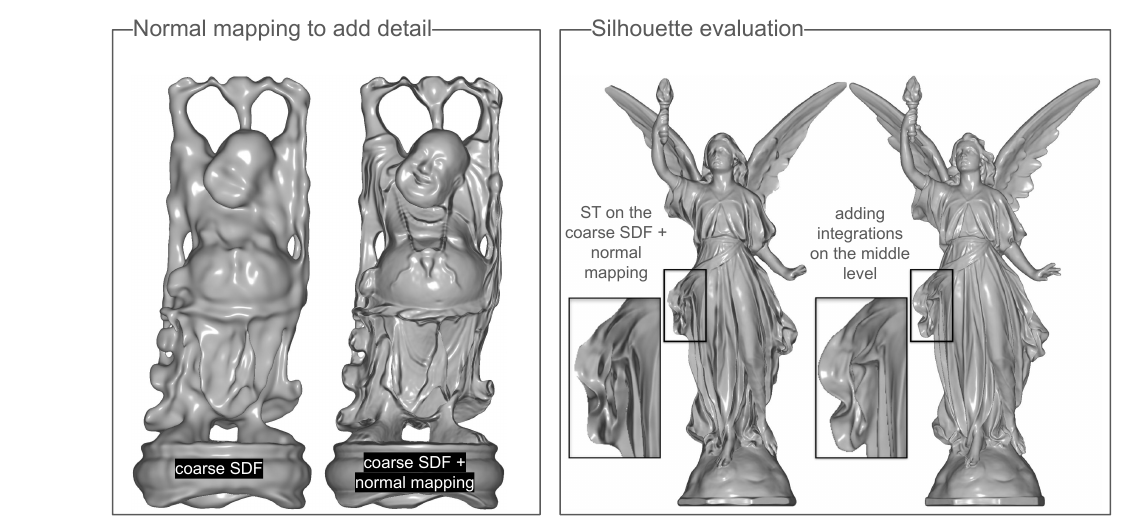}
\caption{Left: neural normal mapping from a finer SDF adds high-frequency detail to a coarse $(64,1)$ SDF (Buddha). Right: silhouette evaluation (Lucy); compared to sphere tracing the coarse SDF alone with normal mapping, adding ST iterations on a medium $(256,2)$ level refines the silhouette.}
\label{f-armadillo_ov}
\end{figure}

\paragraph{Isolating residuals and nested bands.}
Suppl.~Sec.~\ref{s-isolation} adds an ablation isolating our two design axes (the residual composition and the nested-band supervision) against variants trained with full-domain residuals and with independent per-level SDFs, together with robustness studies of coarse-stage under-training (suppl.~Sec.~\ref{s-undertrained}) and outlier contamination (suppl.~Sec.~\ref{s-outliers}).
\paragraph{Neural normal mapping and multiscale ST.}
Fig.~\ref{f-armadillo_ov} shows that neural normal mapping adds high-frequency detail to a coarse SDF without rendering-cost penalty (left), and that adding multiscale ST iterations on a medium-level SDF further refines the silhouette (right). A broader evaluation across other models is in the appendix.

\begin{table}[!ht]
\centering
\scriptsize
\renewcommand{\arraystretch}{0.9}
\begin{minipage}[t]{0.55\textwidth}
\centering
\tiny
\captionof{table}{Real-time CUDA renderer ablation (multiscale ST, GEMM normals, neural normal mapping). 20 ST iterations on the first SDF, 5 on each subsequent level. (NM) denotes normal mapping without additional ST on the last SDF. Images $512^2$; size in KB.}
\label{t-3d}
\begin{tabular}{lrrr}
\toprule
Model & FPS & Speedup & Size \\
\midrule
$(400,4)$ (SIREN baseline)                  & 19  & 1.0$\boldsymbol{\times}$ & 1899 \\
$(128, 2)$ (coarse)                         & 180 & 9.5$\boldsymbol{\times}$ & 68   \\
$(128, 2) \rhd (256, 2)$ (NM)               & 85  & 4.5$\boldsymbol{\times}$ & 331  \\
$(128, 2) \rhd (256, 2)$                    & 70  & 3.7$\boldsymbol{\times}$ & 331  \\
$(128, 2) \rhd (400, 2)$ (NM)               & 47  & 2.5$\boldsymbol{\times}$ & 704  \\
$(128, 2) \rhd (400, 2)$                    & 38  & 2.0$\boldsymbol{\times}$ & 704  \\
$(128, 2) \rhd (256, 2) \rhd (400, 2)$ (NM) & 43  & 2.3$\boldsymbol{\times}$ & 967  \\
$(128, 2) \rhd (256, 2) \rhd (400, 2)$      & 35  & 1.8$\boldsymbol{\times}$ & 967  \\
\bottomrule
\end{tabular}
\end{minipage}
\hfill
\begin{minipage}[t]{0.43\textwidth}
\centering
\tiny
\captionof{table}{Marching cubes runtime (s) on Stanford. Baseline: SIREN $(256,4)$. Multiscale: $(64,1)\!\rhd\!(128,2)\!\rhd\!(400,2)$. Up to $\mathbf{5\times}$ speedup.}
\label{tab:time_mc}
\begin{tabular}{lrrrr}
\toprule
Model & Baseline & No cull. & Cull. & Speedup \\
\midrule
Arm.   & 4.87 & 4.10 & 0.99 & 4.92$\boldsymbol{\times}$ \\
Lucy   & 4.87 & 7.49 & 1.07 & 4.56$\boldsymbol{\times}$ \\
Dragon & 4.88 & 7.28 & 1.71 & 2.85$\boldsymbol{\times}$ \\
Thai   & 4.89 & 7.28 & 1.73 & 2.82$\boldsymbol{\times}$ \\
\bottomrule
\end{tabular}
\end{minipage}
\end{table}

\paragraph{Real-time renderer and mesh extraction.} A CUDA renderer using multiscale ST and GEMM-based normals (via CUTLASS) achieves real-time performance, with neural normal mapping yielding substantial speedups over the SIREN baseline (Tab.~\ref{t-3d}). For mesh extraction (Tab.~\ref{tab:time_mc}), our adaptive grid evaluation skips inference at finer levels far from the level set, achieving up to $5\times$ speedup; smaller surfaces benefit more, since their nesting neighborhoods contain fewer vertices.

\vspace{-2mm}
\section{Conclusion}
\vspace{-1mm}
We propose \method{}, an INR framework modeling neural SDFs as residual SIRENs trained on nested neighborhoods, with multiscale sphere tracing, neural attribute mapping, and GEMM-based analytical normals enabling real-time rendering and accelerated marching-cubes mesh extraction. The nested-band design acts as an implicit regularizer yielding strong noise robustness and extends naturally to differentiable and time-dependent pipelines. \emph{Limitations:} smoothness prevents very sharp edges (addressable with local features), and the maximum in Eq.~\ref{e-deltas} makes the band width sensitive to isolated outlier points. A quantile-based \(\delta\) is a natural robustification (suppl.~Sec.~\ref{s-outliers}), while the adaptive \(\delta\) keeps the hierarchy robust to the coarse stage's training budget (suppl.~Sec.~\ref{s-undertrained}). Future directions include inverse rendering, unsigned distance functions, and fully fused GEMMs~\cite{tiny-cuda-nn}.

\bibliographystyle{plainnat}
\bibliography{references}

\appendix

\section*{Supplementary Material}

\etocsettocstyle{\subsection*{Table of Contents}}{}
\etocsetnexttocdepth{2}
\localtableofcontents

\setcounter{section}{0}
\renewcommand{\thesection}{S\arabic{section}}
\renewcommand{\thefigure}{S\arabic{figure}}
\renewcommand{\thetable}{S\arabic{table}}
\renewcommand{\theequation}{S\arabic{equation}}

\section{Additional qualitative comparisons}

The following figures complement the main-paper qualitative comparisons (Figs.~\ref{coarse} and \ref{all_fines}) with additional illustrative cases referenced in Sec.~4.1 of the main paper.

\begin{figure}[h]
\centering
\includegraphics[width=0.55\linewidth]{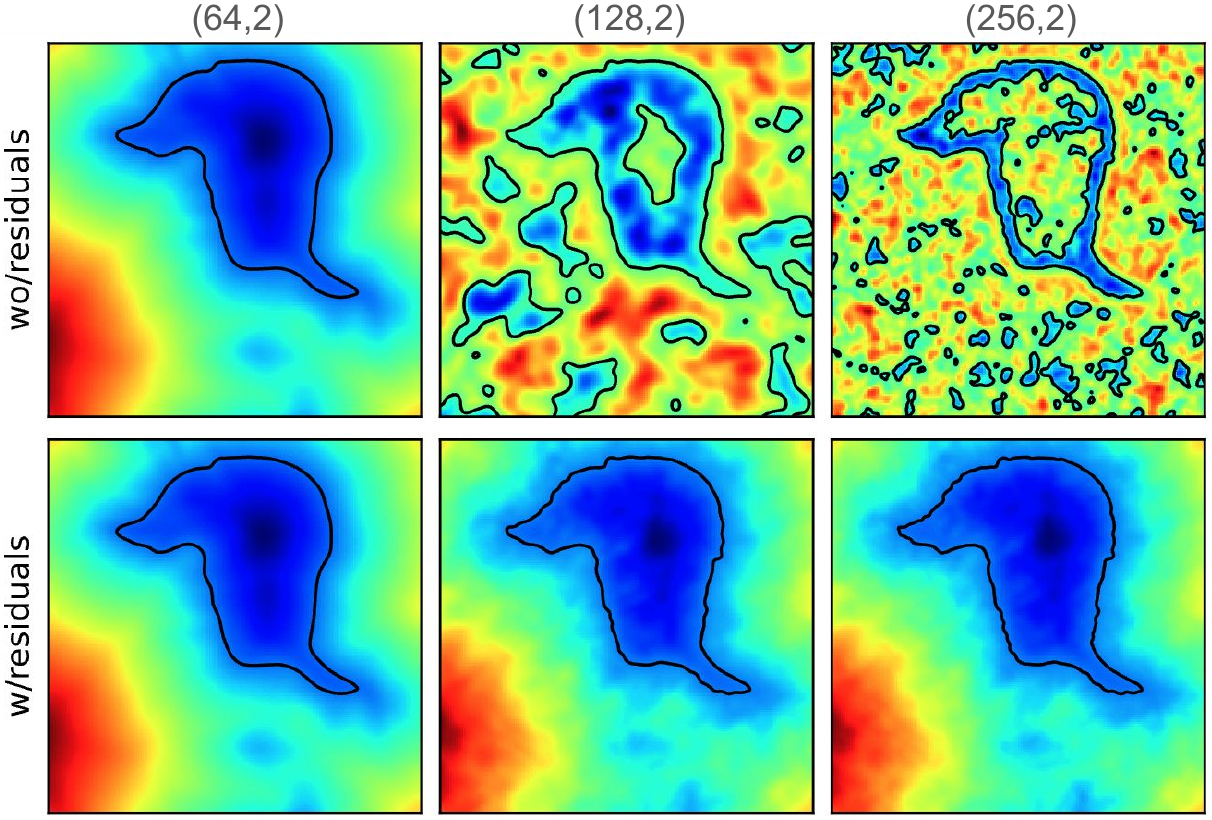}
\caption{Evaluation of the residual approach. Training the SDFs in the neighborhoods (first row: center, right) results in spurious components outside the region. Using the residual approach eliminates those components (second row: center, right).}
\label{f-residual_eval}
\end{figure}

\begin{figure}[h]
\centering
\includegraphics[width=0.35\linewidth]{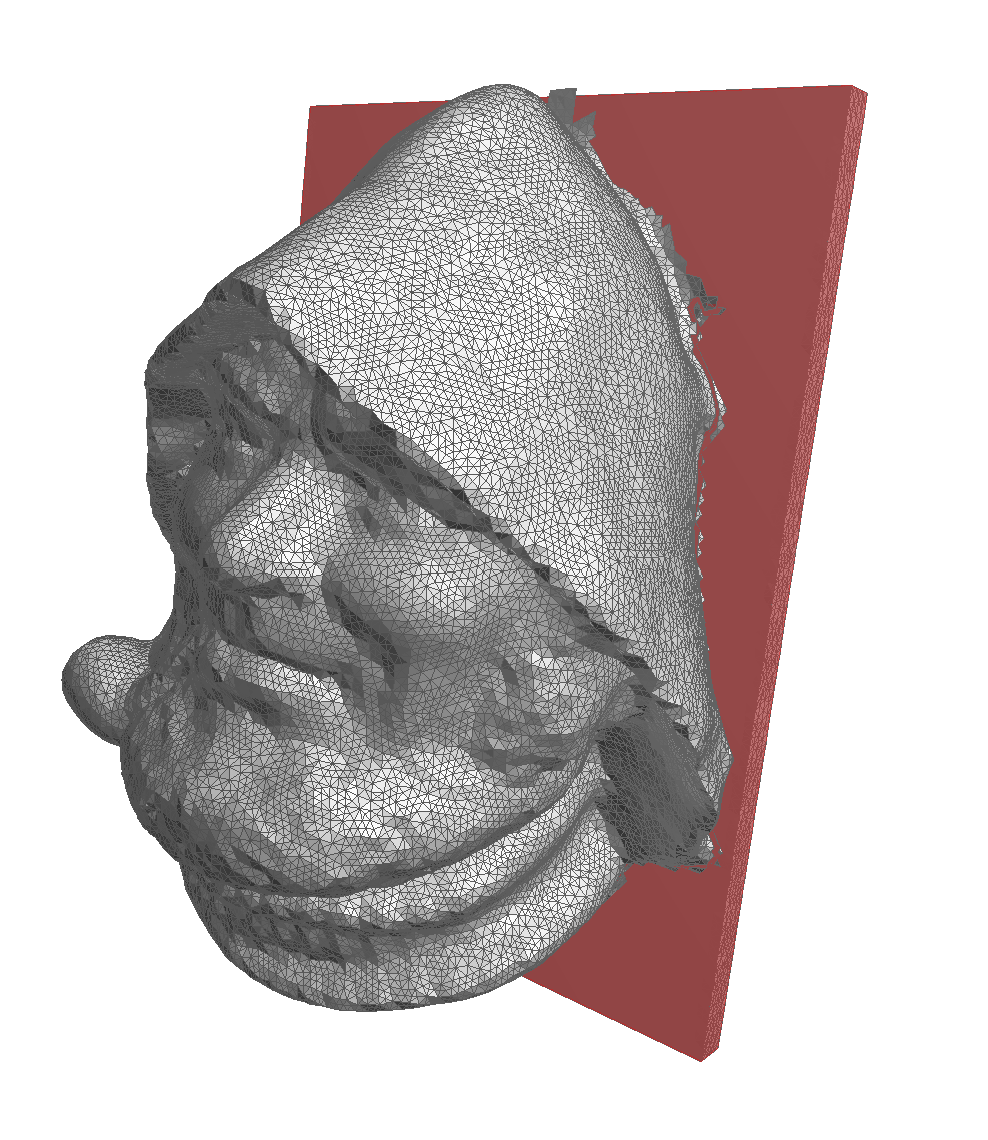}
\caption{Sample from the Thingi32 dataset. The red region corresponds to a separate connected component from the gray surface. Such disconnected structures violate common assumptions of SDF-based representations, making them challenging for methods relying on a single continuous signed distance field.}
\label{sailor}
\end{figure}

\begin{figure}[h]
\centering
\begin{minipage}{0.49\textwidth}
\centering
\includegraphics[width=\linewidth]{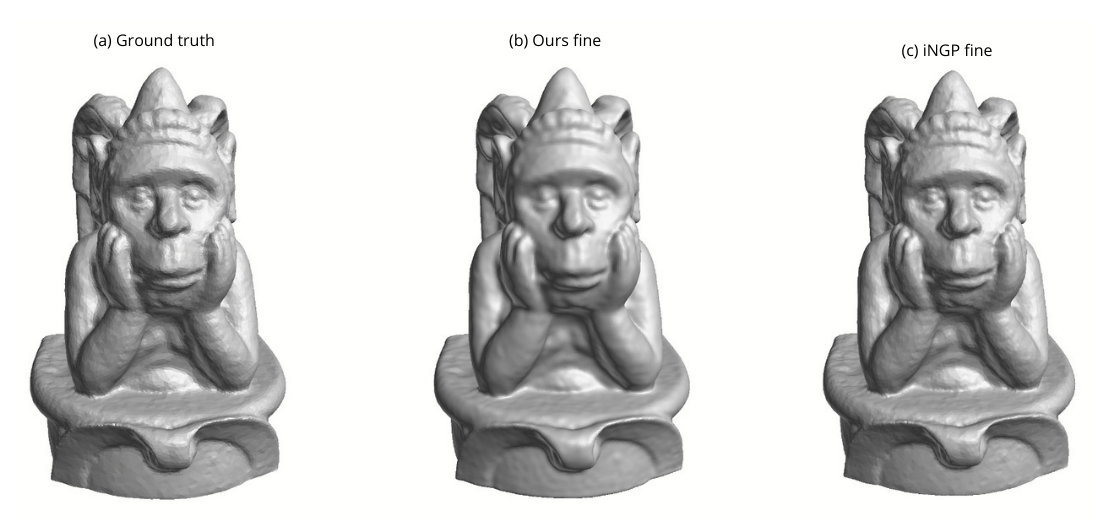}
\caption{Qualitative comparison on a sample from the Thingi32 dataset. (a) Ground-truth reference, (b) our method (fine), and (c) Instant-NGP (fine). iNGP recovers slightly sharper high-frequency details but often introduces a geometric offset relative to the ground truth, which becomes apparent under direct alignment with the reference mesh.}
\label{comparison_ingp}
\end{minipage}
\hfill
\begin{minipage}{0.49\textwidth}
\centering
\includegraphics[width=0.8\linewidth]{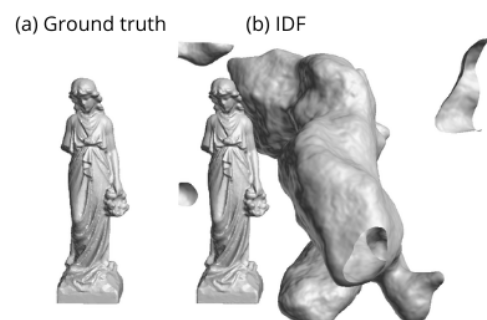}
\caption{Instability observed in IDF reconstructions. (a) Ground-truth reference and (b) corresponding IDF reconstruction; the method generates blob-like artifacts instead of a coherent surface, indicating reduced robustness on complex geometries.}
\label{comparison_idf}
\end{minipage}
\end{figure}

\section{Comparison details}
This section presents detailed model-level results comparisons to enable a deeper analysis of the performance for the evaluated approaches.

\subsection{Per-model metrics}
In the results section of the main paper, we discussed the evaluation of the mean and median performance of coarse and fine representations across several approaches, including the proposed \method{}. Here, we further present the per-model Chamfer Distance values obtained for each approach.

\begin{figure}
    \centering
    \includegraphics[width=1\linewidth]{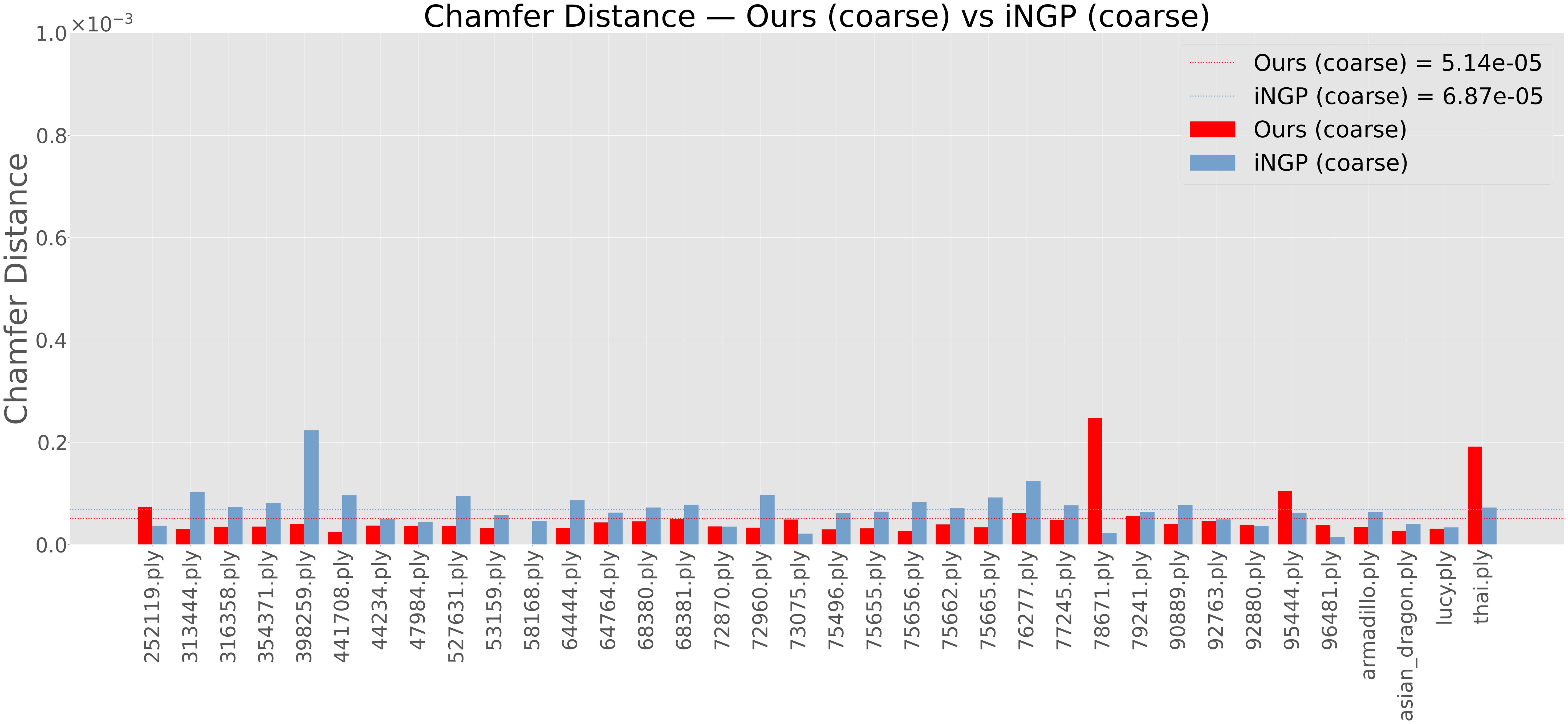}
    \caption{Comparison of Ours Coarse with iNGP Coarse.}
    \label{fig:chamfer_ours+ignp_coarse}
\end{figure}

\begin{figure}
    \centering
    \includegraphics[width=1\linewidth]{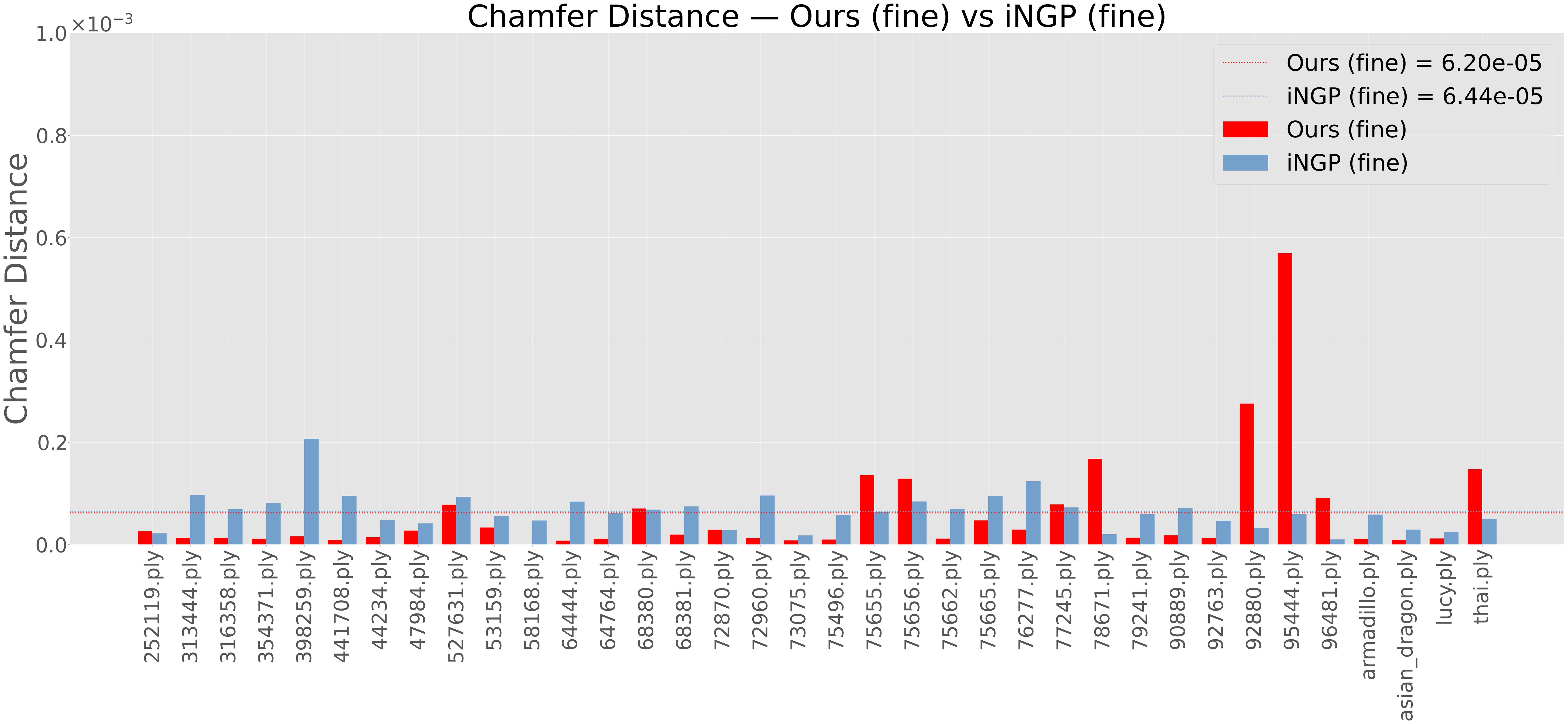}
    \caption{Comparison of Ours Fine with iNGP Fine.}
    \label{fig:chamfer_ours+ignp_fine}
\end{figure}

Figures~\ref{fig:chamfer_ours+ignp_coarse} and~\ref{fig:chamfer_ours+ignp_fine} show the per-model Chamfer Distance comparison between \method{}  and iNGP~\cite{muller2022instant} for the coarse and fine representations, respectively. 
Figures~\ref{fig:chamfer_bacon} and~\ref{fig:chamfer_idf} present the corresponding results for BACON~\cite{lindell2021bacon} and IDF~\cite{wang2022geometry}.

\begin{figure}
    \centering
    \includegraphics[width=0.4\linewidth]{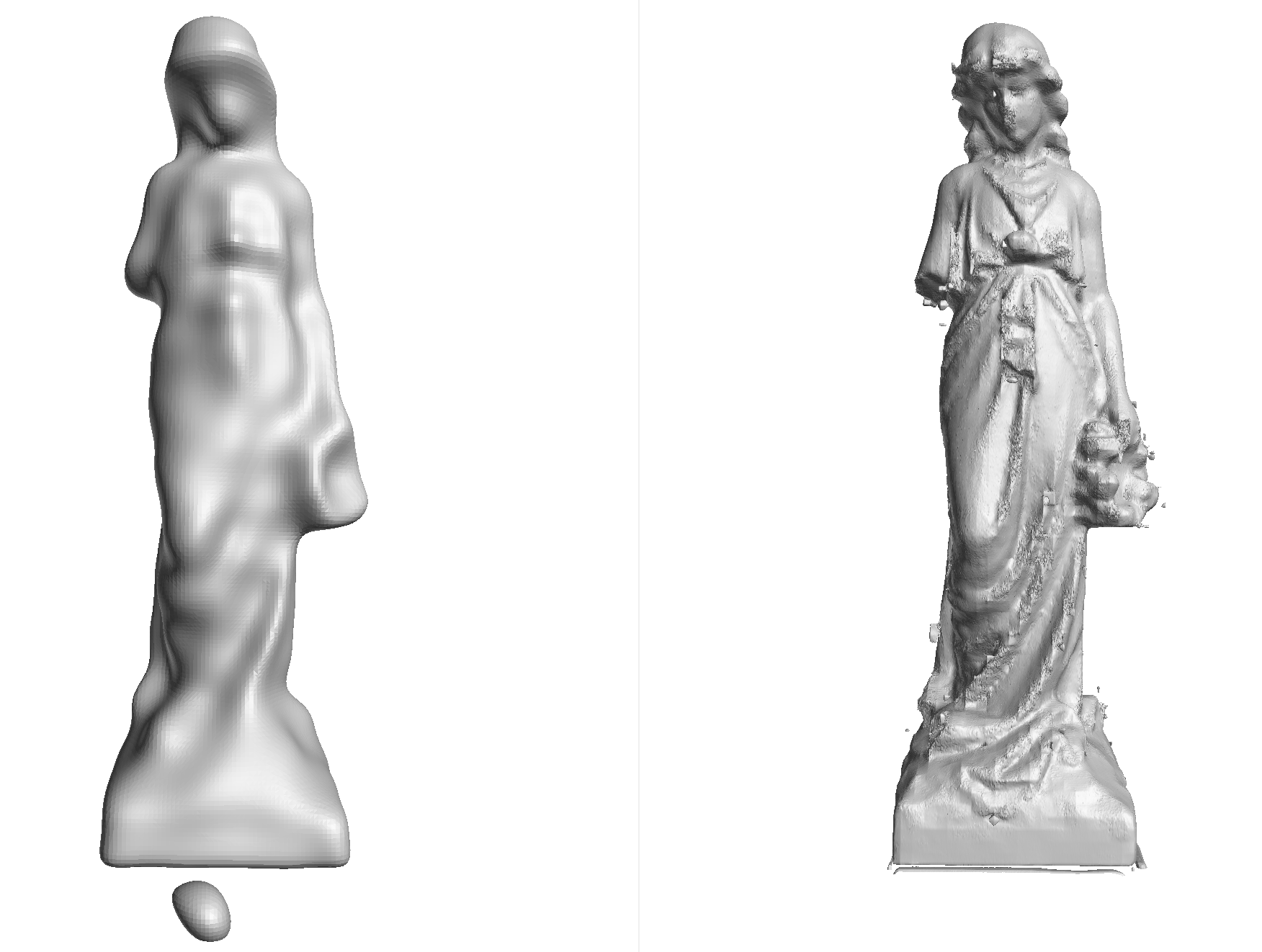}
    \caption{Reconstruction for 78671.ply (ours Coarse vs iNGP Coarse).}
    \label{fig:blob_coarse}
\end{figure}

In Figure~\ref{fig:chamfer_ours+ignp_coarse}, the Chamfer Distance values obtained by \method{} are strongly concentrated around the mean and median, with only a few clearly visible outliers. Notably, the model 78671.ply represents a clear outlier among our coarse reconstructions. As shown in Figure~\ref{fig:blob_coarse}, this behavior is most likely caused by the presence of a blob-like artifact generated during reconstruction, which significantly increases the geometric deviation from the ground truth surface.

When comparing these results with those shown in Figures~\ref{fig:chamfer_bacon} and~\ref{fig:chamfer_idf}, it becomes evident that the Chamfer Distance values obtained by BACON and IDF exhibit a wider dispersion across the dataset. This larger variability indicates less consistent reconstruction quality, with several models presenting higher geometric error. In contrast, the results obtained by \method{} remain more tightly clustered, indicating a more stable reconstruction behavior.

\begin{figure}
    \centering
    \includegraphics[width=1\linewidth]{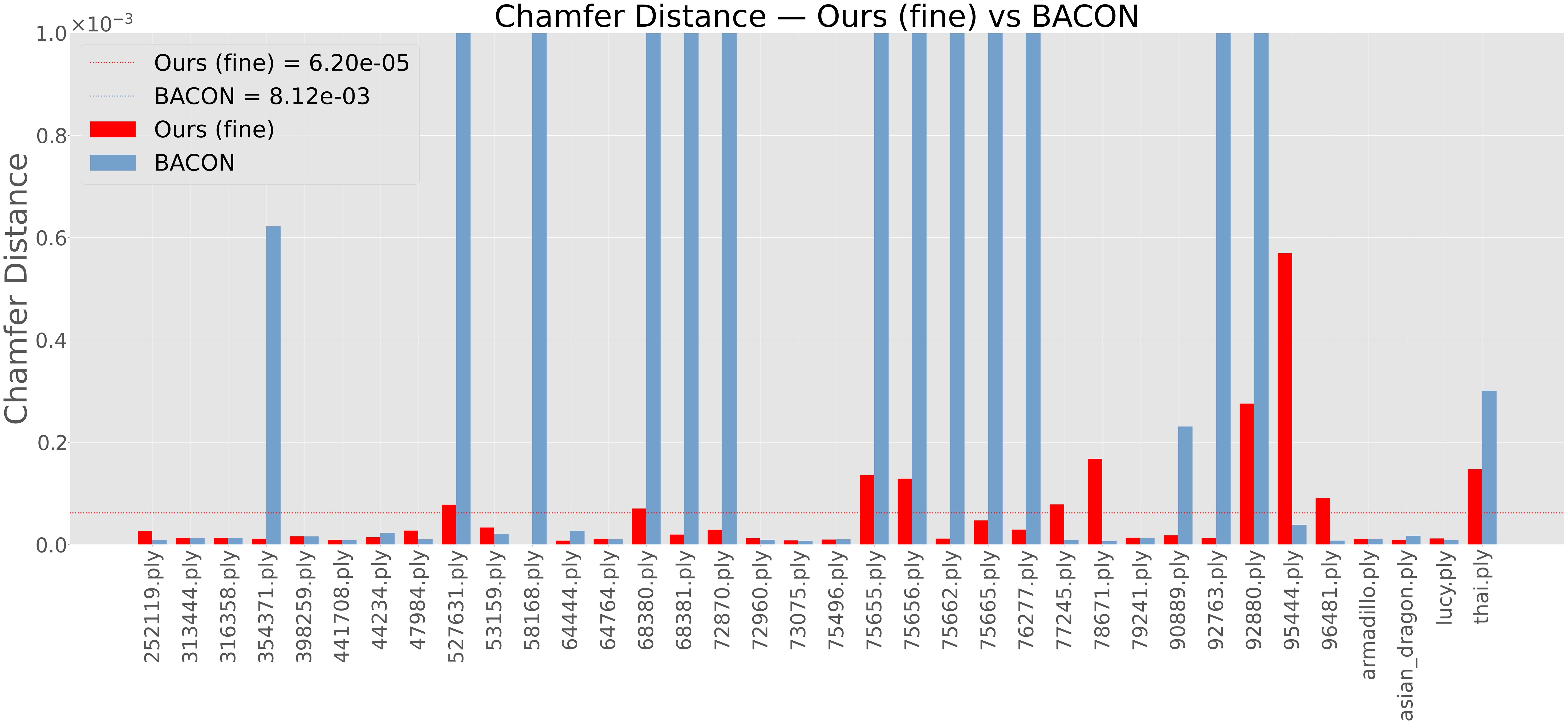}
    \caption{Comparison of Ours Fine with BACON.}
    \label{fig:chamfer_bacon}
\end{figure}

\begin{figure}
    \centering
    \includegraphics[width=1\linewidth]{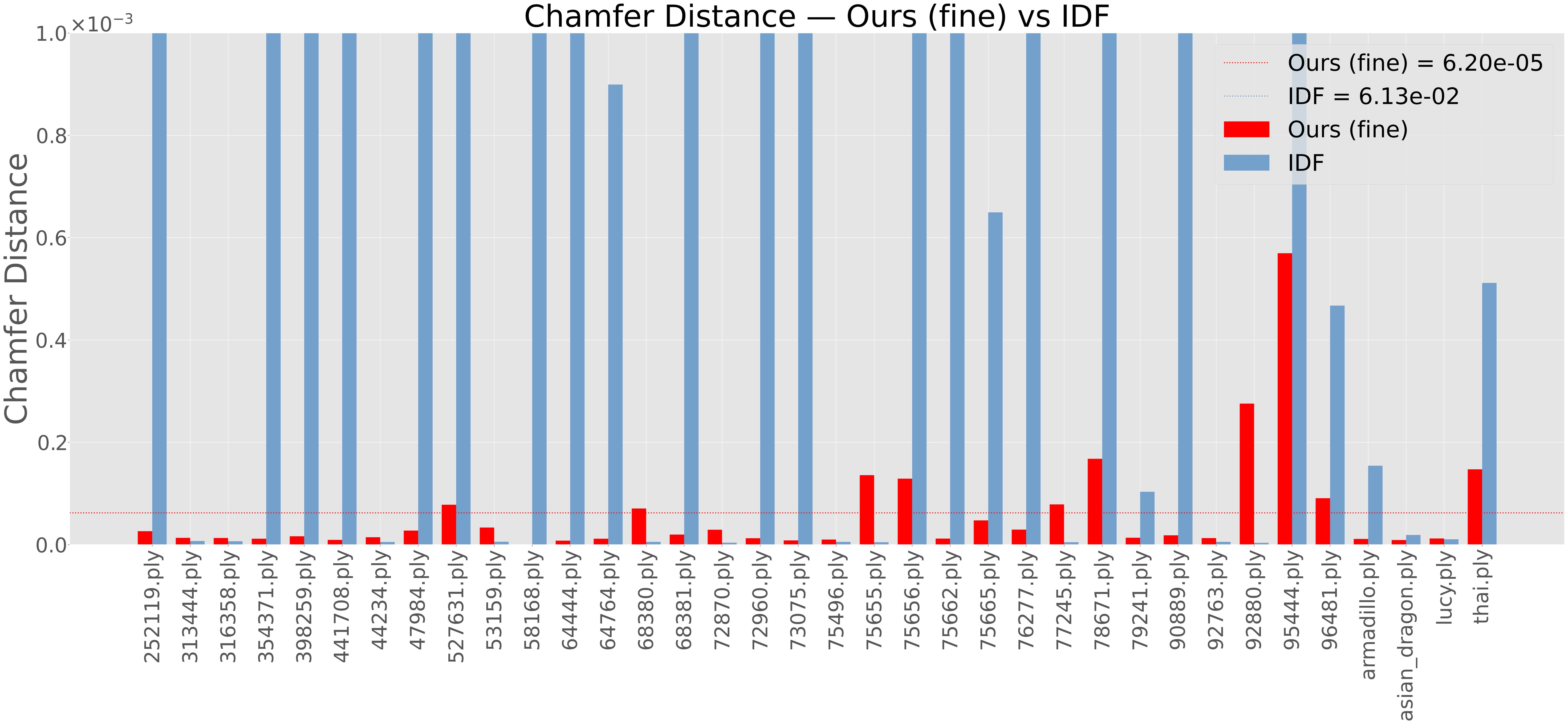}
    \caption{Comparison of Ours Fine with IDF.}
    \label{fig:chamfer_idf}
\end{figure}

Overall, these results suggest that the proposed coarse generation approach provides improved robustness, producing more consistent reconstructions with lower geometric error when compared to iNGP, BACON, and IDF.

In Figure~\ref{fig:chamfer_ours+ignp_fine}, a larger variation in Chamfer Distance values can be observed when compared to the coarse case. Although most reconstructions still present relatively low Chamfer Distance values, the distribution becomes slightly wider for both methods. One factor contributing to this variability is the nature of the Thingi32 dataset itself. Several meshes in this dataset contain geometric characteristics that are not ideal for training Signed Distance Functions, such as multiple disconnected components, non-watertight surfaces, and self-intersections. These properties introduce additional difficulty for implicit reconstruction methods and can lead to localized instability during training. It is also important to note that Table 1 in the main paper shows that the fine case presents improvements in the IoU metric.

Another factor that may contribute to the observed variation is the need for a more refined adjustment of the \method{} parameters for specific objects. In this study, the training procedure was performed using a unified set of parameters for all meshes in the dataset. While this configuration provides a fair and consistent evaluation across models, it does not necessarily yield the optimal reconstruction for every individual shape.

\begin{figure}
    \centering
    \includegraphics[width=0.4\linewidth]{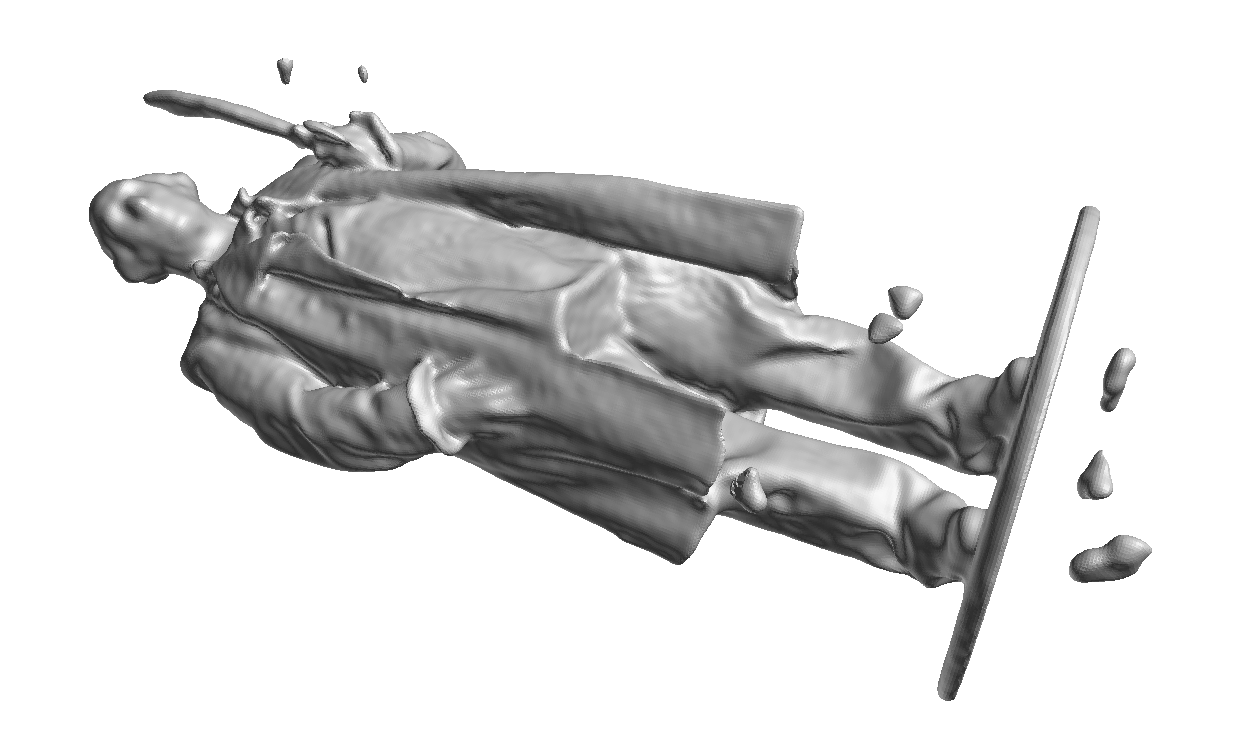}
    \caption{Reconstruction for 95444.ply (ours Fine).}
    \label{fig:blob_95444}
\end{figure}

An example of this behavior can be observed for the model 95444, shown in Figure~\ref{fig:blob_95444}. In this case, the reconstructed mesh presents blob-like artifacts, which increase the geometric deviation from the ground truth surface. These artifacts could likely be reduced with a more careful tuning of the model parameters specifically for this object.

Despite these challenges, \method{} achieves performance comparable to iNGP in terms of geometric robustness and spatial fidelity. The majority of reconstructions remain within the same error scale, indicating that the proposed method is capable of maintaining stable reconstructions even under imperfect training conditions. The next section also shows that our approach is robust to input noise, while iNGP absorbs it.

In contrast, the results presented for BACON and IDF continue to show a higher degree of variability in Chamfer Distance values across the dataset. This behavior suggests that these approaches are more sensitive to geometric irregularities present in the input meshes, leading to less consistent reconstruction quality when compared to the proposed method. Additionally, when compared to our fine model, the obtained Chamfer Distance values tend to be higher, indicating comparatively lower reconstruction accuracy.

\subsection{Per-Model Metrics: Noise Robustness Analysis}

Expanding upon the noise robustness discussion presented in the main paper, this section provides a detailed, per-model analysis of the Chamfer Distance for each evaluated approach under noisy conditions.

\begin{figure}[h]
    \centering
    \includegraphics[width=1\linewidth]{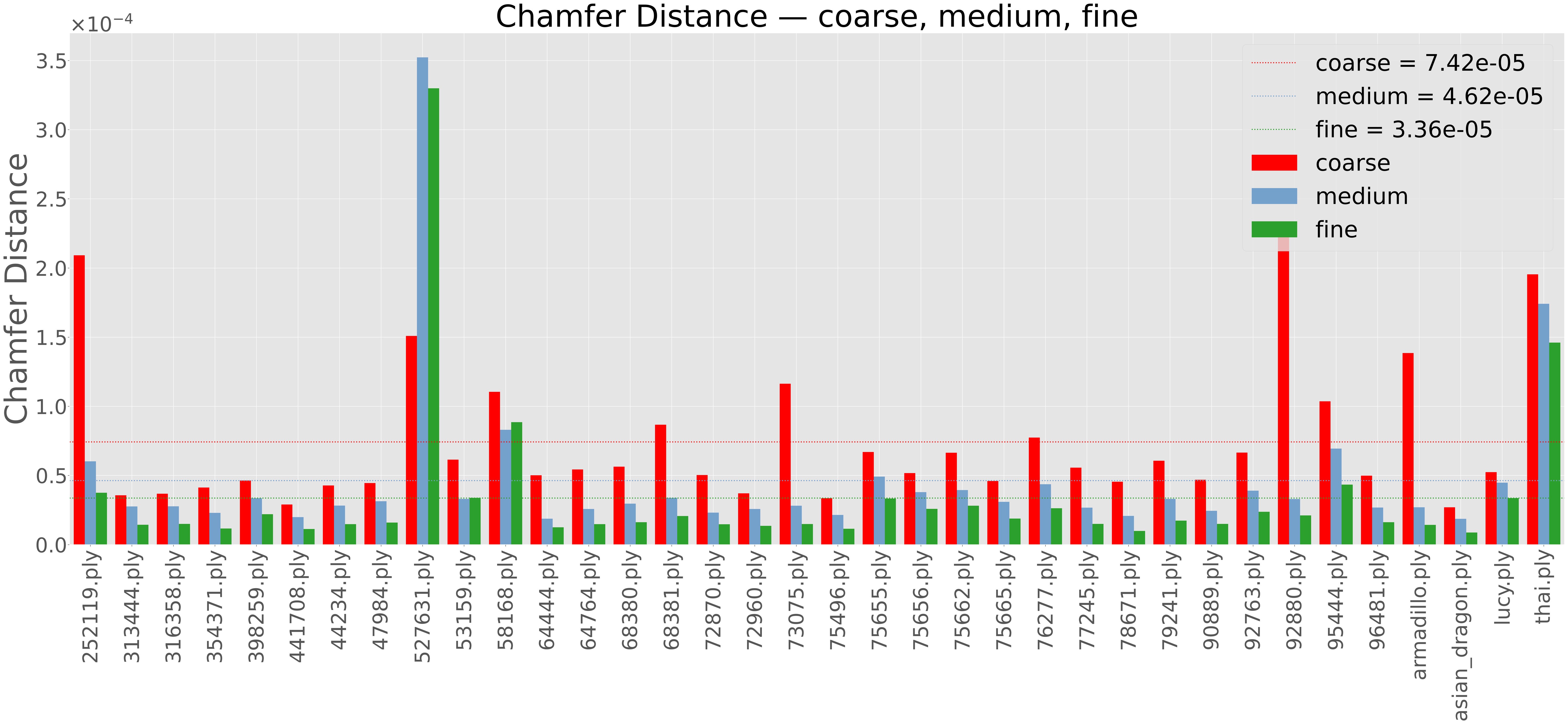}
    \caption{Per-mesh Chamfer Distance for each residual level of our method under 1\% noise.}
    \label{fig:per_mesh_ours_noise_metric}
\end{figure}

In \method{}, the initial coarse level effectively acts as a low-pass filter, significantly attenuating high-frequency noise. This establishes a clean and robust base representation for the subsequent residual levels. As illustrated in Figure \ref{fig:per_mesh_ours_noise_metric}, reconstruction accuracy improves progressively through the hierarchy. The coarse level yields a mean Chamfer Distance of 7.42e-05, which decreases to 4.62e-05 at the medium level, and further improves to 3.36e-05 at the fine level. This progression demonstrates the clear advantage of utilizing the coarse level's low-pass characteristics to guide the finer, high-frequency residual reconstructions without overfitting to the noise.

\begin{figure}[h]
    \centering
    \includegraphics[width=1\linewidth]{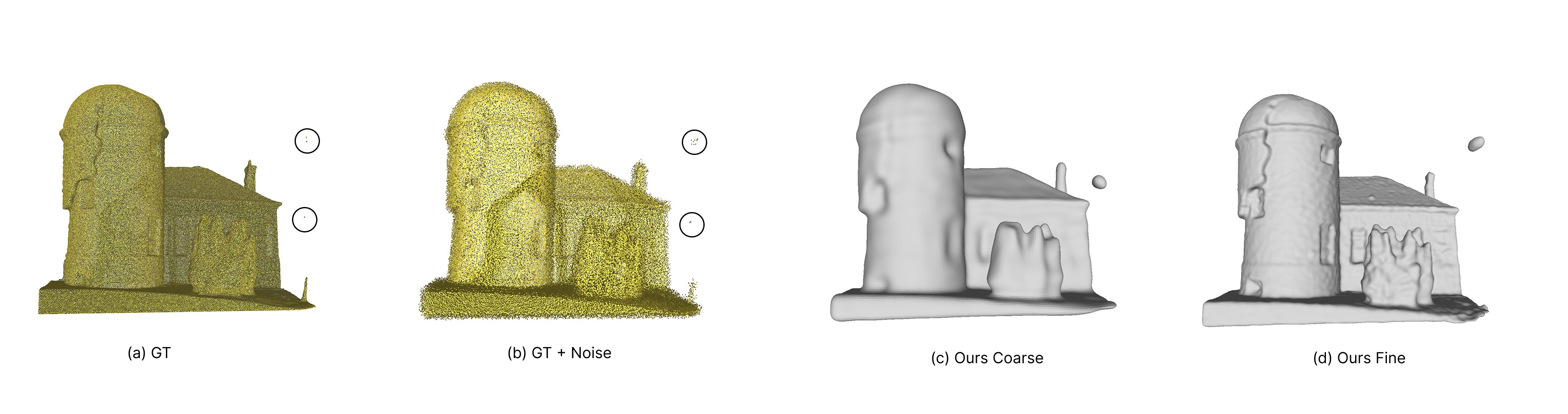}
    \caption{Analysis of outlier sample \texttt{527631.ply} from the Thingi32 Dataset. (a) Original ground truth, (b) ground truth with 1\% noise perturbation, (c) our coarse reconstruction generating an isolated blob, and (d) our fine reconstruction preserving the structure.}
    \label{fig:outlier_analysis}
\end{figure}

One notable outlier in our evaluation is sample \texttt{527631.ply} (see Figure \ref{fig:per_mesh_ours_noise_metric}), where the Chamfer Distance fails to decrease after the initial coarse level. While the general surface noise is successfully removed, structural artifacts remain. As detailed in Figure \ref{fig:outlier_analysis}, the ground truth for this specific sample contains distant, unconnected off-surface point clusters. Consequently, our method accurately smooths the main surface but fits isolated geometry (blobs) to these spatial outliers. The subsequent fine levels preserve these structural blobs, maintaining the higher error metric.

\begin{figure}[h]
    \centering
    \includegraphics[width=1\linewidth]{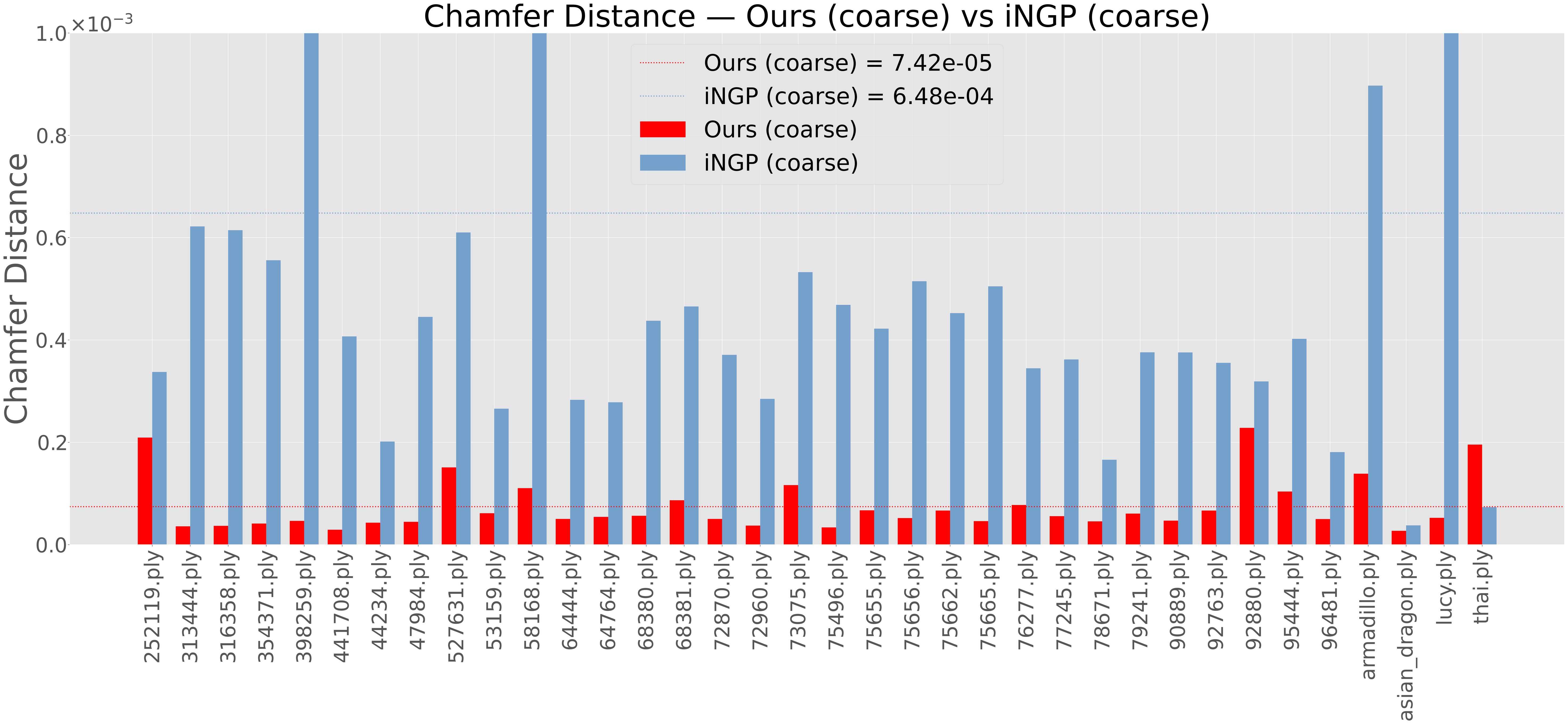}
    \caption{Comparison of our coarse level versus iNGP (coarse) with 1\% noise.}
    \label{fig:our_coarse_vs_ingp_coarse}
\end{figure}

\begin{figure}[h]
    \centering
    \includegraphics[width=1\linewidth]{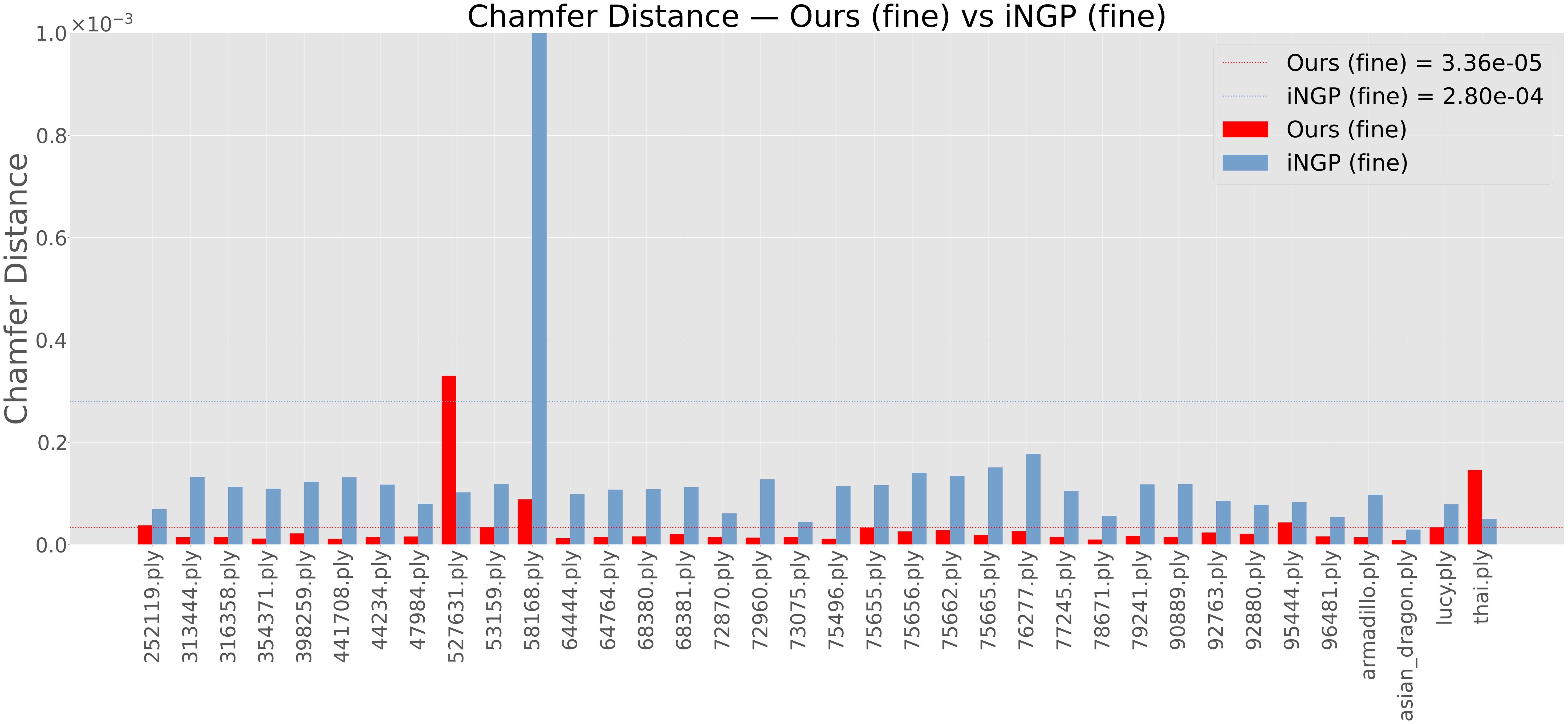}
    \caption{Comparison of our fine level versus iNGP (fine) with 1\% noise.}
    \label{fig:Ours_fine_vs_ingp_fine}
\end{figure}

When compared to baseline methods, our approach demonstrates superior robustness to noise. Figure \ref{fig:our_coarse_vs_ingp_coarse} shows that our coarse model achieves a lower Chamfer Distance than the coarse iNGP model across all evaluated meshes. Furthermore, our fine model outperforms the fine iNGP model in all but two meshes (Figure \ref{fig:Ours_fine_vs_ingp_fine}).

\begin{figure}[h]
    \centering
    \includegraphics[width=1\linewidth]{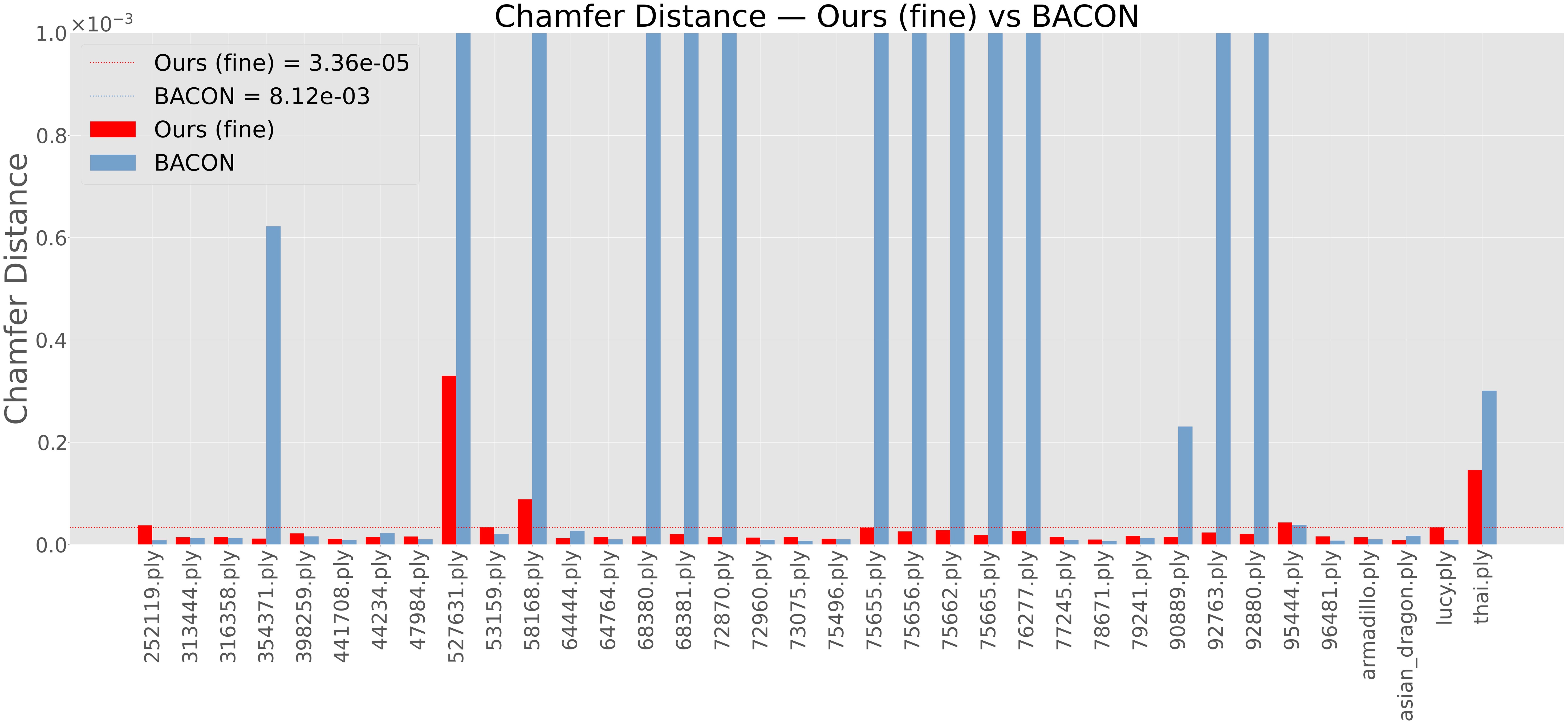}
    \caption{Comparison of our fine level versus BACON with 1\% noise.}
    \label{fig:Ours_fine_vs_BACON}
\end{figure}

\begin{figure}[h]
    \centering
    \includegraphics[width=1\linewidth]{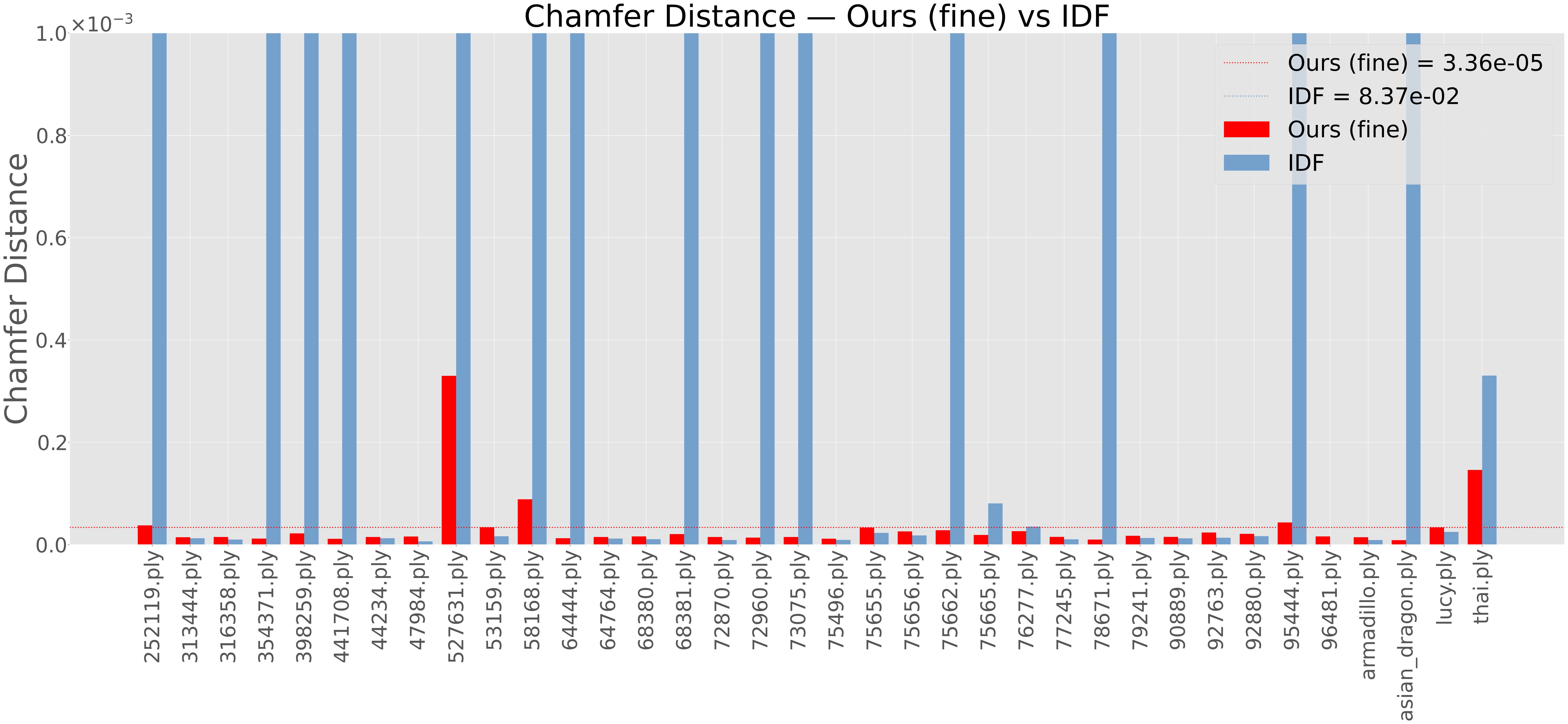}
    \caption{Comparison of our fine level versus IDF with 1\% noise, highlighting IDF's vulnerability to extreme outliers.}
    \label{fig:Ours_fine_vs_IDF}
\end{figure}

Similarly, our fine level consistently outperforms BACON across the entire dataset (Figure \ref{fig:Ours_fine_vs_BACON}). When compared to IDF (Figure \ref{fig:Ours_fine_vs_IDF}), a different trend emerges. While IDF achieves lower errors on several individual meshes, it exhibits instability in the presence of noise. This instability results in extreme outliers, drastically inflating IDF's mean Chamfer Distance (8.37e-02 vs. our 3.36e-05). In contrast, our method remains stable and avoids such failures, ultimately providing a more robust and reliable reconstruction.

\subsection{Comparison with Screened Poisson reconstruction}
\label{s-spsr}

We compare against Screened Poisson surface reconstruction (SPSR)~\cite{kazhdan2013screened} (Open3D, depth 10, with its standard 1\% low-density trim), on the 10-shape subset and evaluation protocol of Sec.~\ref{s-isolation}, using the same clean and noisy inputs as Tab.~3 of the main paper. Table~\ref{tab:spsr} reports mean CD and mean IoU.

\begin{table}[!h]
\centering
\scriptsize
\caption{Comparison with Screened Poisson reconstruction (depth 10, 1\% density trim). Mean CD and IoU over the 10-shape subset of Sec.~\ref{s-isolation}.}
\begin{tabular}{lcc}
\toprule
\textbf{Setting} & \textbf{SPSR CD / IoU} & \textbf{\method{} CD / IoU} \\ \midrule
clean     & 7.16E-05 / \textbf{0.910} & \textbf{3.87E-05} / 0.653 \\
1\% noise & 9.98E-05 / 0.416 & \textbf{6.39E-05} / \textbf{0.687} \\ \bottomrule
\end{tabular}
\label{tab:spsr}
\end{table}

On clean, dense, well-oriented inputs SPSR is excellent, attaining the best mean IoU of any method we evaluated. That strength is expected of a classical direct solver: handed dense samples with exact normals, SPSR fits the occupancy indicator itself (one global linear solve on a depth-10 adaptive octree, capacity growing with the data rather than fixed) so occupancy agreement is close to its objective. A compact neural SDF approximates under a deliberate band-limited prior instead of interpolating. Under the 1\% noise of Tab.~3, SPSR loses more than half its occupancy accuracy (0.910 $\rightarrow$ 0.416) while \method{} holds (0.653 $\rightarrow$ 0.687). Running SPSR \emph{without} its density trim (an outlier-rejection heuristic no other compared method uses) gives mean IoU 0.867 clean and 0.394 under noise. SPSR outputs a triangle mesh, which rasterises in real time once extracted, so we make no rendering-speed claim against it. The differences are in what the representation affords (distance queries away from the surface, the attribute mapping of Sec.~\ref{s-attribute-mapping}, and level of detail from a single compact model).

\subsection{Comparison with BANF via a custom re-implementation}
\label{s-banf}

BANF's released code targets image fitting, NeRFing, and multi-view 3D reconstruction, without an official implementation covering SDF fitting from point clouds. We therefore re-implemented BANF from its method section. The numbers below reflect our custom implementation, not the authors' code, and should be read with that caveat.

\emph{Configuration.} Band-limiting with each level evaluated at its own lattice and reconstructed by linear interpolation, cascaded residual training (level $k$ fits $f - \sum_{l<k} f_l$, recomposed as $\sum_k f_k$), their losses (L2 on ground-truth SDF, Eikonal, Laplacian, with derivatives by finite differences at $\delta = 1/(4r)$), their sampling (500K pool, 40/40/20 on-surface/near-surface/uniform), their schedule ($\{32,64,128,256\}$ lattices; 5K iterations at level 0 with warmup, 10K per finer level; Adam $10^{-3}$, batch 100K, sphere initialization), and the iNGP hash encoding (16 levels $\times$ 2 features, table $2^{19}$, resolutions 16$\rightarrow$2048, $3{\times}32$ decoder).

\emph{Fidelity.} Our cascade produces the resolutions BANF tabulates, so we can check against their published numbers. In their convention (mean Euclidean Chamfer, $\times 10^{-2}$) we obtain at lattices 32/64/128 on the Asian Dragon 1.57 / 1.20 / 0.92 against their 1.23 / 0.65 / 0.32, and on the Thai Statue 1.17 / 0.92 / 0.74 against 1.41 / 0.68 / 0.40. We match at the coarsest level (0.83--1.27$\times$) but fall behind at finer ones (up to 2.9$\times$), likely an optimisation gap (BANF trains inside iNGP's optimised stack, our re-implementation is pure PyTorch). Absolute comparisons should be read with that gap in mind.

\begin{table}[!h]
\centering
\scriptsize
\caption{Comparison with our BANF re-implementation (CD; paper protocol). ``degr.'' is the clean$\rightarrow$noise degradation factor; best of each pair in bold.}
\begin{tabular}{lrrrr}
\toprule
 & \textbf{Dragon} & \textbf{Armadillo} & \textbf{Lucy} & \textbf{Thai} \\ \midrule
\method{} clean & \textbf{8.65E-06} & 7.41E-05 & \textbf{2.18E-05} & 1.94E-04 \\
BANF clean      & 3.03E-04 & \textbf{2.79E-05} & 1.03E-04 & \textbf{8.27E-05} \\ \midrule
\method{} noise & \textbf{8.69E-06} & \textbf{1.38E-05} & \textbf{3.33E-05} & \textbf{4.90E-04} \\
BANF noise      & 3.20E-04 & 2.03E-04 & 1.18E-02 & 1.07E-02 \\ \midrule
\method{} degr. & \textbf{1.0$\times$} & \textbf{0.2$\times$} & \textbf{1.5$\times$} & \textbf{2.5$\times$} \\
BANF degr.      & 1.1$\times$ & 7.3$\times$ & 115$\times$ & 130$\times$ \\ \bottomrule
\end{tabular}
\label{tab:banf}
\end{table}

Two within-method observations survive the re-implementation gap. On clean data BANF is the better method on the Thai Statue and the Armadillo. And its noise degradation tracks thin-structure content (negligible on the solid Dragon, moderate on the Armadillo, catastrophic on Lucy and Thai, whose occupancy collapses to IoU 0.001). On Thai, noise takes the mesh from 81 to 474 connected components. The finer levels turn noise into off-surface geometry, which our nested bands prevent structurally, residuals never being supervised or evaluated outside the $\delta$-band (Fig.~\ref{f-residual_eval}). \method{} stays within 2.5$\times$ throughout. Independently of this re-implementation, cell (b) of the ablation in Sec.~\ref{s-isolation} isolates the cascaded full-domain supervision pattern BANF shares with IDF and BACON, and iNGP covers the grid-based representation axis.

\section{Additional ablation studies}

\paragraph{Loss terms and neighborhood $\delta$.} We performed two additional ablation studies for our approach: (i) loss term assessment, (ii) $\delta$ influence over the reconstructions. %
Tables~\ref{tab:abl-grad},\ref{tab:abl-normal}, and~\ref{tab:abl-sdf} show the ablation results of our loss function using different weights for each component, while maintaining the remaining hyper-parameters fixed. We performed these studies both for the medium level and the fine level. Note that all studies used the Lucy mesh as a baseline. Table~\ref{tab:abl-delta} shows the results for varying the delta values while maintaining the remaining hyper-parameters fixed.

\begin{table}[!h]
\centering
        \scriptsize
\caption{Gradient constraint ablation studies.}
\subfloat[Gradient constraint fine level.]{
\begin{tabular}{rr}
\toprule
\textbf{Gradient constraint} & \textbf{Approx. Error} \\ \midrule
0.0                                           & 0.0013                           \\
10.0                                          & 0.0013                           \\
30.0                                          & 0.0013                           \\
100.0                                         & 0.0012                           \\
300.0                                         & 0.0013                           \\
1000.0                                        & 0.0014                           \\
3000.0                                        & 0.0017                           \\
10000.0                                       & 0.0022                           \\
30000.0                                       & 0.0030                            \\ \bottomrule
\end{tabular}}
\hspace{1cm}
\subfloat[Gradient constraint medium level.]{
\centering
        \scriptsize
\begin{tabular}{rr}
\toprule
\textbf{Gradient Constraint} & \textbf{Approx. Error} \\ \midrule
0.0                          & 0.0086                 \\
10.0                         & 0.0084                 \\
30.0                         & 0.0082                 \\
100.0                        & 0.0078                 \\
300.0                        & 0.0074                 \\
1000.0                       & 0.0069                 \\
3000.0                       & 0.0073                 \\
10000.0                      & 0.0087                 \\
30000.0                      & 0.0116                 \\ \bottomrule
\end{tabular}}
\label{tab:abl-grad}
\end{table}

\begin{table}[!h]
\centering
\scriptsize
\caption{Normal constraint ablation studies.}
\subfloat[Normal constraint fine level.]{
\begin{tabular}{rr}
\toprule
\textbf{Normal Constraint} & \textbf{Approx. Error} \\ \midrule
0.0                        & 0.0017                 \\
10.0                       & 0.0013                 \\
30.0                       & 0.0013                 \\
100.0                      & 0.0013                 \\
300.0                      & 0.0013                 \\
1000.0                     & 0.0013                 \\
3000.0                     & 0.0013                 \\
10000.0                    & 0.0013                 \\
30000.0                    & 0.0013                 \\ \bottomrule
\end{tabular}}
\hspace{1cm}
\subfloat[Normal constraint medium level.]{
\begin{tabular}{rr}
\toprule
\textbf{Normal Constraint} & \textbf{Approx. Error} \\ \midrule
0.0                        & 0.0073                 \\
10.0                       & 0.0074                 \\
30.0                       & 0.0077                 \\
100.0                      & 0.0081                 \\
300.0                      & 0.0083                 \\
1000.0                     & 0.0085                 \\
3000.0                     & 0.0086                 \\
10000.0                    & 0.0087                 \\
30000.0                    & 0.0087                 \\ \bottomrule
\end{tabular}
}
\label{tab:abl-normal}
\end{table}

\begin{table}[]
\centering
\scriptsize
\caption{SDF constraint ablation studies.}
\subfloat[SDF constraint fine level.]{
\begin{tabular}{rr}
\toprule
\textbf{SDF Constraint} & \textbf{Approx. Error} \\ \midrule
0.0                     & 0.0076                 \\
10.0                    & 0.0013                 \\
30.0                    & 0.0013                 \\
100.0                   & 0.0013                 \\
300.0                   & 0.0013                 \\
1000.0                  & 0.0012                 \\
3000.0                  & 0.0013                 \\
10000.0                 & 0.0013                 \\
30000.0                 & 0.0013                 \\ \bottomrule
\end{tabular}}
\hspace{1cm}
\subfloat[SDF constraint medium level.]{
\begin{tabular}{rr}
\toprule
\textbf{SDF Constraint} & \textbf{Approx. Error} \\ \midrule
0.0                     & 0.0490                  \\
10.0                    & 0.0080                  \\
30.0                    & 0.0079                 \\
100.0                   & 0.0079                 \\
300.0                   & 0.0080                  \\
1000.0                  & 0.0081                 \\
3000.0                  & 0.0080                  \\
10000.0                 & 0.0082                 \\
30000.0                 & 0.0082                 \\ \bottomrule
\end{tabular}
}
\label{tab:abl-sdf}
\end{table}

\begin{table}[]
\centering
\scriptsize
\caption{Ablation studies of the delta factor. We multiply the delta by the values in the first column and measure the SDF error compared to the Open3D calculated SDF, which we use an ground-truth.}
\begin{tabular}{rrr}
\toprule
\textbf{Max delta fraction} & \textbf{Medium level error} & \textbf{Fine level error} \\ \midrule
1.01                        & 0.0098                      & 0.0048                    \\
1.05                        & 0.0098                      & 0.0048                    \\
1.10                         & 0.0100                      & 0.0049                    \\
1.20                         & 0.0101                      & 0.0049                    \\
1.30                         & 0.0103                      & 0.0050                    \\
1.50                         & 0.0106                      & 0.0051                    \\
2.00                         & 0.0113                      & 0.0053                    \\
5.00                         & 0.0139                      & 0.0066                    \\ \bottomrule
\end{tabular}
\label{tab:abl-delta}
\end{table}

\subsection{Isolating the residual composition and the nested bands}
\label{s-isolation}

We ablate the two design axes separately, against the full method as the reference, on 10 shapes (Armadillo, Asian Dragon, Lucy, Thai Statue from Stanford; 44234, 64764, 68381, 72870, 73075, 77245 from Thingi32), clean and under the 1\% vertex noise of Tab.~3, with the paper's evaluation protocol (500K samples, $L_2$ Chamfer, Hausdorff distance, voxel IoU, $512^3$ extraction). All cells share the $(128,2)\triangleright(256,2)\triangleright(400,2)$ architectures, the $\omega_0=30/45/100$ schedule, and the same coarse network. On clean inputs cell (a) is the released model. Cell (b) is \emph{residual learning without nested bands} (residuals supervised over the full domain, the IDF/BACON/BANF pattern), cell (c) is \emph{nested bands without residual learning} (independent SDFs, each trained inside the previous level's band). Each cell is extracted with its own sound inference: band-culled hierarchical marching cubes for (a) and (c), full-domain extraction for (b), whose residuals are trained domain-wide, so culling it would be unsound.

\begin{table}[!h]
\centering
\scriptsize
\caption{Isolation ablation: mean CD / median CD / mean Hausdorff / mean IoU over 10 shapes. Best per column in bold.}
\begin{tabular}{lcc}
\toprule
\textbf{Variant} & \textbf{1\% noise} & \textbf{clean} \\ \midrule
(a) w residual, w bands & 6.39E-05 / \textbf{1.49E-05} / \textbf{1.45E-03} / \textbf{0.687} & 3.87E-05 / \textbf{1.38E-05} / 5.86E-03 / 0.653 \\
(b) w residual, w/o bands & \textbf{6.34E-05} / 2.26E-05 / 5.23E-03 / 0.555 & 4.26E-05 / 2.36E-05 / 6.75E-03 / 0.655 \\
(c) w/o residual, w bands & 1.04E-04 / 2.36E-05 / 5.72E-03 / 0.669 & \textbf{3.47E-05} / 1.62E-05 / \textbf{4.01E-03} / \textbf{0.749} \\ \bottomrule
\end{tabular}
\label{tab:isolation}
\end{table}

Comparing (a) and (b) isolates the nested bands. Typical-case Chamfer is a tie, but the tail separates: mean Hausdorff is $3.6\times$ worse without bands (5.23E-03 versus 1.45E-03), and under noise (b) leaves off-surface components on 4 of 10 shapes (reaching 0.135 from the true surface on the Armadillo) against 1 of 10 for the full method, which reconstructs 8 of 10 noisy shapes as a single connected component. This is the quantitative form of Tab.~3's claim: nested bands stop noise from turning into off-surface geometry.

Comparing (a) and (c) isolates the residual sum, which is what noise robustness needs: (c) leads the clean aggregates but degrades $3.0\times$ under noise in mean CD (3.47E-05 $\rightarrow$ 1.04E-04) where the full method stays within $1.7\times$, has higher Hausdorff on 10 of 10 noisy shapes, and fragments even under culled extraction (spurious components on 7 of 10 noisy shapes, up to 373 on 77245). Re-extracting the same noisy checkpoints \emph{without} band culling separates them completely: the residual sum is essentially unchanged (mean CD 6.39E-05 $\rightarrow$ 6.42E-05, 1--4 components per shape), while the independent band-trained SDFs disintegrate into 365 to 4{,}109 components filling the whole domain (mean CD 2.8E-01). That measures the phenomenon Fig.~\ref{f-residual_eval} shows. Without the residual coupling the composed field is only meaningful inside the band, and the design works only if inference culls.

\subsection{Robustness to coarse-stage under-training}
\label{s-undertrained}

To evaluate sensitivity to the coarse stage's convergence, we trained $f_1$ for 25\%, 50\%, and 100\% of its epochs and ran the unchanged pipeline on top (Armadillo and Lucy, protocol of Sec.~\ref{s-isolation}).

\begin{table}[!h]
\centering
\scriptsize
\caption{Coarse-stage under-training: the adaptive band absorbs an under-trained $f_1$.}
\begin{tabular}{lcccc}
\toprule
\textbf{Coarse epochs} & $\boldsymbol{\delta_1}$ \textbf{(Arm.)} & \textbf{final CD (Arm.)} & $\boldsymbol{\delta_1}$ \textbf{(Lucy)} & \textbf{final CD (Lucy)} \\ \midrule
25\%  & 6.20E-02 & 1.11E-05 & 2.83E-02 & 7.13E-05 \\
50\%  & 2.00E-02 & 1.06E-04 & 2.70E-02 & 5.03E-05 \\
100\% & 1.71E-02 & 9.81E-05 & 2.94E-02 & 5.24E-05 \\ \bottomrule
\end{tabular}
\label{tab:undertrained}
\end{table}

The band tracks the predecessor's convergence as designed: with a 25\%-trained coarse stage, the Armadillo's $\delta_1$ widens automatically by $3.6\times$ and the final quality is unchanged or better (CD 1.11E-05, IoU 0.91), the slack giving the residual more room. Lucy is stable across all three budgets. The stages are therefore not fragile to predecessor state. As for over-training, the coarse level's $\omega_0=30$ bounds its spectrum, so it cannot absorb high-frequency noise regardless of epochs because of the low-pass mechanism behind Tab.~3.

\subsection{Outlier contamination}
\label{s-outliers}

Following the setting of outliers without noise, we add 0.1\% and 1\% stray points, drawn uniformly inside the unit ball with random normals, to the clean inputs of the Armadillo and Lucy, and evaluate against the clean ground truth (IoU). For reference we run SPSR on the same contaminated clouds, both with and without its low-density trim (an outlier-rejection heuristic).

\begin{table}[!h]
\centering
\scriptsize
\caption{Outlier contamination (IoU against clean ground truth).}
\begin{tabular}{lccc}
\toprule
\textbf{Contamination} & \textbf{\method{}} & \textbf{SPSR (no trim)} & \textbf{SPSR (trim)} \\ \midrule
Armadillo 0.1\% & 0.909 & 0.982 & 0.978 \\
Armadillo 1\%   & 0.863 & 0.985 & 0.981 \\
Lucy 0.1\%      & 0.799 & 0.995 & 0.986 \\
Lucy 1\%        & 0.002 & 0.004 & 0.938 \\ \bottomrule
\end{tabular}
\label{tab:outliers}
\end{table}

Sparse contamination is handled (the Armadillo holds IoU 0.86 at 1\%), while heavy contamination on Lucy defeats our method and, without its trim, defeats SPSR just as completely (0.004). Its advantage there comes from the rejection heuristic instead of the representation. The mechanism on our side is narrow: Eq.~5 of the main paper sets each band from the \emph{maximum} residual error over the input points, so a single stray point inflates it (on Lucy, $\delta_1$ grows from 2.7E-02 to 1.1 at 0.1\% contamination, degenerating the band to the whole domain). A quantile-based $\delta$ in place of the max is the natural robustification. This behaviour is consistent with the outlier analysis of Fig.~\ref{fig:outlier_analysis}. The robustness claims of Tab.~3 concern measurement noise, for which the mechanism is intact, rather than outlier contamination, for which classical pipelines prescribe prefiltering.

\subsection{Measured adaptive band widths}
\label{s-deltas}

Table~\ref{tab:deltas} reports the adaptive band widths of Eq.~5 ($\varepsilon = 0.3$) measured from the released models on the unit-sphere domain (radius 1). Every band is under 3\% of the domain radius and shrinks from level to level as the approximation improves. On noisy inputs the bands widen (Thai: $\delta_1 = 6.9$E-02, $\delta_2 = 5.2$E-02), which is the intended adaptive behaviour. A noisier fit leaves more residual error at the data, so the next level gets more room.

\begin{table}[!h]
\centering
\scriptsize
\caption{Measured band widths (clean inputs, unit-sphere domain).}
\begin{tabular}{lcc}
\toprule
\textbf{Shape} & $\boldsymbol{\delta_1}$ & $\boldsymbol{\delta_2}$ \\ \midrule
Armadillo    & 1.68E-02 & 1.11E-02 \\
Asian Dragon & 2.93E-02 & 1.65E-02 \\
Lucy         & 2.67E-02 & 1.43E-02 \\
Thai Statue  & 2.71E-02 & 1.57E-02 \\ \bottomrule
\end{tabular}
\label{tab:deltas}
\end{table}

\section{Rendering and mesh extraction details}
\label{s-rendering}

To render the zero level set \(f^{-1}(0)\) of a SDF \(f\), two common strategies can be used: \textbf{sphere tracing} (ST)~\cite{hart1989ray}, which directly traces rays through the SDF field, and \textbf{marching cubes}~\cite{lorensen1987marching}, which extracts an explicit mesh followed by standard rasterization. When the SDF is represented using \method{}, both rendering approaches become more efficient.

\begin{figure}[htb]
    \centering
    \begin{minipage}[t]{0.48\textwidth}
        \includegraphics[width=1.0\textwidth]{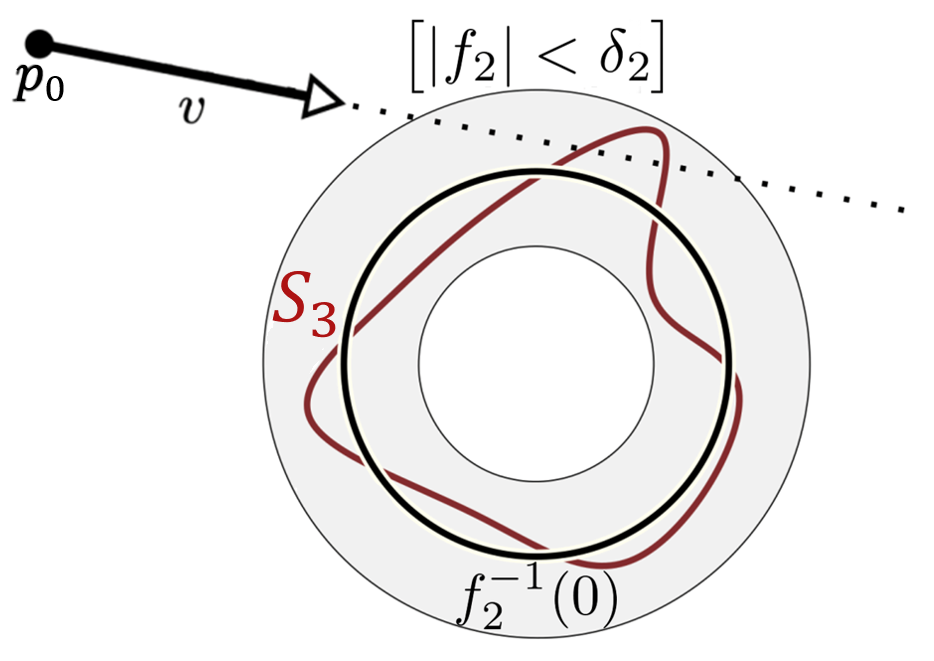}
        \caption{Ray intersecting $S_3$ nested in a $\delta$-neighborhood of a coarse SDF $f_2$. Notice that sphere tracing $f_2$ directly would lead to a false negative, thus we use $\big[|f_{2}|<\delta_2\big]$ instead.}
        \label{f-sphere_tracing}
    \end{minipage}
     \hfill
    \begin{minipage}[t]{0.48\textwidth}
        \centering
        \includegraphics[width=0.85\textwidth]{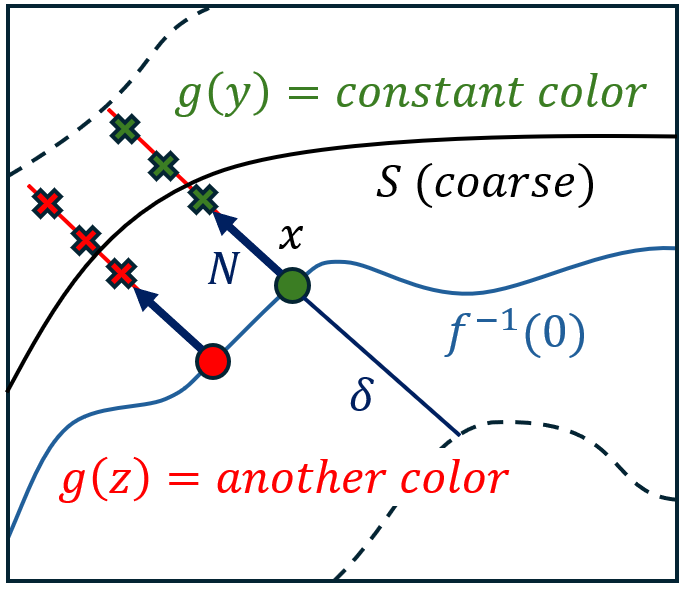}
        \caption{Volumetric texture mapping. The texture $g$ should be constant along the normals $N$ near the coarse surface $S$ (red/green). 
        Having such volumetric representation in the $\delta$-neighborhood ensures that $g$ can be assigned to any point in the coarse surface~$S$.}
        \label{f-texture_mapping_scheme}
    \end{minipage}
\end{figure}

We first introduce a \textbf{multiscale sphere tracing} scheme. Given a view ray \(\gamma(t) = p_0 + t v\), with origin at point \(p_0\), unit direction \(v\), and intersecting the zero level set \(f^{-1}(0)\), the standard sphere tracing (ST) approximates the first intersection point by iterating
\[
p_{i+1} = p_i + v f(p_i)
\]
along \(\gamma\). However, querying a high-capacity neural SDF at each step can be computationally expensive. To reduce this cost, we leverage the multiscale SDF hierarchy \(\{f_i\}\), using coarser networks to guide the early steps of the tracing process.
Thanks to the nesting condition introduced in Eq.~3 of the paper, coarse levels can safely be used to trace offset surfaces before switching to finer levels closer to the surface. The ray initially traces the offset surface \(f_1^{-1}(\delta_1)\) using \(f_1\), proceeds to \(f_2^{-1}(\delta_2)\) with \(f_2\), and finally goes to the target surface \(f_3^{-1}(0)\) using \(f_3\). Each coarser level performs offset tracing via
\[
p_{i+1} = p_i + v \left(f_j(p_i) - \delta_j\right),
\]
which ensures convergence toward the true surface with minimal reliance on high-cost evaluations. Figure~\ref{f-sphere_tracing} illustrates this procedure, focusing on how the ray reaches \(S_3\) by tracing within the neighborhood \(\{|f_2| < \delta_2\}\). For neural SDF inference, we use the GEMM algorithm~\cite{dongarra1990set}.

Importantly, if \(\gamma \cap S_3 \neq \emptyset\), the multiscale ST approximates the first intersection point between \(\gamma\) and \(S_3\). This is guaranteed by the nesting condition, which implies that if \(\gamma \cap S_3 \neq \emptyset\), then necessarily \(\gamma \cap f_2^{-1}(\delta_2) \neq \emptyset\), and thus also \(\gamma \cap f_1^{-1}(\delta_1) \neq \emptyset\). The values \(\delta_i\) play a critical role in this process, as setting them appropriately helps avoid failures, as illustrated in Fig.~4 of the paper. Equation~5 (in the main paper) provides a principled definition of \(\delta_i\), linking them to network training.

Finally, \method{} also accelerates mesh extraction via \textbf{marching cubes}. We propose an adaptive grid inference strategy: we first evaluate the coarse SDF \(f_1\) to cull grid vertices, and only evaluate finer SDFs for vertices inside the \(\delta_1\)-neighborhood. This yields efficient and focused SDF sampling, as depicted in Fig.~5 of the paper. Fig.~\ref{fig:mip_mc_culling} shows reconstructions using this approach.

\section{Normal and texture mapping details}
\label{s-attribute-mapping}
\vspace{-0.2cm}
Let \(S\) be a surface nested within a \(\delta\)-neighborhood of the zero level set of a neural SDF \(f\), i.e., \(S \subset \big[|f| \leq \delta\big]\). 
Assume \(f\) is a finer neural SDF. Then, the \textit{neural normal mapping} assigns to each point \(p \in S\) the attribute
\[
g(p) := \nabla f(p).
\]
This corresponds to restricting \(\nabla f\) to \(S\), effectively transferring the normal of \(f^{-1}(0)\) along the shortest path connecting it to \(p\). Since \(f\) is a signed distance function, the gradient \(\nabla f\) remains constant along such paths.

We consider two cases. First, let \(S\) be a triangle mesh. We use neural normal mapping to transfer detailed normals from the level sets of \(f\) onto \(S\). This approach is analogous to classical normal mapping, which typically relies on UV parameterizations. However, since our method is volumetric, such parameterizations are not required (see Fig.~7 in the paper, middle).

In the second case, let \(S\) be the zero level set of a coarser neural SDF. Here, neural normal mapping allows us to bypass additional sphere tracing iterations (see Fig.~7 in the paper, left). In this case, surface extraction via marching cubes is not necessary.

Similarly, we define a neural network \(g: \mathbb{R}^3 \to \mathcal{C}\) to encode a \textit{texture} over the \(\delta\)-neighborhood of \(f\), where the codomain \(\mathcal{C}\) is the \texttt{RGB} color space. The attribute mapping associated with the triple \(\{S, f, g\}\) is referred to as \textit{neural texture mapping}.

To train the parameters \(\phi\) of \(g\), we use the following loss functional:
\[
\mathscr{T}(\phi) = \int_{f^{-1}(0)} (g - \mathscr{g})^2 \, dx + \int_{\big[|f| \leq \delta\big]} \langle \nabla g, \nabla f \rangle^2 \, dx,
\]
where the first term encourages \(g\) to match the \textit{ground-truth} texture \(\mathscr{g}\), and the second term enforces consistency of \(g\) along the gradient paths of \(f\), regularizing the network within the \(\delta\)-neighborhood. Figure~\ref{f-texture_mapping_scheme} illustrates the texture mapping scheme.

\section{GEMM-based Analytical Normal Calculation for MLPs details}
\label{s-gemm}

We propose a GEMM-based analytical computation of normals, which are continuous and do not need auto-differentiation, resulting in smooth normals.
To compute them, we recall that a MLP with $n-1$ hidden layers has the following form $f(x)\!=\!W_n\circ h_{n-1}\circ \cdots \circ h_{0}(x)+b_n,$ where $h_{i}(x_i)\!=\!\varphi(W_i x_i \!+\! b_i)$ is the i-layer.
The \textit{activation} $\varphi$ is applied on each coordinate of the linear map $W_i\!:\!\R^{N_i}\!\!\to\!\!\R^{N_{i+1}}$ translated by $b_i\!\in\!\R^{N_{i+1}}$.
The gradient of $f$ is given using the \textit{chain~rule}:
\begin{align}\label{e-neural_implicit_gradient}
    \grad{f}(x)\!=\!W_n\cdott \jac h_{n-1} (x_{n-1})\cdott \cdots \cdott \jac h_0(x), \quad\text{with} \quad\jac h_{i}(x_i)=W_i\odot \varphi'\big[a_i|\cdots|a_i\big]
\end{align}
$\jac$ is the \textit{Jacobian}, $x_i\!:=\!\!h_{i-1}\!\circ\! \cdots \!\circ\! h_{0}(x)$,
$\odot$ is the \textit{Hadamard} product, and $a_i\!\!=\!W_i(x_i)\!+\!b_i$.
Eq.~\ref{e-neural_implicit_gradient} is used in~\cite{gropp2020implicit, novello21diff} to compute the level set normals analytically.

We now use Eq.~\ref{e-neural_implicit_gradient} to derive a GEMM-based algorithm for computing the normals ($\nabla f$) in real-time. The gradient $\nabla f$ is given by a sequence of matrix multiplications which is not appropriate for a GEMM setting because $\jac{h_0}(x)\in \R^{3\times N_1}$.
The GEMM algorithm organizes the input points into a matrix, where its lines correspond to the points and its columns organize them and enable parallelism.
We can solve this problem using three GEMMs, one for each normal coordinate. Therefore, each GEMM starts with a column of $\jac{h_0}(x)$, eliminating one of the dimensions. The resulting multiplications can be asynchronous since they are completely~independent.

The $j$-coord of $\grad{f}$ is given by $G_n\!\!=\!\!W_n\!\cdott\! G_{n-1}$, where $G_{n-1}$ is given by iterating $G_{i}\!\!=\!\!\jac h_{i}(x_{i})\!\cdott\! G_{i-1}$, with the initial condition $G_0=W_0[j] \odot \varphi'(a_0)$. The vector $W_0[j]$ denotes the $j$-column of $W_0$.
We use a kernel and a GEMM to compute $G_0$ and $G_n$. For $G_{i}$ with $0\!<\!i\!<\!n$, observe~that
\begin{align*}\label{e-Gi}
G_{i}\!=\!\left(W_i\odot \varphi'\left[a_i|\cdots|a_i\right]\right)\cdott G_{i-1}\!=\!(W_i\cdott G_{i-1})\odot \varphi'(a_i).
\end{align*}
The first equality comes from Eq.~\ref{e-neural_implicit_gradient} and the second from a commutative property of the Hadamard product.
The second expression needs fewer computations and is solved using a GEMM followed by a kernel.

Algorithm~\ref{a-normal_computation} presents the gradient computation for a batch of points. 
The input is a matrix $P\in\R^{3\times k}$ with columns storing the $k$ points generated by the GEMM version of the sphere tracing algorithm. 
The output is a matrix $\grad{f_\theta}(P)\in\R^{3\times k}$, where its $j$-column is the gradient of $f_\theta$ evaluated at $P[j]$.
Lines $2-5$ are responsible for computing $G_0$, Lines $6-11$ compute $G_{n-1}$, and Line $13$ provides the result gradient $G_n$.
Table~\ref{tab:time_auto} shows a comparison between this algorithm and automatic differentiation using PyTorch.

\begin{figure}[htb]
    \centering
    \begin{minipage}[t]{0.48\textwidth}
        \scriptsize
        \centering
        \captionof{table}{Runtime comparison, in seconds, between Pytorch autograd and our algorithm to calculate the normals. Ours performs $2\times$ faster. Samples evaluated are from Stanford dataset.}
        \begin{tabular}{lrrr}
        \toprule
        \textbf{Model}           & \textbf{Autograd} & \textbf{Ours}    & \textbf{Resolution} \\ \midrule
        Armadillo 256x2 & 0.007    & \textbf{0.003}   & 512x512        \\ 
        Armadillo 256x2 & 0.024    & \textbf{0.010}   & 1024x1024       \\ 
        Armadillo 256x3 & 0.010    & \textbf{0.005}   & 512x512        \\ 
        Armadillo 256x3 & 0.025    & \textbf{0.012}   & 1024x1024       \\ \hline
        Buddha 256x2    & 0.008    & \textbf{0.005}   & 512x512        \\ 
        Buddha 256x2    & 0.021    & \textbf{0.014}   & 1024x1024       \\ 
        Buddha 256x3    & 0.011    & \textbf{0.005}   & 512x512        \\ 
        Buddha 256x3    & 0.024    & \textbf{0.012}   & 1024x1024       \\ \hline
        Lucy 256x2      & 0.007    & \textbf{0.004}   & 512x512        \\ 
        Lucy 256x2      & 0.021    & \textbf{0.012}   & 1024x1024       \\ 
        Lucy 256x3      & 0.011    & \textbf{0.007}   & 512x512        \\ 
        Lucy 256x3      & 0.025    & \textbf{0.015}   & 1024x1024       \\ \bottomrule
        \end{tabular}
        \label{tab:time_auto}
    \end{minipage}
    \hfill
    \begin{minipage}[t]{0.48\textwidth}
        {\small
        \begin{algorithm}[H]
        \SetAlgoLined
         \KwIn{neural SDF $f_{\theta}$, positions $P$}
         \KwOut{Gradients $\grad{f_\theta}(P)$}
        
        \For{ $j = 0$ to $2$ (async)} 
        {
            using a GEMM: \tcp{Input Layer}
                \Indp $A_0 = W_0\cdott P + b_0$ \\
            \Indm using a kernel: \\
                \Indp $G_0 = W_0[j] \odot \varphi'(A_0)$;\,
                $P_0 = \varphi(A_0)$ \\
            \Indm
            \tcp{Hidden layers}
            \For{layer $i=1$ to $n-1$}
            {
                    using GEMMs: \\
                        \Indp $A_i = W_i\cdott P_{i-1} + b_i$;\,
                        $G_i = W_i \cdott G_{i-1}$ \\
                    \Indm using a kernel: \\
                        \Indp $G_i = G_i \odot \varphi'( A_i)$;\,
                        $P_i = \varphi(A_i)$ \\
            }
            using a GEMM: \tcp{Output layer}
                \Indp $G_n = W_n \cdott G_{n-1}$
        }
         \caption{Normal computation}
         \label{a-normal_computation}
        \end{algorithm}}
    \end{minipage}
\end{figure}

\section{Additional applications}
\subsection{Surface reconstruction from posed images}

We demonstrate that \method{} can be seamlessly integrated into image-based reconstruction pipelines, such as NeuS~\cite{wang2021neus}. To this end, we replace the standard neural SDF used in NeuS with our multiscale SDF architecture, composed of two neural networks: a coarse-level network and a higher-resolution refinement network.
We compare our modified pipeline against the baseline NeuS to evaluate the impact of our approach both quantitatively and qualitatively. Specifically, we implement the coarse network with 4 hidden layers of 128 neurons each, and the fine-level network with 5 hidden layers of 256 neurons. Despite having 37\% fewer parameters than the original NeuS architecture, our multiscale approach achieves improved performance.
Quantitatively, under the default volume rendering configuration, our method yields an average PSNR improvement of 3.74\% across models from the DTU dataset~\cite{jensen2014dtu}. Table~\ref{neus} summarizes the PSNR comparisons between our approach and the baseline NeuS.
\begin{table}[h]
\centering
\caption{PSNR comparison between the baseline NeuS and our multiscale method on selected scans from the DTU dataset~\cite{jensen2014dtu}. Our method consistently improves reconstruction quality while using fewer network parameters. All values are reported in dB.}
\begin{tabular}{lcccccc}
\toprule
          & Scan 24 & Scan 37 & Scan 40 & Scan 55 & Scan 63 & Avg.   \\
\midrule
NeuS      & 28.20   & 27.10   & 28.13   & 28.80   & 32.05   & 28.86  \\
Ours      & 31.50   & 26.50   & 27.78   & 29.01   & 34.89   & 29.94  \\
\bottomrule
\end{tabular}
\label{neus}
\end{table}

\begin{figure}[!h]
\centering
\includegraphics[width=0.9\textwidth]{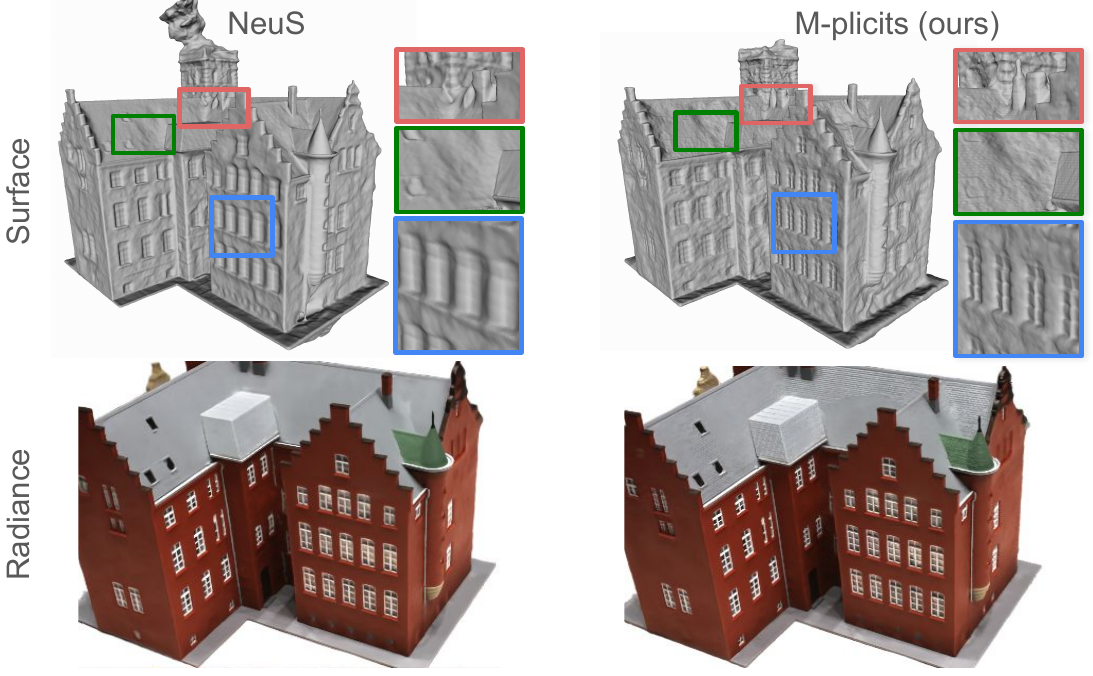}
\caption{Comparison between baseline NeuS and our multiscale variant on Scan 24 of the DTU dataset. Top: extracted surface meshes with zoomed-in details. Bottom: rendered appearance. Our approach recovers finer geometric details, as evident in architectural structures and window boundaries.}
\label{f-neus}
\end{figure}

Figure~\ref{f-neus} presents a comparison between NeuS and our multiscale variant on Scan 24 of the DTU dataset. The top row shows the reconstructed mesh geometry, including zoomed-in insets that highlight fine surface details. The bottom row displays renderings showing that ours provide reconstruction with sharp details. Our method yields improved geometric fidelity and cleaner surface reconstructions, especially in regions with architectural features.

\subsection{Neural implicit surface evolution}
\begin{figure}[h]
\centering
    \includegraphics[width=0.9\textwidth]{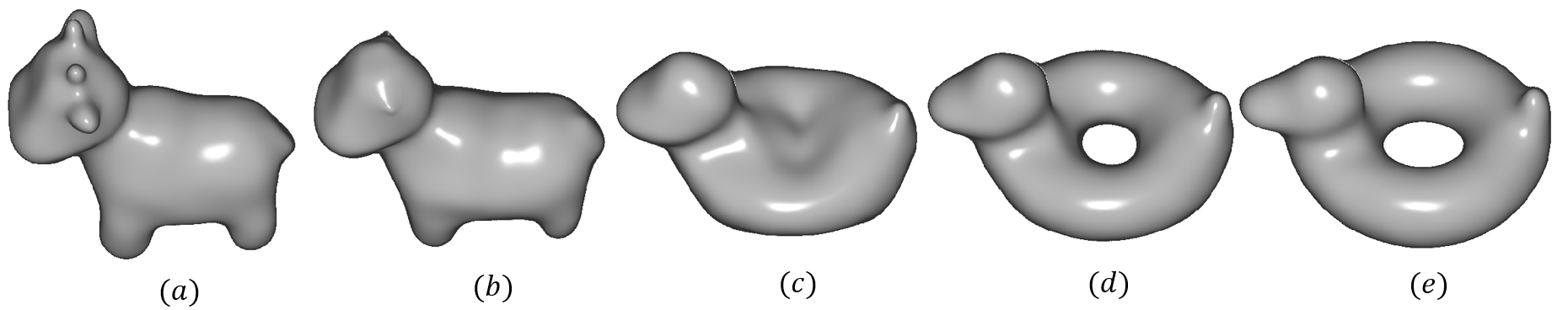}
    \caption{A dynamic multiscale SDF is trained using the pipeline from \cite{novello21neuralAnimation}. Note the change in topology (c-d), which is challenging to handle using meshes. Also, octree/mesh-based approaches require generating a surface for each time, an overhead that our model avoids.}
    \label{f-4d}
\end{figure}

Note that neural SDFs provide a smooth representation of a static scene. By adding an additional input coordinate, we can encode time into the representation. We leverage this approach to train dynamic evolutions of static neural SDFs, following the training schemes introduced in \cite{novello21neuralAnimation}. Fig.~\ref{f-4d} presents an example of interpolation between the Spot and Bob models using this method. Importantly, the implicit model handles topology changes, demonstrating that our representation can be integrated into differentiable pipelines. The visualization is in real-time (120 FPS) using an extension of our multiscale ST to dynamic SDFs.

\subsection{Additional experiments}
\label{s-additional_experiments}

\paragraph{Normals.} We compare our GEMM normal calculation against \texttt{torch.autograd}. As shown in Tab.~\ref{tab:time_auto}, ours performs $2\times$ faster. We tested $6$ different INRs trained for Stanford dataset samples: Armadillo, Buddha, and Lucy, varying between 2-3 hidden~layers.

\paragraph{Point cloud from images:} 

\begin{figure}[!h]%
        \centering
        \includegraphics[width=\linewidth]{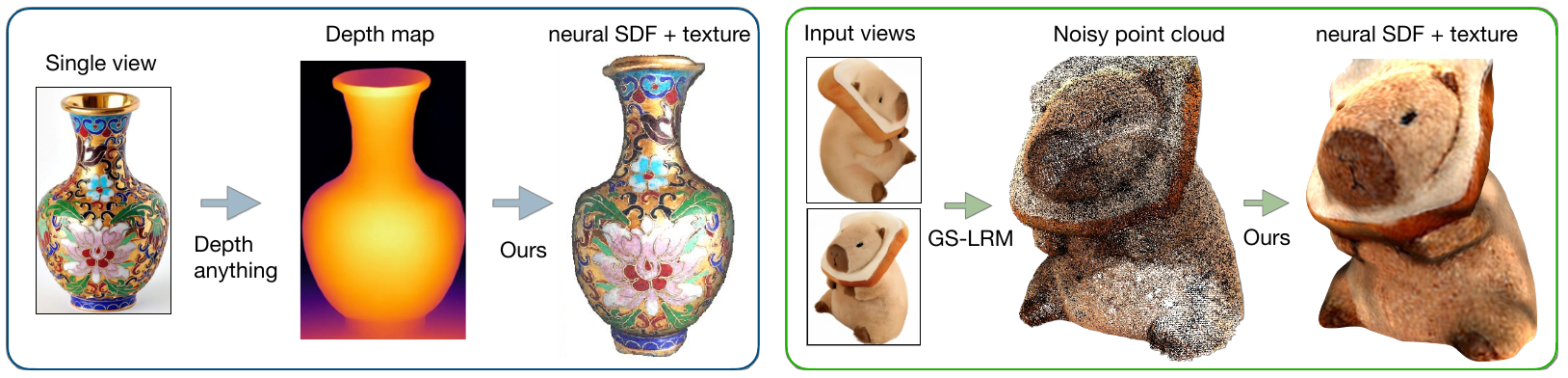}
            \caption[Training a textured SDF from images/noisy point cloud.]{Training a textured SDF from images/noisy point cloud. On the left, our model (neural SDF + texture) is trained using the unprojection of a depth map, which is computed from a single view using Depth Anything. The resulting vase is rendered at 32.1 FPS. On the right, we show a reconstruction derived from a noisy point cloud, extracted from multiple views using GS-LRM~\protect\cite{zhang-gslrm2024}. By combining our method with this feed-forward 3D model (GS-LRM), we achieve fast reconstruction of the SDF with texture.}
    \label{fig:point-cloud}
\end{figure}

Fig~\ref{fig:point-cloud} shows our model trained with a point cloud reconstructed from an image. We use Depth Anything~\cite{Yang_2024_CVPR} to generate the depth of the pixels and use that depth to create the point cloud based on the view.

\begin{figure}[!h]
    \centering
    \includegraphics[width=\linewidth]{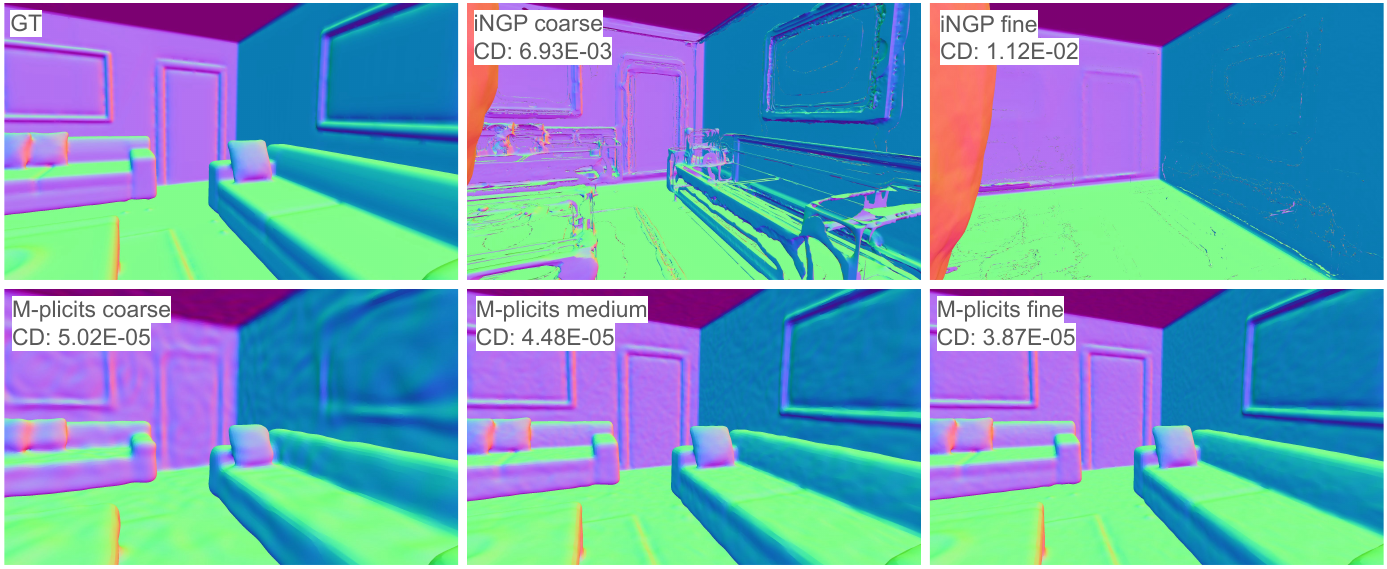}
    \caption{
        \textbf{Scene-scale reconstruction (10M+ points).} 
        Comparison between Instant-NGP (coarse and fine) and M-plicits at three scales (coarse, medium, fine). 
        Instant-NGP fails to reconstruct large regions of the geometry and introduces strong artifacts, even when increasing the number of levels. 
        In contrast, M-plicits yields accurate and stable reconstructions at all scales, closely matching the ground-truth surface, as reflected in the reported Chamfer distances.
    }
    \label{fig:room}
\end{figure}

\paragraph{Scene-scale point cloud.}
We also evaluated our method on a scene-scale point cloud containing more than 10 million points. 
Figure~\ref{fig:room} shows that M-plicits successfully reconstructs the entire scene across all scales (coarse, medium, and fine), whereas Instant-NGP fails in our tests using both 3 levels (coarse) and 16 levels (fine). 
The Chamfer distance further corroborates these observations: M-plicits achieves {2--3 orders of magnitude lower Chamfer distance} compared to Instant-NGP.

\begin{figure*}
    \centering
    \begin{tabular*}{0.8\textwidth}{@{\extracolsep{\fill}}cccc}
    \textbf{Coarse} & \textbf{Normal mapping} & \textbf{Multiscale ST} & \textbf{Baseline} \\
    \end{tabular*}
    \includegraphics[width=\textwidth]{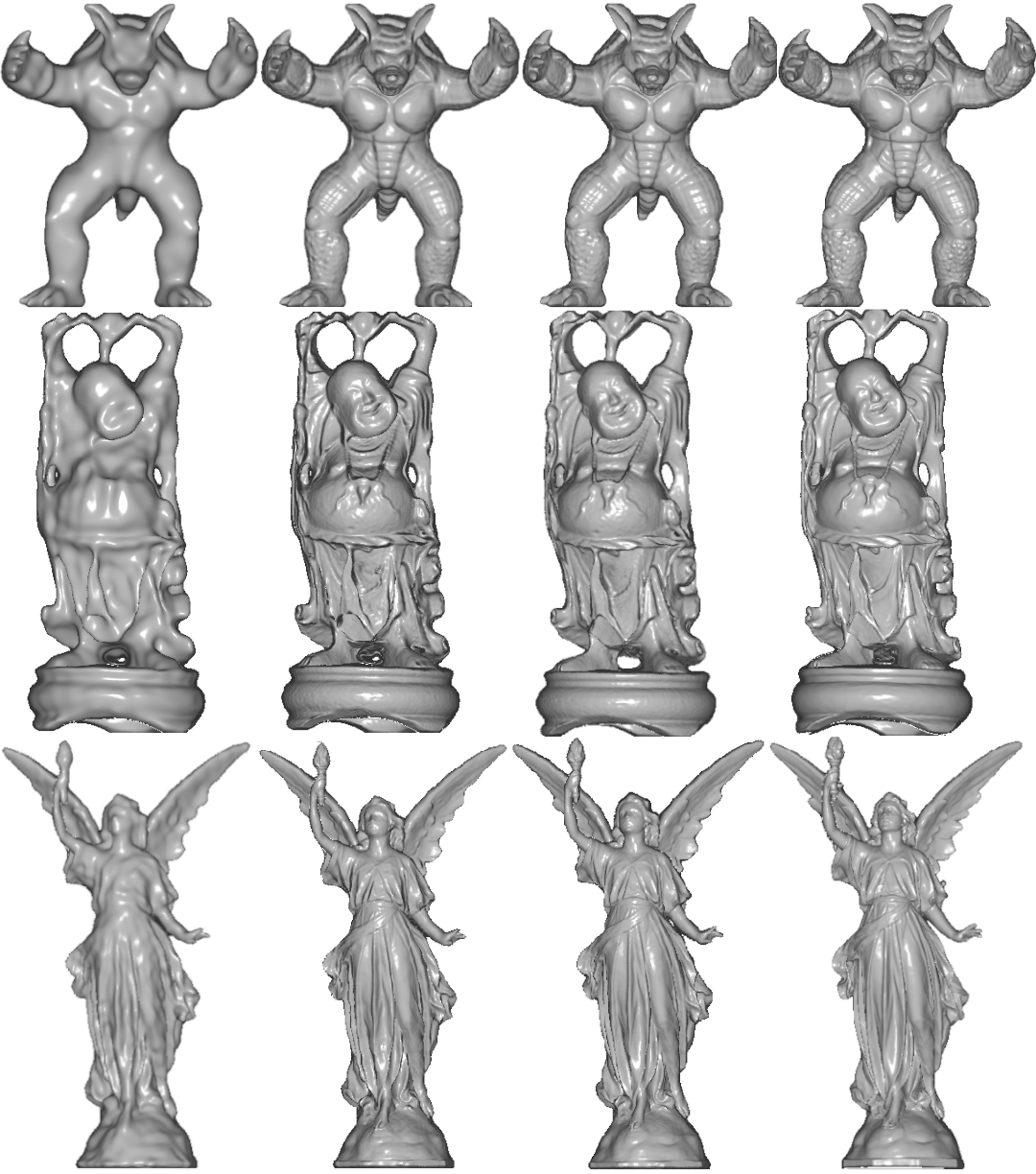}
    \caption{Comparison between our method and the SIREN baseline. The columns represent different configurations. From left to right: $\boldsymbol{(64, 2)}$, $\boldsymbol{(64, 2) \rhd (256, 2)}$, and the baseline $\boldsymbol{(256, 4)}$. The second column uses neural normal mapping and the third uses multiscale sphere tracing. Notice that fidelity is improved in the second column and the third column refines the results.}
    \label{f-rendering}
\end{figure*}

\begin{figure*}
\centering
    \begin{tabular}{cc}
    \vspace{0.3cm}
    \textbf{Baseline} & \textbf{Ours} \\
    \includegraphics[width=0.49\textwidth]{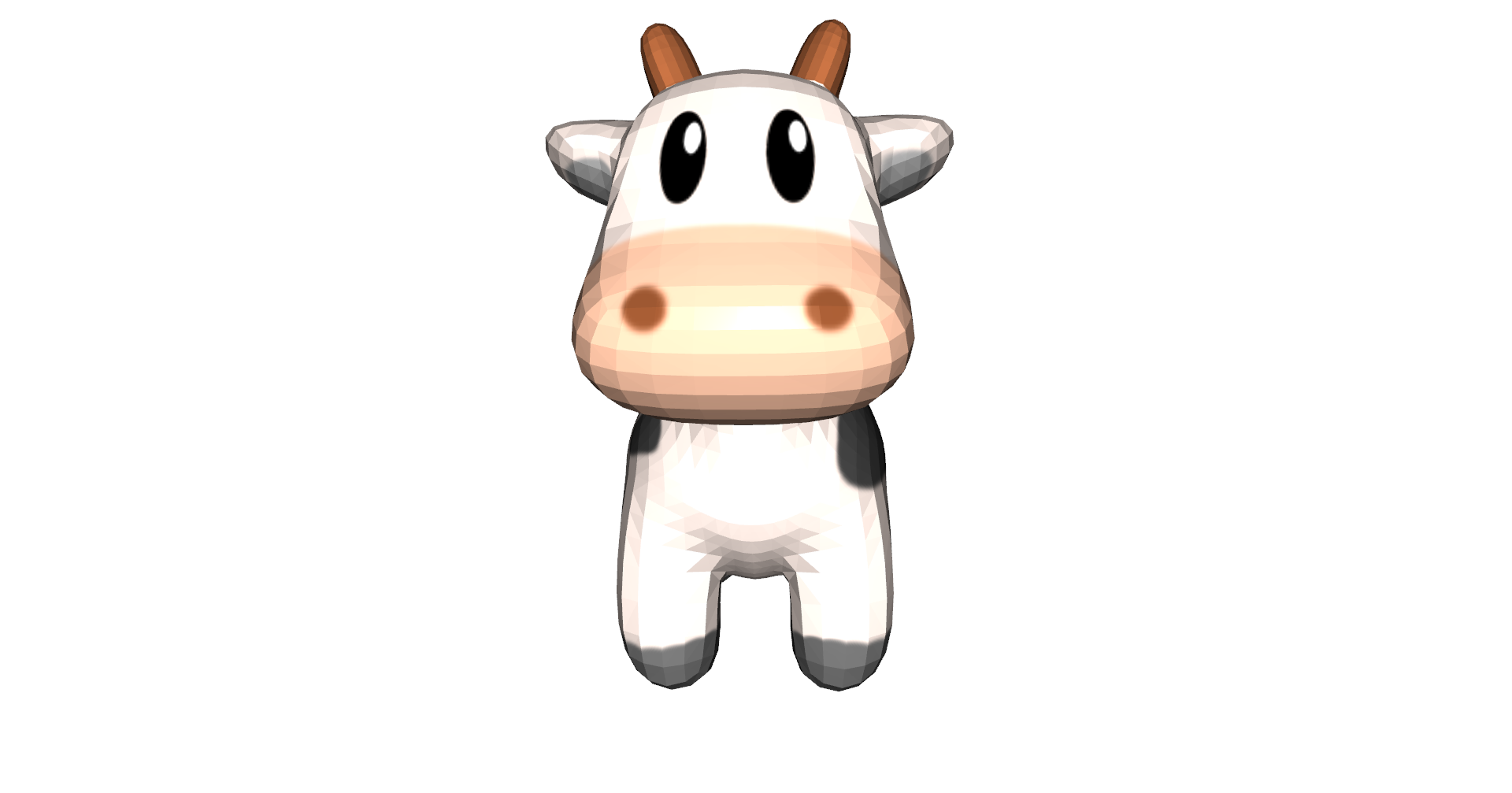} &
    \includegraphics[width=0.49\textwidth]{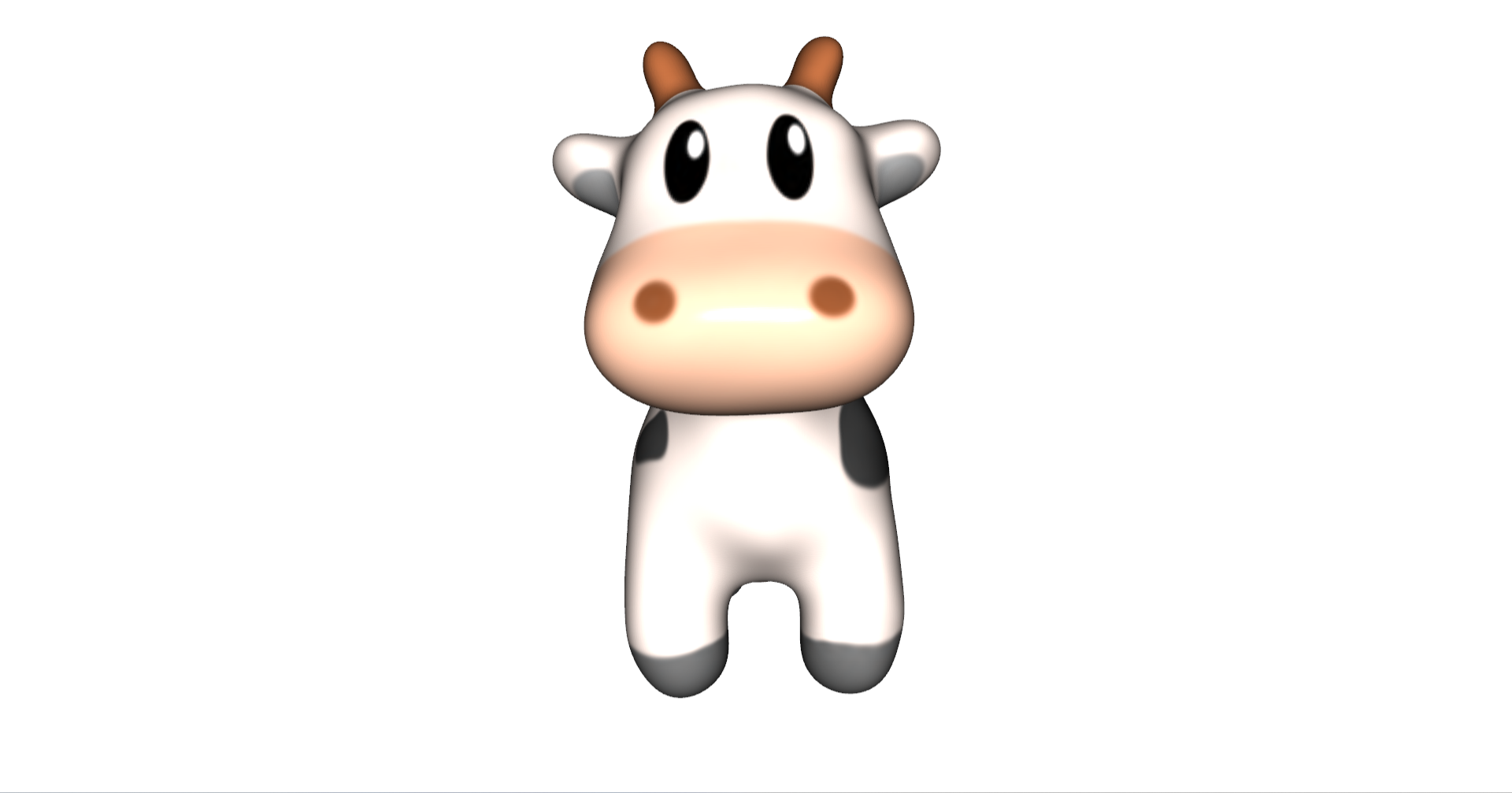} \\
    \includegraphics[width=0.49\textwidth]{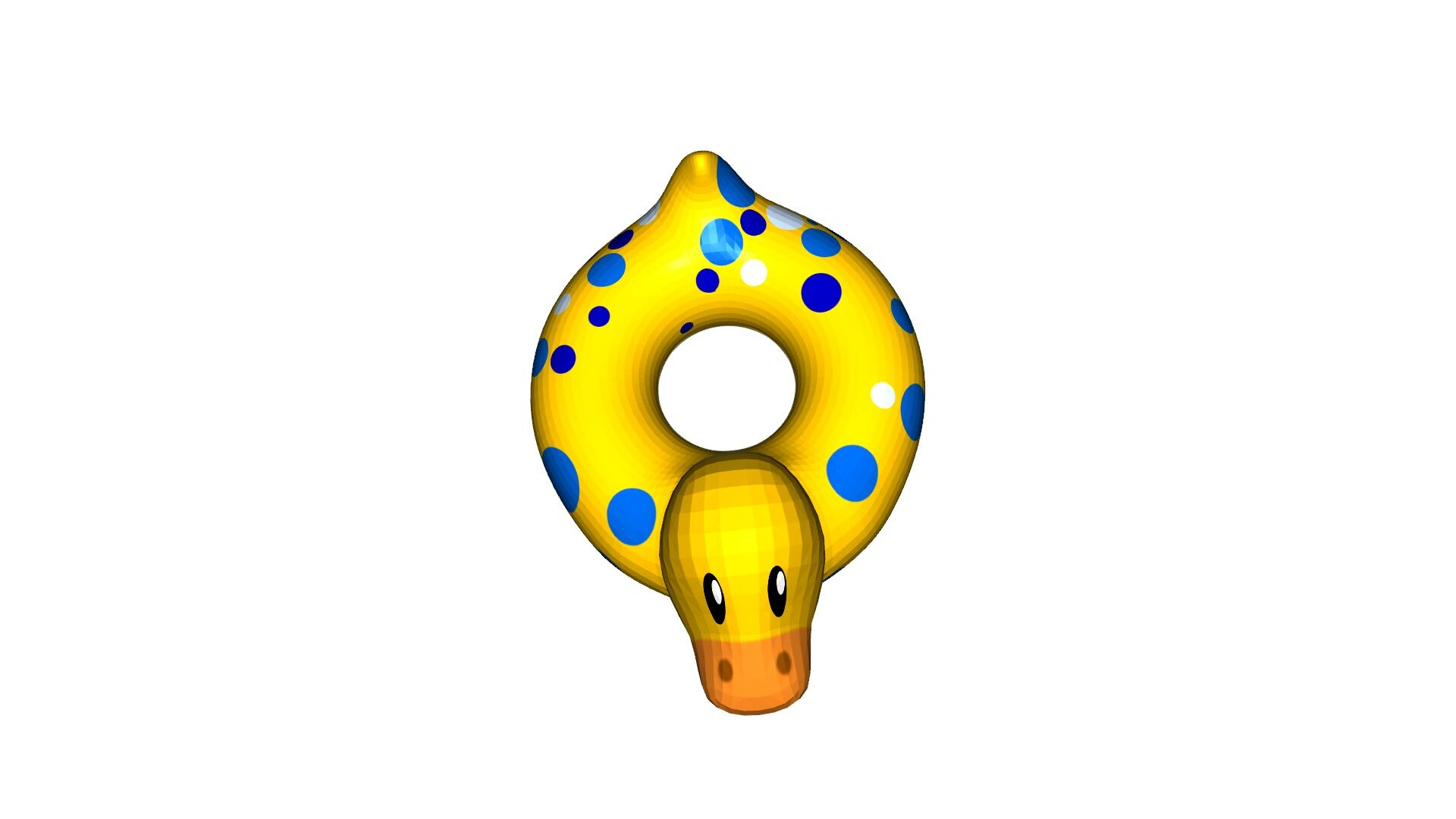} &
    \includegraphics[width=0.49\textwidth]{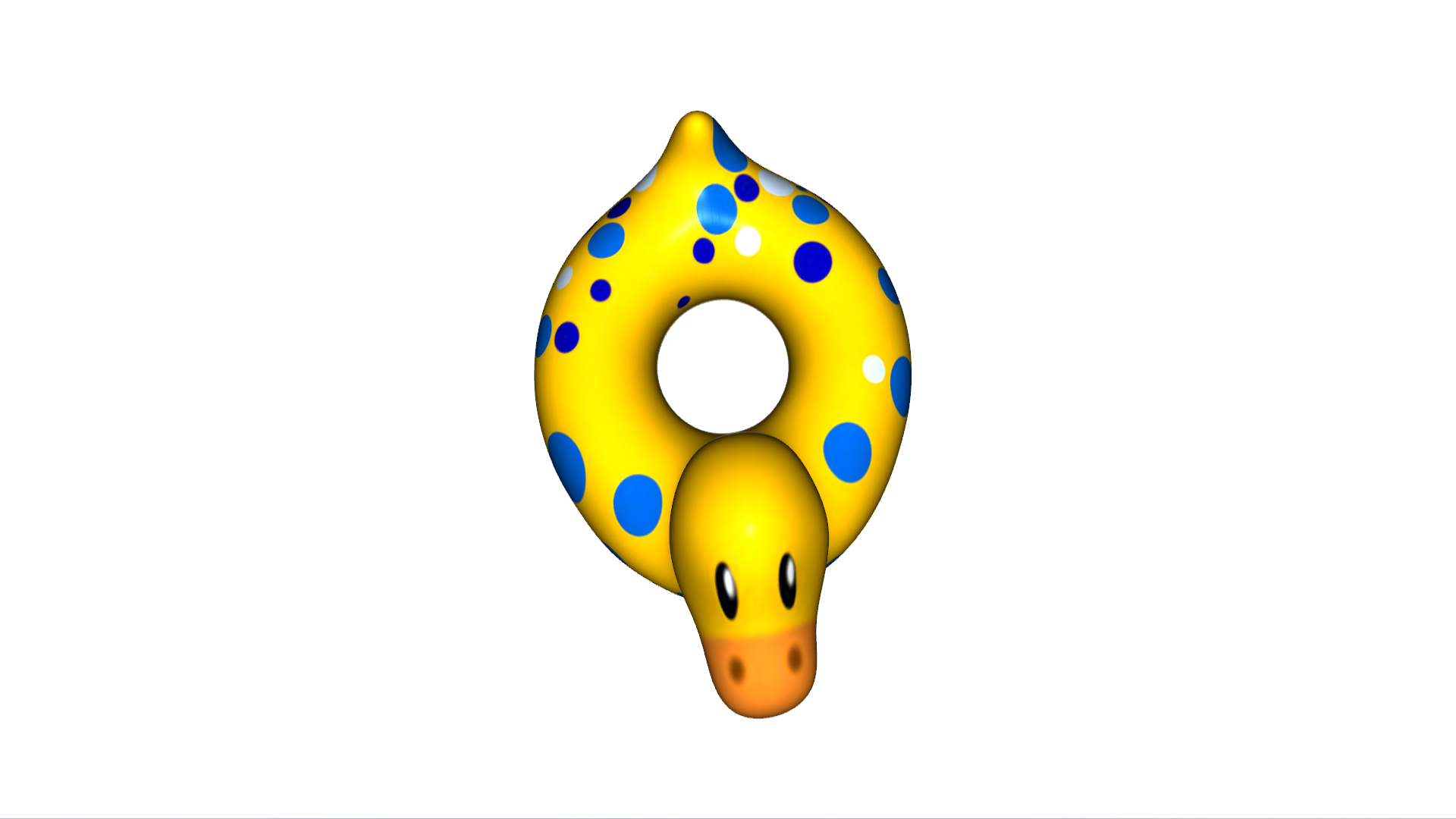} \\
    \includegraphics[width=0.49\textwidth]{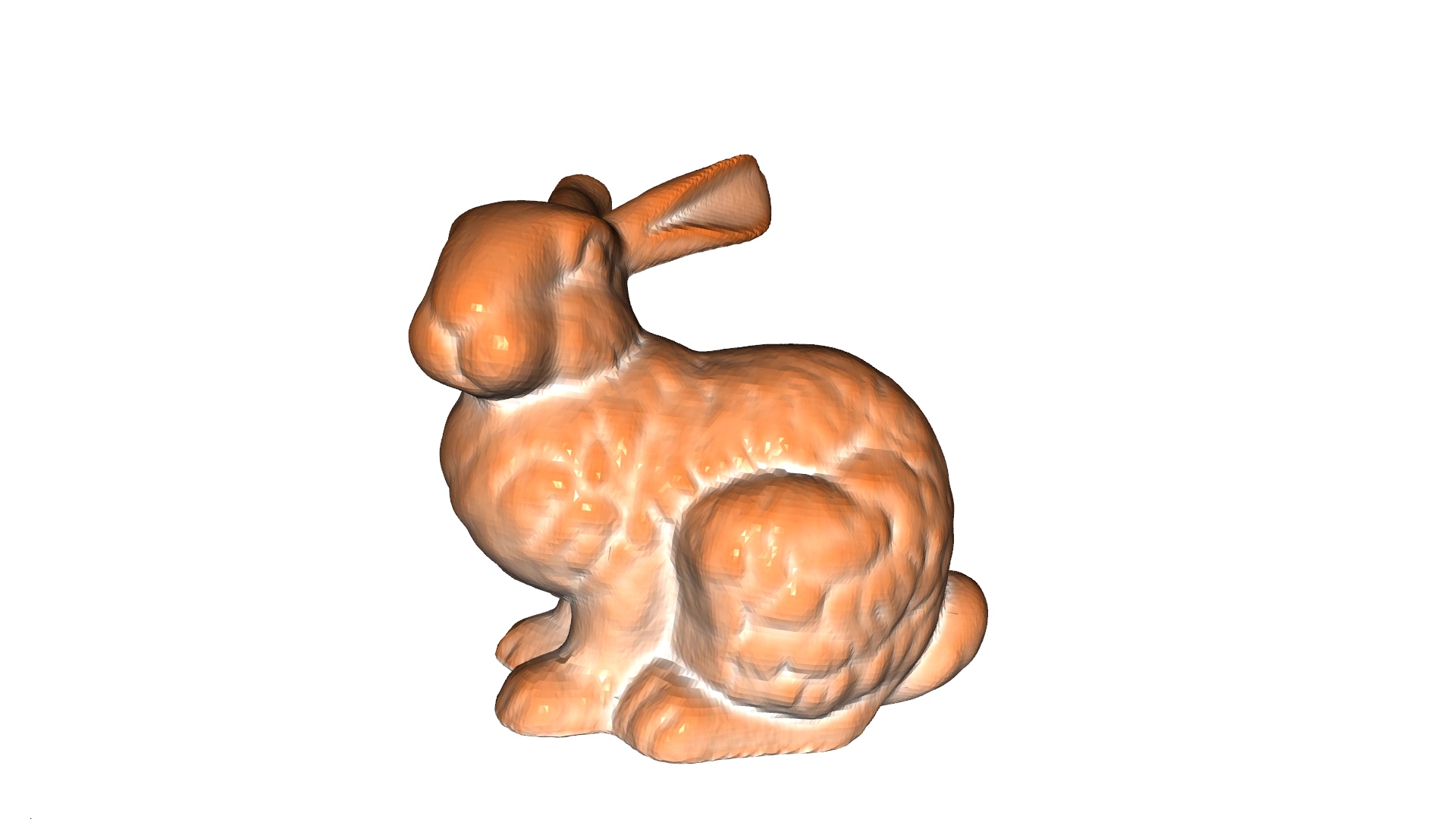} & 
    \includegraphics[width=0.49\textwidth]{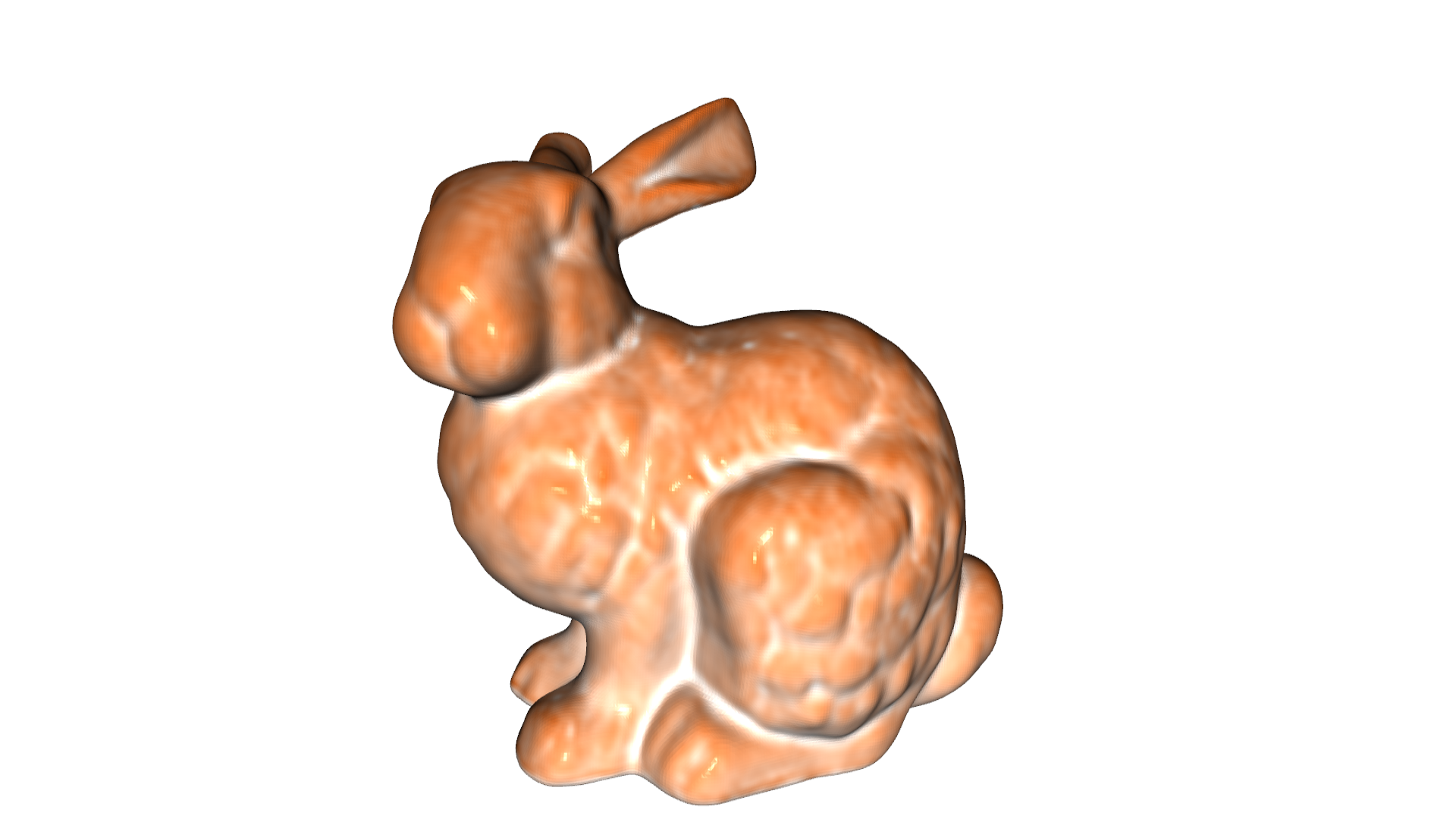} \\
    \includegraphics[width=0.49\textwidth]{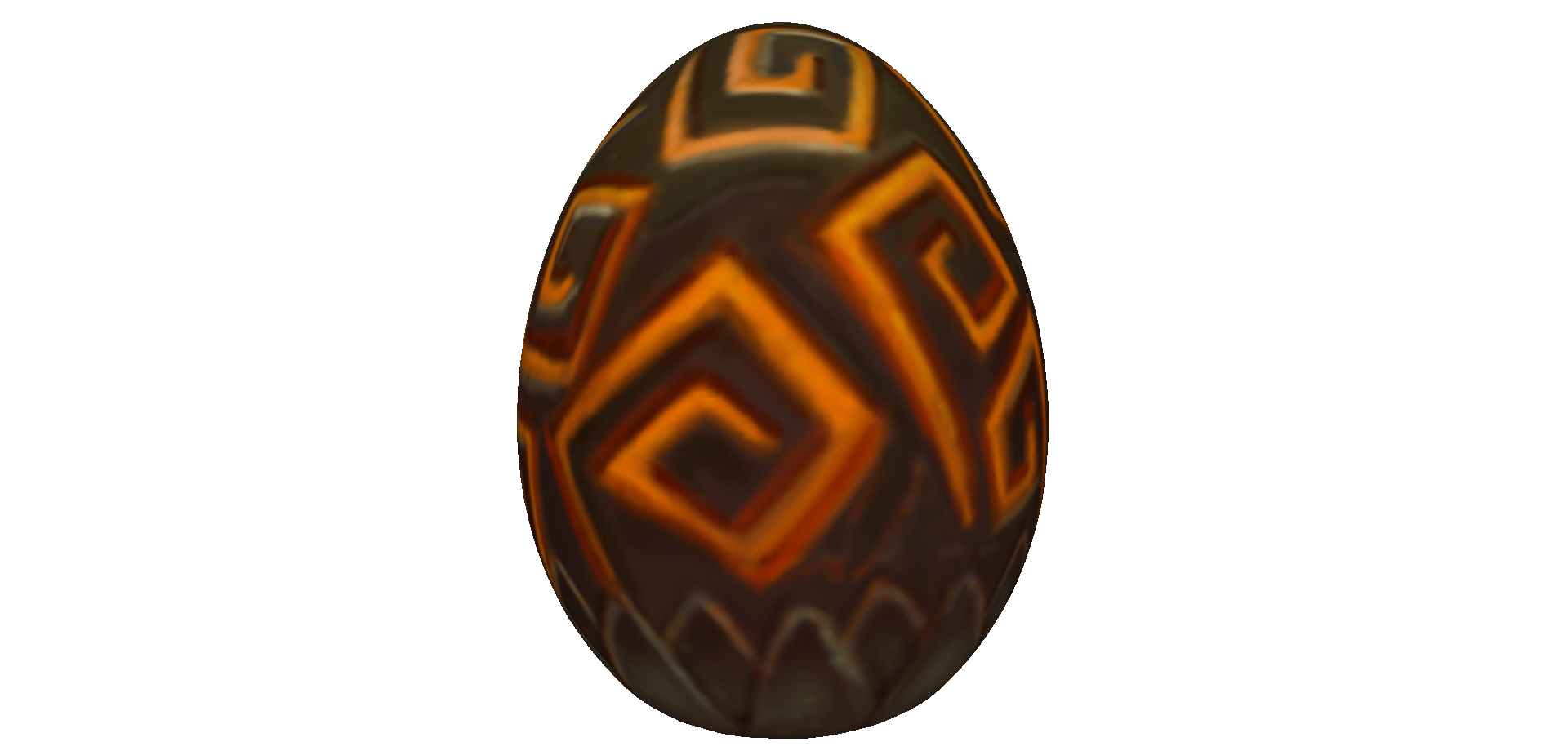} &
    \includegraphics[width=0.49\textwidth]{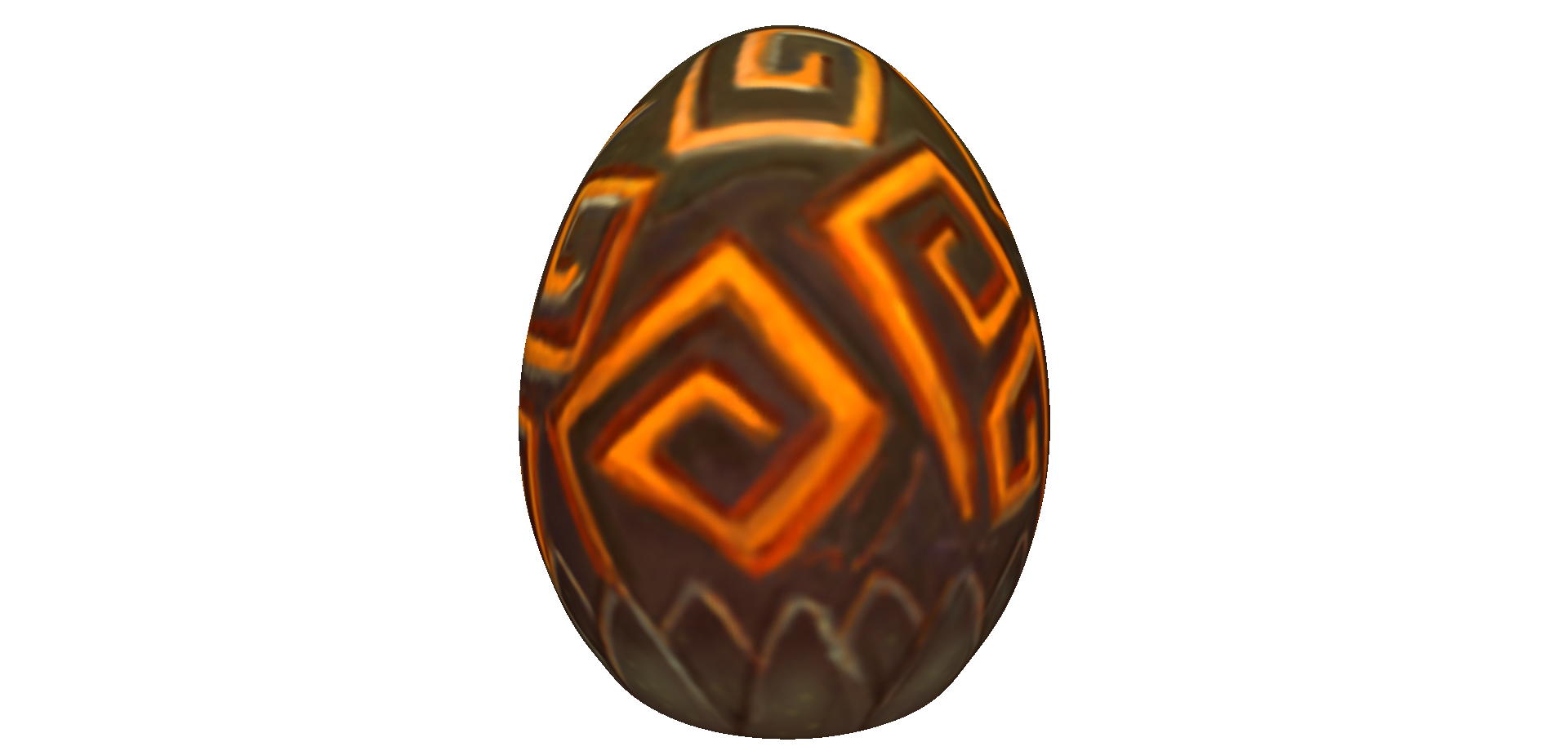} \\
    \includegraphics[width=0.49\textwidth]{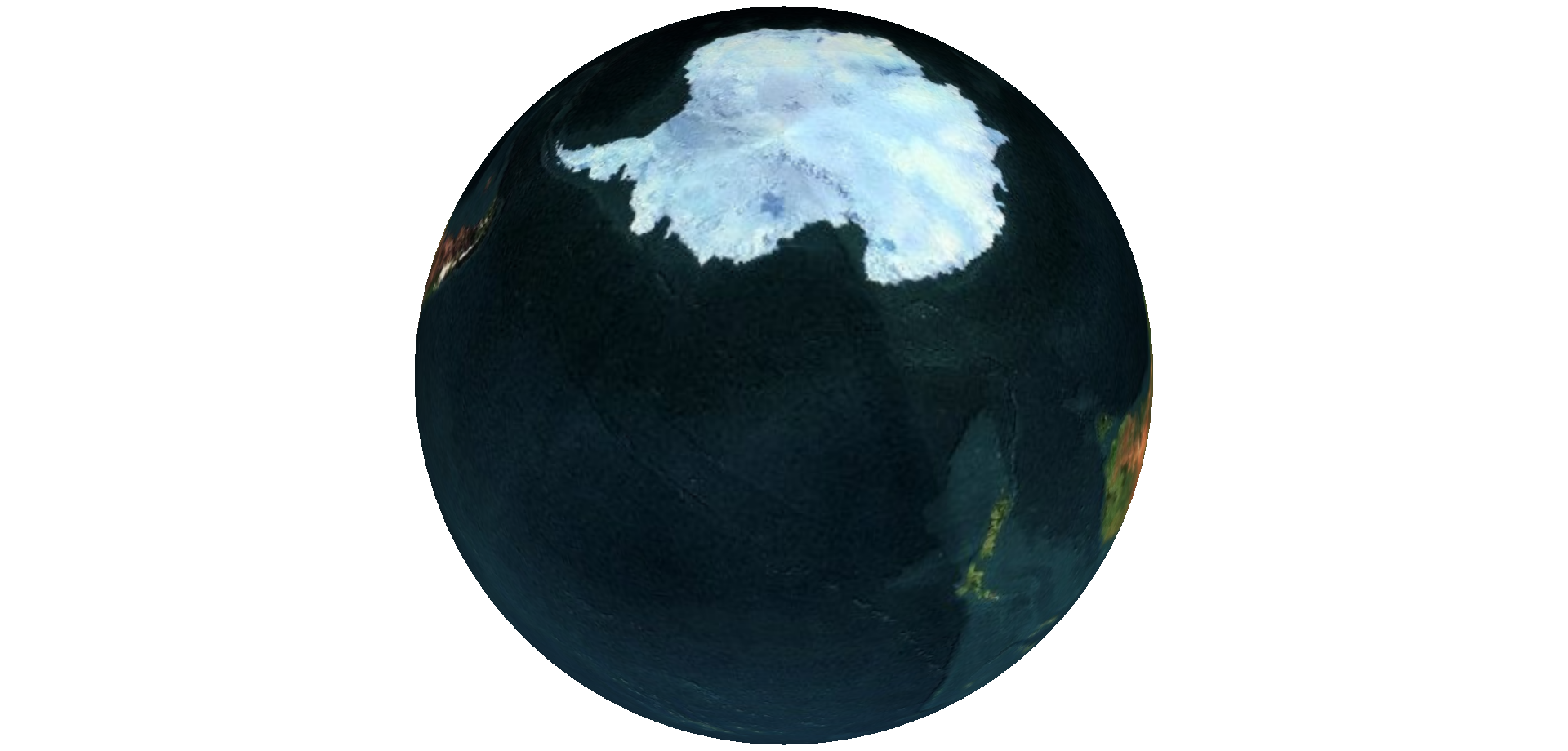} &
    \includegraphics[width=0.49\textwidth]{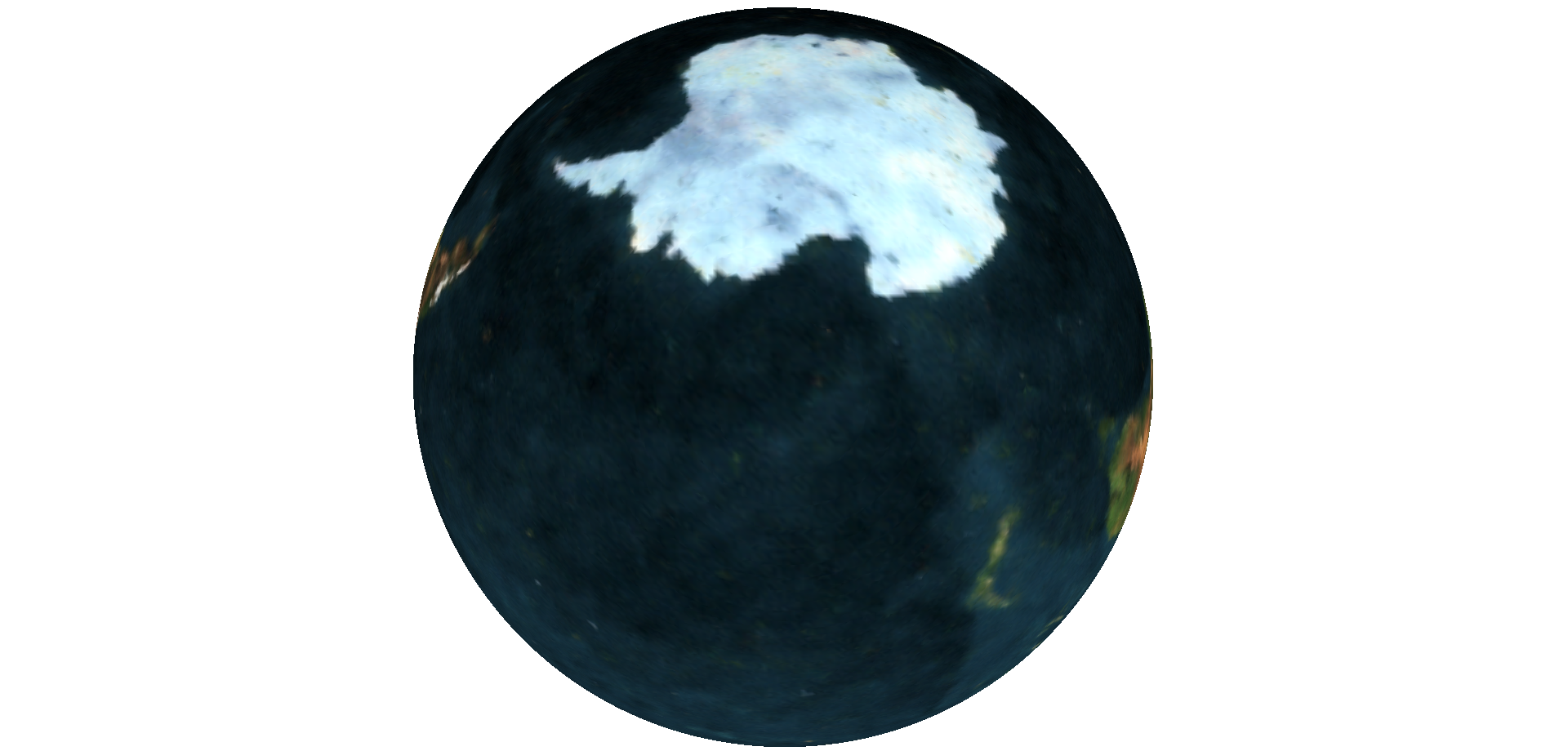}
    \end{tabular}
    \caption{Images we use to calculate the MSE between the ground-truth textured meshes and our approach.}
    \label{f-nerf}
\end{figure*}

\paragraph{Broader perceptual evaluation:} Fig.~\ref{f-rendering} shows a broader perceptual evaluation of the multiscale sphere tracing and the neural normal mapping using several models. Fig.~\ref{f-nerf} also shows the images we use to calculate the MSE to compare the neural texture mapping with the rendering baseline.

\begin{figure}[t]
    \centering
    \includegraphics[width=0.45\linewidth]{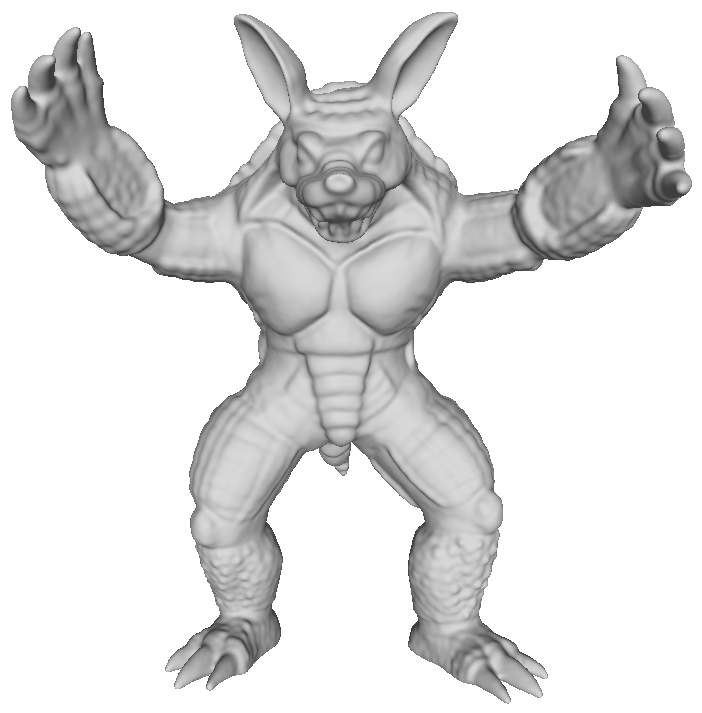}
    \includegraphics[width=0.22\linewidth]{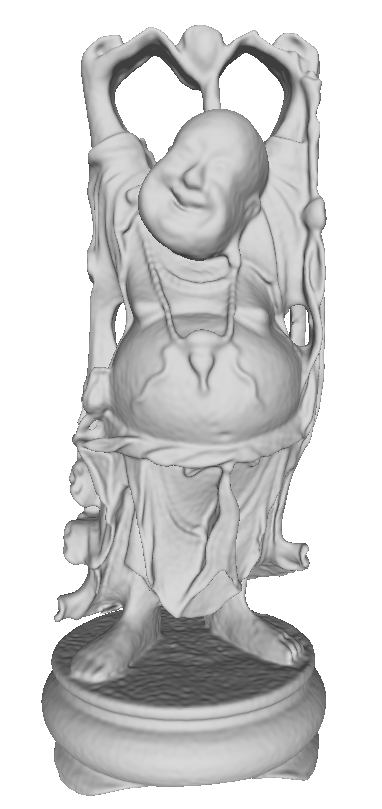}
    \includegraphics[width=0.31\linewidth]{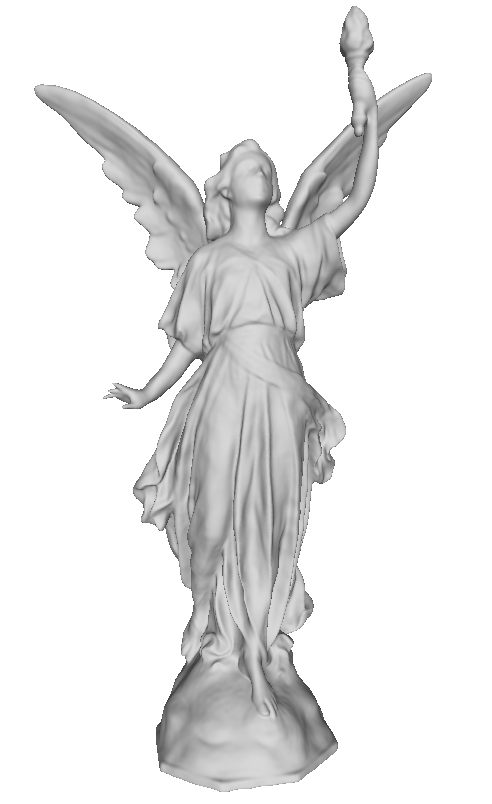}
    \caption{From left to right: Marching cubes reconstruction of Armadillo, Buddha and Lucy using our proposed grid culling method.}
    \label{fig:mip_mc_culling}
\end{figure}

\paragraph{Accelerated Marching Cubes qualitative evaluation:} Fig.~\ref{fig:mip_mc_culling} shows high-fidelity reconstructions computed using our acceleration for the marching cubes algorithm.

\paragraph{Residuals.} 
To evaluate our nested neighborhood scheme and loss design, we compare against a \textbf{baseline} residual variant \(f_2 = f_1 + r_1\), where both \(f_1\) and \(r_1\) are trained using SIREN’s original loss and sampling strategy, following the baseline in IDF~\cite{wang2022geometry}. 
In contrast, our method employs the loss in Eq. 4 in the main paper together with \emph{nested neighborhoods} to supervise the ground-truth SDF within a narrow band around the previous stage’s zero level set. This exploits the property \(f(\mathbf{x}_j + t\,\mathbf{N}_j) = t\) within a tubular neighborhood, enabling supervision beyond the original samples \(\mathbf{x}_j\) to improve stability and prevent error growth.  
This strategy also reduces network complexity, allowing us to use much smaller architectures: a base network with a single hidden layer of 128 neurons (\(\omega_0=30\)) and \(r_1\) with a single hidden layer of 256 neurons (\(\omega_0=45\)).  

We compare this baseline with \method{} on the Thingi32 dataset, where the baseline achieves an average CD of 6.2e{-2}, while our method reaches 1.2e{-2}, demonstrating that \method{} yields substantially better reconstructions.  
We also evaluate the residual approach qualitatively. Fig. 13 in the main paper shows that residuals decrease spurious components when combined with neighborhood training. We further exploit this property to accelerate marching cubes in scenarios where mesh extraction is required.  

\begin{figure}[h]
\centering
    \includegraphics[width=0.8\textwidth]{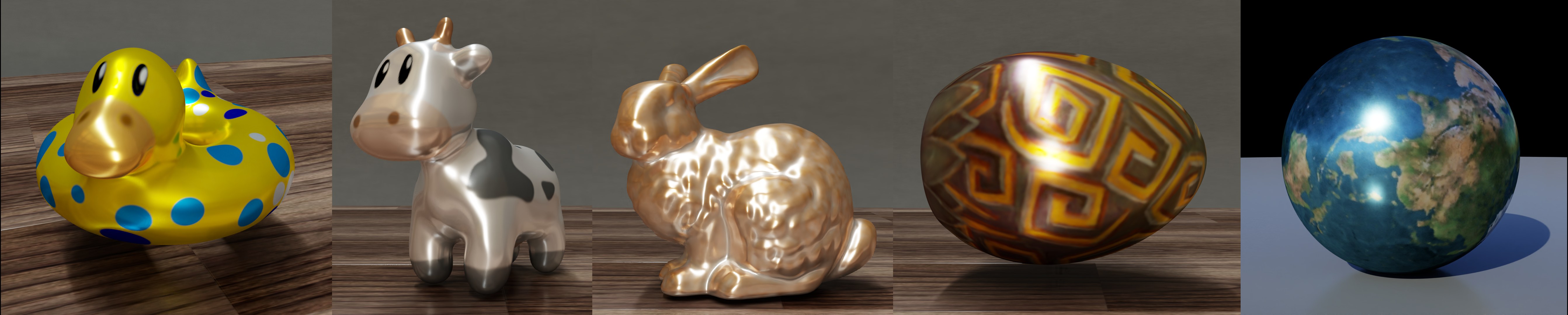}
    \caption{{Neural texture mapping}. All networks are ${(256, 3)}$, except for the the earth, which is ${(512,3)}$. The surfaces are marching cubes of ${(64,1)}$ SDFs, except for the bunny, which is ${(128,2)}$. 
    }
    \label{f-neural_texture}
\end{figure}

\paragraph{Textures.}
We define textures directly in a neighborhood of the surface, removing the need for a UV map. 
This formulation produces visually convincing appearance while decoupling texture from geometry in a compositional way. 
To assess accuracy, we compared our approach against traditional UV-textured meshes by measuring the MSE between rendered images. 
Across five test models—Spot, Bob, Bunny, Egg, and Earth—we obtained MSEs of 0.0329, 0.0434, 0.0720, 0.0291, and 0.0033, respectively. 
Figure~\ref{f-neural_texture} illustrates neural texture mapping applied to coarse surfaces.  

\paragraph{Scene scale test.} We also evaluated our method on a scene-scale point cloud (Figure~\ref{fig:room}). M-plicits successfully reconstructs the entire scene across all scales, whereas Instant-NGP fails in our tests. The Chamfer distance further corroborates these observations: M-plicits achieves {2--3 orders of magnitude lower Chamfer distance}.

\paragraph{Normals.} We compare our GEMM normal calculation against \texttt{torch.autograd}.
Our method performs $2\times$ faster across $6$ different INRs trained for the Armadillo, Buddha and Lucy, with architectures varying between 2 and 3 hidden~layers. More details are provided in Tab.~\ref{tab:time_auto}.

\end{document}